\documentclass{article}

\usepackage[table,xcdraw]{xcolor}
\usepackage{microtype}
\usepackage{graphicx}
\usepackage{subfigure}
\usepackage{booktabs} 

\usepackage{hyperref}

\usepackage[accepted]{icml2024}

\usepackage{amsmath}
\usepackage{amssymb}
\usepackage{mathtools}
\usepackage{amsthm}

\usepackage[capitalize,noabbrev]{cleveref}

\theoremstyle{plain}

\theoremstyle{definition}

\theoremstyle{remark}

\usepackage[textsize=tiny]{todonotes}

\usepackage{amsmath,amsfonts,bm}

\def\eqref#1{equation~\ref{#1}}

\def\1{\bm{1}}

\def\vtheta{{\bm{\theta}}}

\def\vb{{\bm{b}}}

\def\vh{{\bm{h}}}

\def\vp{{\bm{p}}}

\DeclareMathAlphabet{\mathsfit}{\encodingdefault}{\sfdefault}{m}{sl}
\SetMathAlphabet{\mathsfit}{bold}{\encodingdefault}{\sfdefault}{bx}{n}

\newcommand{\softmax}{\mathrm{softmax}}

\newcommand{\nl}[1]{\textit{``{#1}''}}

\DeclareMathOperator*{\argmax}{arg\,max}

\def\methodName{\texttt{Patchscope}}
\def\methodNamePlural{\texttt{Patchscopes}}
\def\methodFamilyName{\texttt{Patchscopes}}

\def\sourceSeqLen{n}
\def\sourcePosition{i}
\def\sourcePrompt{S}
\def\sourceLayer{\ell}
\def\numSourceLayers{L}
\def\sourceModel{\mathcal{M}}
\def\sourceHidden{\vh}
\def\sourceHiddenDim{d}

\def\targetSeqLen{m}
\def\targetPosition{i^*}
\def\targetPrompt{T}
\def\targetLayer{\ell^*}
\def\numTargetLayers{L^*}
\def\targetModel{\mathcal{M}^*}
\def\targetHidden{\bar{\vh}}
\def\targetHiddenDim{d^*}

\def\transformation{f}
\def\identity{\mathbb{I}}
\def\constantZero{0}

\def\subject{\sigma}
\def\relation{\rho}
\def\object{\omega}
\def\objects{\Omega}

\def\is{\leftarrow}

\usepackage{fontawesome}
\usepackage{multirow}
\usepackage{graphicx}

\usepackage{enumitem}
\usepackage{wrapfig}
\usepackage{soul}
\usepackage{comment}

\icmltitlerunning{\methodFamilyName{}: A Unifying Framework for Inspecting Hidden Representations of Language Models}

\begin{document}

\twocolumn[
\icmltitle{
\includegraphics[height=18pt]{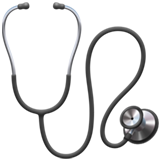}
\methodFamilyName: A Unifying Framework for Inspecting \\Hidden Representations of Language Models}



\icmlsetsymbol{equal}{*}

\begin{icmlauthorlist}
\icmlauthor{Asma Ghandeharioun}{equal,google}
\icmlauthor{Avi Caciularu}{equal,google}
\icmlauthor{Adam Pearce}{google}
\icmlauthor{Lucas Dixon}{google}
\icmlauthor{Mor Geva}{google,uni}
\end{icmlauthorlist}

\icmlaffiliation{uni}{Tel Aviv University}
\icmlaffiliation{google}{Google Research}

\icmlcorrespondingauthor{Asma Ghandeharioun}{aghandeharioun@google.com}
\icmlkeywords{Mechanistic Interpretability, Language Models, Patching, Representation, Probing}

\vskip 0.3in
]


\printAffiliationsAndNotice{\icmlEqualContribution} 

\begin{abstract}

Understanding the internal representations of large language models (LLMs) can help explain models' behavior and verify their alignment with human values. Given the capabilities of LLMs in generating human-understandable text, we propose leveraging the model itself to explain its internal representations in natural language. We introduce a framework called \methodFamilyName{} and show how it can be used to answer a wide range of questions about an LLM's computation. We show that many prior interpretability methods based on projecting representations into the vocabulary space and intervening on the LLM computation can be viewed as instances of this framework. Moreover, several of their shortcomings such as failure in inspecting early layers or lack of expressivity can be mitigated by \methodFamilyName{}. Beyond unifying prior inspection techniques, \methodFamilyName{} also opens up \textit{new} possibilities such as using a more capable model to explain the representations of a smaller model, and multihop reasoning error correction\footnote{Code is publicly available at \url{https://pair-code.github.io/interpretability/patchscopes}.}.

\end{abstract}

\section{Introduction}
\label{sec:introduction}

\begin{figure}[!ht]
    \centering
    \includegraphics[width=\columnwidth]{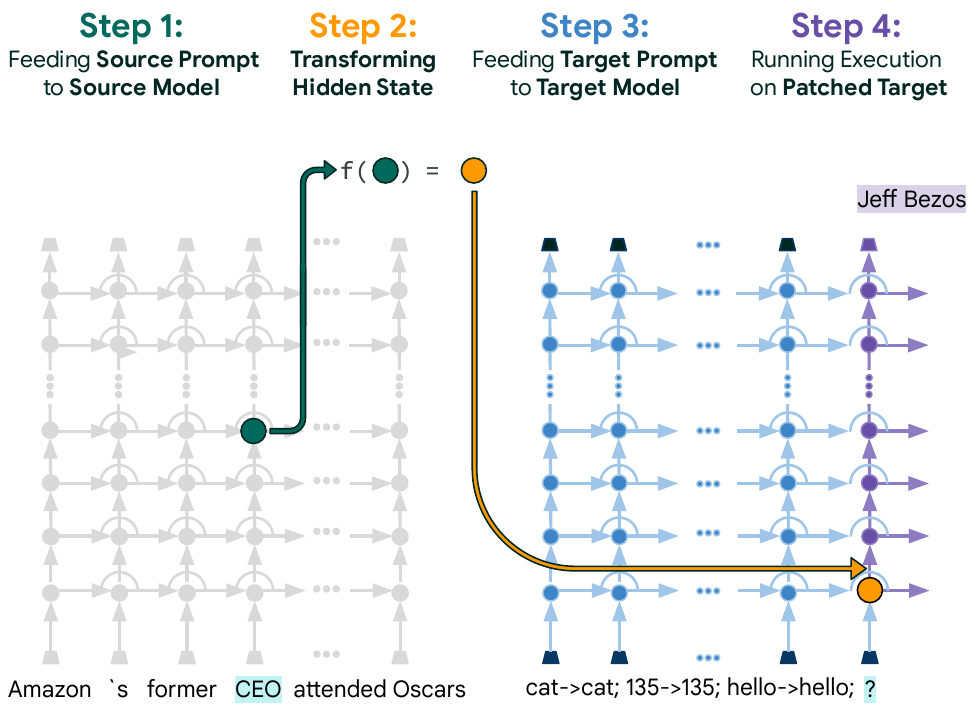}
    \vspace{-.5cm}
    \caption{Illustration of our framework, showing a \methodName{} for decoding 
    the hidden representation of \textit{``CEO''} in the source prompt (left).
    We use a target prompt (right) comprised of few-shot demonstrations of string repetitions to encourage the LLM to explain its internal representation.
    \textbf{Step~1}: Run the forward computation on the source prompt in the source model. \textbf{Step~2}: Optionally transform the hidden state at the source layer. \textbf{Step~3}: Run the forward computation on the target prompt up to the target layer in the target model. \textbf{Step~4}: Patch the target representation of the token \textit{``?''} at the target layer with the transformed representation (from step 2), then continue the forward computation from that layer onward. The modularity of \methodFamilyName{} allows designing a variety of methods by configuring the target prompt, model and transformation.
    }
    \label{fig:method}
\end{figure}

The question of what information is captured in the hidden representations of large language models (LLMs) is of key importance in control and understanding of modern generative AI, and has drawn substantial attention recently \cite{casper2022sok, madsen2022post, patel2021mapping, nanda-etal-2023-emergent}. To tackle this question, prior work has introduced a diverse array of interpretability methods, which largely rely on three prominent approaches: training linear classifiers, called probes, on top of hidden representations \cite{belinkov2019analysis, belinkov2022probing, alain2017understanding}, projecting representations to the model's vocabulary space \cite{logitlens,din2023jump,belrose2023eliciting}, and intervening on the computation to identify if a representation is critical for certain predictions \citep{meng2022locating,singh-etal-2020-bertnesia,wang2022interpretability, conmy2023automated,geva-etal-2023-dissecting}. 

Despite the wide success of these methods, they each exhibit practical shortcomings. First, probing relies on supervised training for pre-defined classes, which is hard to scale when there is a large number of classes or when all the categories are not known a priori. Second, the accuracy of vocabulary projections substantially decreases in early layers and the outputs are often hard to interpret. Last, all the above methods are not as expressive as one might like: they provide class probabilities or most likely tokens, as opposed to a high-quality explanation in natural language.

In this work, we argue that the advanced capabilities of LLMs in generating human-like text can be leveraged for ``translating'' the information in their representations for humans. We introduce a modular framework, called \methodFamilyName{} (\S\ref{sec:method}), that can be configured to query various kinds of information from LLM representations.
\methodFamilyName{} decode specific information from a representation within an LLM by ``patching'' it into the inference pass on a different prompt that has been designed to encourage the extraction of that information.\footnote{While patching (or ``activation patching'') by itself is not a new technique, to the best of our knowledge, we are the first to propose using it for decoding information from hidden representations in a configurable and expressive manner. See more details in \S\ref{sec:encompass_prior_methods}.} Such configuration (a \methodName{}) can be viewed as an inspection tool geared towards a particular objective, as illustrated in Fig.~\ref{fig:method}.

\newlength\lengthone
\setlength\lengthone{5em}

\newlength\lengthtwo
\setlength\lengthtwo{19em}

\newlength\lengththree
\setlength\lengththree{5em}

\newlength\lengthfour
\setlength\lengthfour{6.2em}

\newlength\lengthfive
\setlength\lengthfive{3em}

\begin{table}[!t]
\setlength\tabcolsep{3pt}
\centering
\caption{
Many prior inspection methods with various objectives can be viewed as \methodNamePlural{}. The rows highlighted in green show \methodName{} configurations that overcome several limitations of prior methods through more expressive inspection that is training-data free and is more robust across layers. 
}
\vspace{0.3em}
\label{tab:intro}
\resizebox{1\columnwidth}{!}{%
\renewcommand{\arraystretch}{1.2}
\begin{tabular}{m{\lengthone}m{\lengthtwo}m{\lengththree}m{\lengthfour}m{\lengthfive}}
\toprule
\textbf{Inspection Objective} & &
  \textbf{Expressive} &
  \textbf{Training \newline Data \newline Free} &
  \textbf{Robust Across Layers}\\ 
 \toprule
  \multirow{6}{*}{\parbox{\lengthone}{\textbf{Inspecting \newline Output \newline Distribution}}} &
  \parbox{\lengthtwo}{\cellcolor[HTML]{DFF6CC} Few-Shot Token Identity \methodName{} (\S\ref{sec:ntp_decoding})} &
  \cellcolor[HTML]{DFF6CC} \color[HTML]{34A853} {\faCheck \faCheck} &
  \cellcolor[HTML]{DFF6CC} \color[HTML]{34A853} {\faCheck} &
  \cellcolor[HTML]{DFF6CC} \color[HTML]{34A853}{\faCheck \faCheck} \\
 &
  \parbox{\lengthtwo}{Logit Lens \cite{logitlens}, \newline Embedding Space Analysis \cite{dar2023analyzing}} &
   \color[HTML]{34A853}{\faCheck} &
   \color[HTML]{34A853}{\faCheck} &
   \color[HTML]{ff0000}{\faClose} \\
 &
  \parbox{\lengthtwo}{Tuned Lens \cite{belrose2023eliciting}} &
   \color[HTML]{34A853}{\faCheck} &
  {For learning \newline mappings} &
   \color[HTML]{34A853}{\faCheck} \\
  &
  \parbox{\lengthtwo}{Future Lens \cite{pal2023future}} &
   \color[HTML]{34A853}{\faCheck} &
  {For learning \newline mappings} &
   \color[HTML]{34A853}{\faCheck \faCheck} \\ \hline
   
  \multirow{5}{*}{\parbox{\lengthone}{\textbf{Feature \newline Extraction}}} &
  \parbox{\lengthtwo}{\cellcolor[HTML]{DFF6CC} Zero-Shot Feat. Ext. \methodName{} (\S\ref{sec:feature_extraction})} &
  \cellcolor[HTML]{DFF6CC} \color[HTML]{34A853}{\faCheck \faCheck} &
  \cellcolor[HTML]{DFF6CC} \color[HTML]{34A853}{\faCheck} &
  \cellcolor[HTML]{DFF6CC} \color[HTML]{34A853}{\faCheck \faCheck} \\
 &
  \parbox{\lengthtwo}{LRE Attribute Lens \cite{hernandez2023linearity}} &
  \color[HTML]{34A853}{\faCheck}  &
  \multicolumn{1}{l}{\parbox{\lengthfour}{For linear \newline relation approx.}} &
  \color[HTML]{34A853}{\faCheck \faCheck}\\
  & 
 \parbox{\lengthtwo}{Probing \citep[e.g.,][]{belinkov2019analysis, belinkov2022probing, alain2017understanding, wang2023gaussian}} & \color[HTML]{ff0000}{\faClose}  & 
 {For training \newline probe} &
 \color[HTML]{34A853}{\faCheck} \\ \hline
 
  \multirow{5.5}{*}{\parbox{\lengthone}{\textbf{Entity \newline Resolution}}} &
  \parbox{\lengthtwo}{\cellcolor[HTML]{DFF6CC} Entity Description \methodName{} (\S\ref{sec:input_processing})} &
  \cellcolor[HTML]{DFF6CC} \color[HTML]{34A853}{\faCheck \faCheck} &
  \cellcolor[HTML]{DFF6CC} \color[HTML]{34A853}{\faCheck} &
  \cellcolor[HTML]{DFF6CC} \color[HTML]{34A853}{\faCheck \faCheck} \\
  
  & \parbox{\lengthtwo}{\cellcolor[HTML]{DFF6CC} X-Model Entity Desc. \methodName{} (\S\ref{sec:cross_model})} &
  \cellcolor[HTML]{DFF6CC} \color[HTML]{34A853}{\faCheck \faCheck \faCheck} &
  \cellcolor[HTML]{DFF6CC}{For learning \newline mappings} &
  \cellcolor[HTML]{DFF6CC} \color[HTML]{34A853}{\faCheck \faCheck} \\
  
  &
  \parbox{\lengthtwo}{Causal Tracing \cite{meng2022locating}} &
   \color[HTML]{ff0000}{\faClose} &
   \color[HTML]{34A853}{\faCheck} &
   \color[HTML]{34A853}{\faCheck \faCheck} \\
  &
  \parbox{\lengthtwo}{Attention Knockout \cite{wang2022interpretability, conmy2023automated, geva-etal-2023-dissecting}} &
   \color[HTML]{ff0000}{\faClose} &
   \color[HTML]{34A853}{\faCheck} &
   \color[HTML]{34A853}{\faCheck \faCheck} \\
  \midrule \midrule
  
 \multirow{2.5}{*}{{\parbox{\lengthone}{\textbf{Inspection \newline Application}}}} &
  \parbox{\lengthtwo}{Early Exiting, e.g., \newline Linear Shortcuts \cite{din2023jump}} &
  \color[HTML]{34A853}{\faCheck} &
   {For learning \newline mappings} &
  \color[HTML]{34A853}{\faCheck} \\
  
  & 
  \parbox{\lengthtwo}{Caption Generation, e.g., \newline Linear Mapping \cite{merullo2022linearly}} & 
  \color[HTML]{34A853}{\faCheck} &
  {For learning \newline mappings} & 
  \color[HTML]{34A853}{\faCheck} \\ 
 \bottomrule
\end{tabular}
}
\vspace{-1em}
\end{table}

We show that many existing methods, including those that rely on vocabulary projections and computation interventions, can be cast as \methodNamePlural{}. Moreover, new configurations of our framework introduce more effective tools in addressing the same questions, while mitigating several limitations of prior approaches. Also, \methodFamilyName{} enables addressing underexplored questions, such as fine-grained analysis of the input contextualization process and the extent to which a more expressive model can be used to inspect hidden representations of a smaller model.

We conduct a series of experiments to evaluate the benefits and opportunities introduced by \methodFamilyName{}, focusing on auto-regressive LLMs. First, we consider the problem of estimating the model's next-token prediction from its intermediate representations (see \S\ref{sec:ntp_decoding}). Across multiple LLMs, we show that using a few-shot token identity prompt, a prompt in the form of $``\texttt{tok}_1 \rightarrow \texttt{tok}_1 \texttt{; tok}_{2} \rightarrow \texttt{tok}_2 ; \ldots; \texttt{tok}_{k}"$ where $\texttt{tok}_{i}$ refers to a random token, leads to substantial gains over vocabulary projection methods. Next, we evaluate how well \methodFamilyName{} can decode specific attributes of an entity from its LLM representations, when these are detached from the original context (see \S\ref{sec:feature_extraction}). We observe that, despite using no training data, \methodFamilyName{} significantly outperforms probing in six out of twelve commonsense and factual reasoning tasks, and works comparably well in all but one of the remaining six.

Beyond output estimation and attribute decoding, \methodFamilyName{} can address questions that are hard to answer with existing methods. 
In \S\ref{sec:input_processing}, we apply \methodFamilyName{} to study how LLMs contextualize input entity names in early layers, where vocabulary projections mostly fail and other methods, at best, provide only a binary signal of whether the entity has been resolved~\cite{youssef2023give, tenney2019bert}. With a new \methodName{}, we are able to \textit{verbalize} the gradual entity resolution process.
For example, we show that, as the model processes the final token of \textit{``Alexander the Great''} throughout the layers, it reflects different entities starting from \textit{``Great Britain''}, to \textit{``the Great Depression''}, to finally resolving \textit{``Alexander the Great"}. Then, in \S\ref{sec:cross_model} we show how one can further improve \methodName{} expressivity by using a stronger target model, e.g., Vicuna 13B instead of Vicuna 7B.

Lastly, we showcase the utility of \methodFamilyName{} for fixing latent multi-hop reasoning errors, particularly when the model is capable of conducting each reasoning step correctly, but fails when they need to be composed in-context (\S\ref{sec:multihop}). Building on top of the data provided by \citet{hernandez2023linearity}, we introduce a more complex task that requires two steps of factual reasoning. \methodName{} achieves 50\% accuracy on this task, outperforming chain-of-thought \cite{wei2022chain} (35.71\%) and vanilla generations (19.57\%).

To conclude, our work makes the following contributions: We propose \methodFamilyName{}, a general modular framework for decoding information from the hidden representations in LLMs. We show that prominent interpretability methods can be viewed as instances of \methodFamilyName{}, and new configurations result in more expressive, robust across layers, and training-data free alternatives that mitigate their shortcomings.  In addition, novel configurations introduce unexplored possibilities of stronger inspection techniques, as well as practical benefits, such as correcting multi-hop reasoning errors.

\section{Related Work}
\label{sec:related_work}
Activation patching is a causal intervention, commonly used as a tool for studying if certain activations play a key role in a model's computation \cite{geiger2021causal, vig2020investigating}. 
Patching has been used largely for localizing specific information to specific layers and token positions \citep{goldowsky2023localizing, meng2022locating, meng2022mass, stolfo-etal-2023-mechanistic, merullo2023mechanism}, and for finding information propagation paths in the computation \cite{wang2022interpretability, geva-etal-2023-dissecting, hendel2023context, hanna2023how, lieberum2023does}. Prior works have also used specific forms of cross-model patching called stitching, in non-transformer architectures, mostly to analyze representational similarity \citep[e.g.,][]{bansal2021revisiting, csiszarik2021similarity, lenc2015understanding}. Despite certain limitations \cite{hase2023does, zhang2023towards}, patching remains a principal tool for mechanistic interpretability \cite{conmy2023automated}.

Given promising results from emerging interpretability efforts that employ LLMs to generate human-like text for inspection \citep[e.g.,][]{mousi2023llms, slobodkin-etal-2023-curious,bills2023language}, we argue that using patching only for localization purposes is myopic, and propose to use it for ``translating'' LLM representations into natural language. Very recently, patching has been used to study new problem setups \citep[e.g.,][]{pal2023future, hernandez2023linearity}, all of which can be seen as different configurations of our proposed framework (see \S\ref{sec:encompass_prior_methods}).

Among the growing research efforts in inspecting hidden representations of neural networks, probing classifiers are perhaps the most common \citep[e.g.,][]{alain2017understanding,belinkov2019analysis,belinkov2022probing, wang2023gaussian}, and methods using projections into the vocabulary space or their extensions to other domains are another key category \citep[e.g.,][]{merullo2023mechanism, geva-etal-2022-transformer,logitlens, belrose2023eliciting, dar2023analyzing, din2023jump, langedijk2023decoderlens, vilas2023analyzing}. While various other latent inspection methods exist \citep[e.g.,][]{zhou2018interpreting, strobelt2017lstmvis, ghandeharioun2021dissect, kim2018interpretability}, the above are the most relevant to this work.

\section{\methodFamilyName{}}
\label{sec:method}

In this section, we introduce \methodFamilyName{} and show how it extends prior interpretability methods with new capabilities. While not limited to particular LLM architectures, this work focuses on auto-regressive transformer-based LLMs.

\subsection{Framework Description}
\label{sec:framework}

The key idea in \methodFamilyName{} is to leverage the advanced capabilities of LLMs to generate human-like text for ``translating'' the information encoded in their own hidden representations. Concretely, given a hidden representation obtained from an LLM inference pass, we propose to decode specific information from it by \textit{patching} it into a different inference pass (of the same or a different LLM) that encourages the translation of that specific information.

Notably, the rest of the forward computation after patching can augment the representation with additional information, hence, this approach does not guarantee that the patched representation itself stores \textit{all} that information. However, dispatching the representation from its original context (the source prompt) stops contextualization and guarantees that no further information from the source prompt is incorporated in the post-patching computation. Thus, our framework reveals if specific information \textit{can be decoded from the patched representation via the post-patching computation}.

Given an input sequence of $\sourceSeqLen$ tokens $\sourcePrompt = \langle s_1, ..., s_{\sourceSeqLen} \rangle$ and a model $\sourceModel$ with $\numSourceLayers$ layers, $\sourceHidden_{\sourcePosition}^{\sourceLayer}$ denotes the hidden representation obtained at layer $\sourceLayer \in [1, \ldots, \numSourceLayers]$ and position $\sourcePosition \in [1, \ldots, \sourceSeqLen]$, when running $\sourceModel$ on $\sourcePrompt$. To inspect $\sourceHidden_{\sourcePosition}^{\sourceLayer}$, we consider a separate inference pass of a model $\targetModel$ with $\numTargetLayers$ layers on a target sequence $\targetPrompt = \langle t_1, \ldots, t_{\targetSeqLen} \rangle$ of $\targetSeqLen$ tokens. Specifically, we choose a hidden representation $\targetHidden_{\targetPosition}^{\targetLayer}$ at layer $\targetLayer \in [1, \ldots, \numTargetLayers]$ and position ${\targetPosition} \in [1, \ldots, \targetSeqLen]$ in the execution of $\targetModel$ on $\targetPrompt$. Moreover, we define a mapping function $\transformation({\sourceHidden}; \vtheta): \mathbb{R}^{\sourceHiddenDim} \mapsto \mathbb{R}^{\targetHiddenDim}$ parameterized by $\vtheta$ that operates on hidden representations of $\sourceModel$, where $\sourceHiddenDim$ and $\targetHiddenDim$ denote the hidden dimension of representations in $\sourceModel$ and $\targetModel$, respectively. This function can be the identity function, a linear or affine function learned on task-specific pairs of representations, or even more complex functions that incorporate other sources of data.
The \textit{patching} operation refers to dynamically replacing the representation $\targetHidden_{\targetPosition}^{\targetLayer}$ during the inference of $\targetModel$ on $\targetPrompt$ with $\transformation(\sourceHidden_{\sourcePosition}^{\sourceLayer})$.
Namely, by applying ${\targetHidden}_{\targetPosition}^{\targetLayer} \is \transformation(\sourceHidden_{\sourcePosition}^{\sourceLayer})$, we intervene on the generation process and modify the computation after layer $\targetLayer$.

Overall, a $\methodName$ intervention applied to a representation determined by $(\sourcePrompt,\sourcePosition,\sourceModel,\sourceLayer)$, 
is defined by a quintuplet $(\targetPrompt,\targetPosition,\transformation,\targetModel,\targetLayer)$ of a target prompt $\targetPrompt$, a target position $\targetPosition$ in this prompt, a mapping function $\transformation$, a target model $\targetModel$, and a target layer $\targetLayer$ of this model.
It is possible that $\sourceModel$ and $\targetModel$ are the same model, $\sourcePrompt$ and $\targetPrompt$ are the same prompt, and $\transformation$ is the identity function $\identity$ (i.e., $\identity(\sourceHidden) = \sourceHidden$).
Next, we show how this formulation covers prior interpretability methods and further extends them with new capabilities.

\subsection{\methodFamilyName{} Encompasses Prior Methods}
\label{sec:encompass_prior_methods}

Many recent methods inspect LLM representations by projecting them to the output vocabulary space \cite{logitlens,din2023jump,belrose2023eliciting}. 
Formally, an estimation of the output distribution is obtained from the representation $\vh_i^\ell$ at position $i$ and layer $\ell$ by:
$$\vp_i^\ell = \softmax(W_Uf(\vh_i^\ell)) \in \mathbb{R}^{|V|} ,$$ 
where $W_U \in \mathbb{R}^{|V|\times d}$ is the model's unembedding matrix and $f$ is a simple mapping function, such as the identity function or an affine mapping. We note that the operation applied to $f(\vh_i^\ell)$ is the same computation applied by the model to the last-layer representation for obtaining the next-token prediction. Therefore, prior methods that inspect representations in the vocabulary space can be viewed as a class of \methodFamilyName{} that maps representations from any source layer $\sourceLayer$ to the last target layer $\numTargetLayers$. Differences between these methods lie in the choice of $\transformation$; logit lens \cite{logitlens, dar2023analyzing} applies the identity function, linear shortcuts \cite{din2023jump} uses a linear mapping function, and tuned lens \cite{belrose2023eliciting} trains an affine mapping. Recently, \citet{hernandez2023linearity} introduced LRE Attribute Lens that builds $\transformation$ based on a relation linearity assumption, and showed its effectiveness in attribute extraction.

This class of methods has proven to be effective for different applications, for example, in improving inference efficiency via early exiting \cite{din2023jump}. While the majority of methods and applications in this category use a single model ($\targetModel = \sourceModel$), \citet{merullo2022linearly} had demonstrated successful caption generation with a generative image model as $\sourceModel$ and a language model as $\targetModel$. 

Another category of inspection methods intervene on the LLM computation. Contemporary to our work, \citet{pal2023future} have investigated whether it is possible to anticipate multiple generated tokens ahead from a given hidden representation, rather than estimating just the next-token prediction. Their method (Future Lens) uses a target prompt different from the original prompt (i.e., $T \neq S$) and is designed to decode subsequent tokens from information encoded in a hidden representation $\vh_i^\ell$. Example target prompts are \nl{The multi-tokens present here are } and \nl{Hello! Could you please tell me more about }.
Future Lens can be cast as another \methodName{} with $\targetModel = \sourceModel$ and $\targetLayer = \sourceLayer$.

More broadly, \methodFamilyName{} also cover recent mechanistic interpretability methods that analyze internal processes in LLMs with 
inference computation
interventions. Specifically, causal tracing \cite{meng2022locating} uses a source prompt augmented with Gaussian noise as the target prompt. Other previous methods have intervened on one or more target layers during inference by patching zero vectors to the computation \cite{wang2022interpretability, conmy2023automated, geva-etal-2023-dissecting}, namely, setting $\transformation(\sourceHidden) = \mathbf{0}$. For a configuration summary of how these interpretability methods can be cast as \methodName{} instances, see \S\ref{sec:method-configs}, Tab. \ref{tab:method-configs}.

\subsection{\methodFamilyName{} Enables Novel Inspection Methods}
\label{sec:novel-methods}

Prior work has utilized specific patching configurations for interpretability, largely patching the same model while using the same prompt (i.e., $\mathcal{M}^* = \mathcal{M}$, $T=S$).
The framing of \methodFamilyName{} introduces a wide range of unexplored setups potentially unlocking new inspection capabilities.

Specifically, we observe that modifying the target prompt enables an expressive decoding of a wide range of features, detached from the source prompt computation. For instance, we can use the prompt \textit{``The capital of X is''} to check if the capital city of a given country is extractable from its (last token) hidden representation at a specific layer.
Similarly, a prompt like \textit{``Tell me facts about X''} can be leveraged to assess whether the model has resolved the entity name in a specific layer.
Contrary to probing, this approach is not restricted by the number of classes of the chosen feature.

Moreover, when the inspected model is not expressive enough to answer certain queries, patching representations into a more capable model could be useful  \citep{hernandez2022natural,singh2023explaining,schwettmann2023multimodal}.

\section{Experiments}
\label{sec:experiments}

In this section, we evaluate our framework on decoding next-token predictions (\S\ref{sec:ntp_decoding}), extracting attributes (\S\ref{sec:feature_extraction}), analyzing the contextualization of entity names (\S\ref{sec:input_processing}), and leveraging stronger models for inspection via cross-model patching (\S\ref{sec:cross_model}). See a summary in Tab.~\ref{tab:intro}.

\subsection{Decoding of Next-Token Predictions}
\label{sec:ntp_decoding}

As introduced in \S\ref{sec:encompass_prior_methods}, let $\vp^\numSourceLayers$ be the output probability distribution for some input, obtained by multiplying the final-layer last-position hidden representation $\sourceHidden^\numSourceLayers$ by the unembedding matrix $W_U\in \mathbb{R}^{|V|\times d}$. We wish to estimate $\vp^L$ from intermediate representations $\mathbf{h}^\ell$ s.t. $\ell < L$.
Particularly, we ask how early in the computation the model has concluded its final prediction from the given context.
In our experiments, we consider multiple LLMs -- LLaMA2 (13B) \cite{touvron2023llama}, Vicuna (13B) \cite{vicuna2023}, GPT-J (6B) \cite{gpt-j}, and Pythia (12B) \cite{pythia} (see more details in \S\ref{section:ntp_models}).

\begin{figure}[t]
    \centering
    \includegraphics[width=0.49\textwidth]{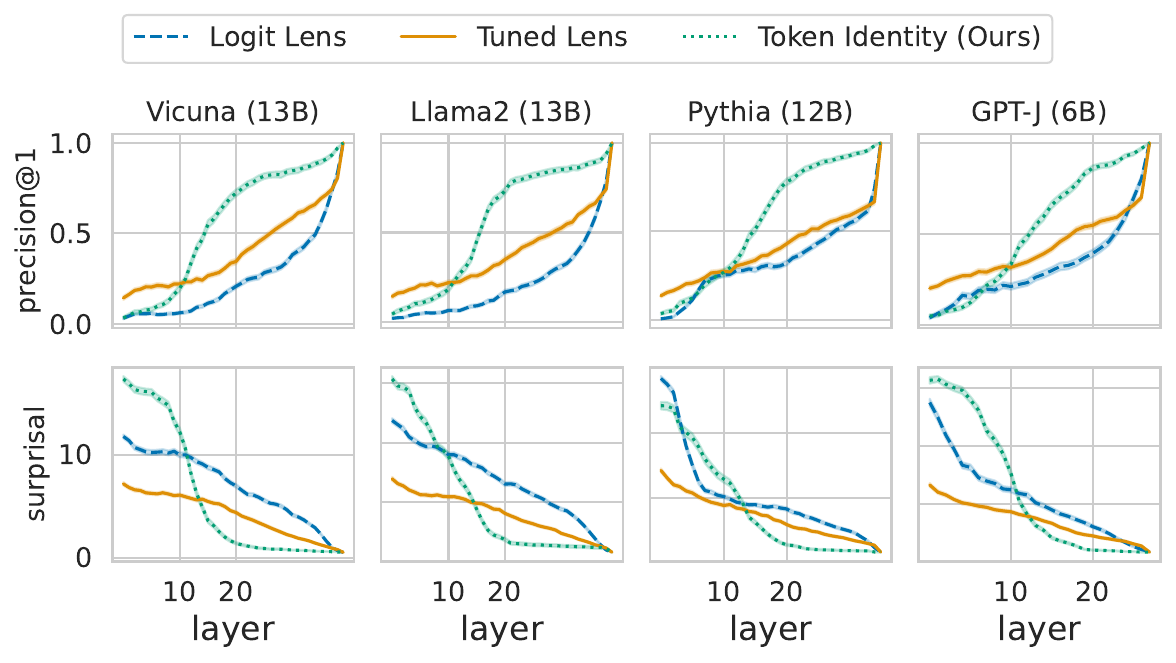}\vspace{-1em}
    \caption{Precision@1 ($\uparrow$ is better) and Surprisal ($\downarrow$ is better) of next-token prediction estimation in multiple models. From layer 10 and upwards, the token identity method (ours) consistently outperforms the rest of the baselines across all the models.}
    \label{fig:next_token}
\end{figure}

\paragraph{Methods} 
We compare vocabulary projection methods (\S\ref{sec:encompass_prior_methods}) with a new \methodName{}.
Each method yields an estimated output probability $\tilde{\vp}^\sourceLayer$ by patching an intermediate representation $\sourceHidden^\sourceLayer$ to the model's final layer. Here, we focus on the common setting where $\sourceModel = \targetModel$, and discuss extensions to $\sourceModel \neq \targetModel$ in \S\ref{sec:cross_model}.

\begin{itemize}[itemsep=1pt, topsep=0pt,leftmargin=*]
    \item \textbf{Logit Lens:} Following prior work \citep{logitlens, geva-etal-2022-lm}, we define $f$ as the identity function, meaning no change is applied to the patched representation. That is, $f(\vh) := \mathbb{I}(\vh)$.

    \item \textbf{Tuned Lens:} Motivated by \citet{belrose2023eliciting,din2023jump}, we employ an affine mapping function between representations at layer $\sourceLayer$ and the final layer $\numSourceLayers$.
    Specifically, we feed the model examples from a training set $\mathcal{T}$ and for each example $s\in\mathcal{T}$ obtain a pair  $(\vh_{s}^\ell, \vh_{s}^L)$ of hidden representations. Then, we fit a linear regression model to find the matrix $A^{\sourceLayer} \in \mathbb{R}^{d \times d}$ and the bias vector $\vb^{\sourceLayer}\in \mathbb{R}^{d}$ that are numerical minimizers for $\sum_{s \in \mathcal{T}} || A \vh_{s}^\ell - \vh_{s}^{L} + \vb ||^2$.
    We define $\transformation$ as:
        $f(\vh^\ell) \coloneqq A^{\ell} \vh^\ell + \vb^{\ell}$.
    
    \item \textbf{Token Identity} \methodName{}: Unlike the previous methods, here we use a target prompt that is different from the source prompt ($\targetPrompt \neq \sourcePrompt$) and is meant to encourage decoding the token identity of the hidden representation. Also, while the above methods skip the computation between layers $l$ and $L$, here we modify it such that all the information from the source prompt computation is discarded,
 except for the patched representation.
    We craft a prompt with $k$ demonstrations representing an identity-like function, formatted as ``\texttt{tok$_1$\,$\rightarrow$\,tok$_1$\,;\,tok$_2$\,$\rightarrow$\,tok$_2$\,;\,$\ldots$\,;\,tok$_k$}''. See \S\ref{section:add_token_id} for further details and an experiment showing this method's robustness to different demonstrations.
    Note that this \methodName{} does not require any training.
\end{itemize}

\paragraph{Evaluation} 
Following \citet{din2023jump}, we use Pile evaluation set and the following metrics (see details in \S\ref{section:ntp_data}):
\begin{itemize}[itemsep=1pt, topsep=0pt,leftmargin=*]
    \item \textbf{Precision@1} ($\uparrow$ is better): The portion of examples for which the highest-probability token $t$ in the estimated probability distribution matches the highest-probability token in the original output distribution. That is, if $\argmax_{t}(\tilde{\vp}^\sourceLayer_t) = \argmax_{t}(\vp^\numSourceLayers_t)$.
    \item \textbf{Surprisal} ($\downarrow$ is better): The minus log-probability of the highest-probability token in the predicted distribution $\tilde{\vp}^\ell$ according to $\vp^L$, i.e., $-\log \vp^L_{\tilde{t}}$, where $\tilde{t}=\argmax_t(\tilde{\vp}^\sourceLayer_t).$
\end{itemize} 

\paragraph{Results} Across all the models, from layer 10 and upwards, the token identity \methodName{} consistently outperforms the other baselines, obtaining a gain of up to 98\% in layers 18-22 (see Fig.~\ref{fig:next_token}). 
This demonstrates the utility of leveraging the model's decoding procedure for inspecting representations of different source prompts, and shows that \textit{in most cases} hidden representations in early layers carry the prediction information regardless of their context.

In the first 10 layers, performance of all methods is worsened, with the token identity prompt performing on-par with logit lens, and tuned lens performing slightly better, which could be due to the additional training of its mappings. Low performance in these layers is expected, as it is where the input contextualization happens. In \S\ref{sec:input_processing}, we introduce a \methodName{} geared towards unraveling this process.

\subsection{Extraction of Specific Attributes}
\label{sec:feature_extraction}

Classification probes are arguably the most commonly used method for checking if certain attributes are encoded in hidden representations \cite{belinkov2022probing, belinkov2019analysis}. However, they need to be trained, and the range of attribute classes needs to be known a priori. Here we show that repurposing \methodFamilyName{} for attribute extraction overcomes these limitations. First, it does not require training. Second, it is not limited by a predefined set of labels, but rather benefits from an open vocabulary. In addition, by taking advantage of the model's nonlinearities, it can capture more complex relations compared to linear probes.

\paragraph{Experimental Setup}
Consider factual and commonsense knowledge represented as triplets $(\subject, \relation, \object)$ of a subject (e.g., $``\texttt{United States}"$), a relation (e.g., $``\texttt{largest city of}"$), and an object (e.g., $``\texttt{New York City}"$).
We investigate to what extent the object $\object$ can be extracted from the last token representation of the subject $\subject$ in an arbitrary input context. 
To this end, we conduct experiments on 8 commonsense and 25 factual knowledge tasks curated by \citet{hernandez2023linearity}. This dataset includes $(\subject, \relation, \object)$ triplets for different relations, along with prompt templates that verbalize them in natural language. 
We conduct experiments with GPT-J (6B) \citep{gpt-j}, filtering the data to keep only the examples where $\object$ appears in the the model's continuation of the prompt up to 20 tokens.
The choice of 20 balances computation cost with accommodating the open-ended nature of Patchscopes, as the ground truth token does not necessarily appear in the next immediate token in a fluent response.
For each example, we sample 5 utterances from the WikiText-103 dataset \cite{merity2016pointer} that include $\subject$ and use them as $\sourcePrompt$. Lastly, we keep tasks with at least 15 samples, which results in 5 commonsense and 7 factual tasks with a total of 1,453 datapoints.
See details in \S{\ref{sec:appendix-attribute-extraction}}.

\begin{table}[t!]
    \centering
    \caption{Feature extraction accuracy (mean$\pm$std). Comparing zero-shot feature extraction \methodName{} to a logistic regression probe shows that despite using \textit{no training data}, it has a significantly higher accuracy than baseline in 6 out of 12 tasks. We use pairwise t-test with Bonferroni correction for comparing the two methods. $^{**}$ and $^*$ indicate $p\!<1e\!-\!5$ and $p\!<\!1e\!-\!4$, respectively.}
    \vspace{1em}
    \label{tab:attribute_extraction}
    \resizebox{.97\columnwidth}{!}{%
    \bgroup
    \def\arraystretch{1.1}
    \begin{tabular}{m{.1em}m{7.3em}m{6em}m{6em}}
        \toprule
         \multicolumn{2}{c}{\centering \textbf{Task}}  & {\textbf{Probe}} & { \textbf{\methodName}} \\
        \toprule
        \multirow{5}{.1em}{\rotatebox{90}{Commonsense}} & Fruit inside color & $37.4 \pm 6.6$ & $38.0 \pm 18.7$ \\
         & Fruit outside color &  $35.5 \pm 3.1$ & $\mathbf{71.0 \pm 13.3^{**}}$ \\
         & Object superclass & $\mathbf{65.6 \pm 10.5^{*}}$ & $54.8 \pm 11.3$ \\
         & Substance phase & $73.8 \pm 3.7$ & $\mathbf{91.9 \pm 1.7^{**}}$ \\
         & Task done by tool & $10.1 \pm 3.2$ & $\mathbf{48.1 \pm 13.2^{**}}$ \\
        \midrule
        \multirow{7}{.1em}{\rotatebox{90}{Factual}} & Company CEO  & $5.0 \pm 2.6$ & $\mathbf{47.8 \pm 13.9^{**}}$ \\
         & Country currency & $17.7 \pm 2.2$ & $\mathbf{51.0 \pm 8.9^{**}}$ \\
         & Food from country & $5.1 \pm 3.7$ & $\mathbf{63.8 \pm 11.3^{**}}$ \\
         & {Plays pos. in sport} & {\centering $75.9 \pm 9.1$} & $72.2 \pm 7.2$ \\
         & {Plays pro sport} & $53.8 \pm 10.3$ & $46.3 \pm 14.2$ \\
         & {Product by co.} & $58.9 \pm 7.2$  & $63.2 \pm 10.7$ \\
         & Star constellation & $17.5 \pm 5.3$ & $18.4 \pm 5.1$ \\
        \bottomrule
    \end{tabular}
    \egroup
    }
\end{table}

\paragraph{Methods} 
We compare our proposed \methodName{} against linear probing \cite{kohn2015s, gupta2015distributional}.

\begin{itemize}[itemsep=1pt, topsep=0pt,leftmargin=*]
    \item \textbf{Zero-shot Feature Extraction} \methodName{}: 
    We craft $\targetPrompt$ as a general verbalization of $\relation$ followed by a placeholder for $\subject$, such that $\targetPosition = \targetSeqLen$. For example, we use $\targetPrompt \is $ ``\texttt{The largest city in x}'' with ``\texttt{x}'' as a placeholder for the subject. To extract the object from the entity representation in $\sourcePrompt$, we patch the representation of token ``\texttt{x}'' at layer $\targetLayer$ with the representation of ``\texttt{States}'' from layer $\sourceLayer$, and consider if
 the generated text includes $\object$. 
    The remaining configurations of this \methodName{} are $\transformation \is \identity, \targetModel \is \sourceModel, \sourcePosition \is $ the last token of $\subject$ in $\sourcePrompt$. 
    We consider all combinations of $\sourceLayer \in [1, \ldots, \numSourceLayers] \times \targetLayer \in [1, \ldots, \numTargetLayers]$. Later in this section, we discuss the role of $\sourceLayer$ pertaining to attribute extraction. 
    \item \textbf{Logistic Regression Probe}: Let $\objects$ represent the range of possible objects for a given relation. We use the set of unique values of $\object$ in the training set as a proxy for $\objects$. We train a logistic regression probe \cite{kohn2015s, gupta2015distributional} for each layer that predicts $\object \in \objects$ from the last token representation of $\subject$. Given that 6 out of 12 tasks have fewer than 40 datapoints, we use three-fold cross-validation for training and evaluation of this baseline. Note that we have excluded tasks where the probe fails completely due to insufficient number of training examples (fewer than 15 datapoints).
\end{itemize}

\paragraph{Evaluation} We measure the average attribute extraction accuracy. For a given sample, the \methodName{} is considered correct if $ \exists \, \targetLayer \in [1, \ldots, \numTargetLayers]$ where the generated text up to 20 tokens includes $\object$. For the probe, a prediction is correct if the highest probability is assigned to $\object$.

\paragraph{Results} Tab.~\ref{tab:attribute_extraction} summarizes the results, averaged over $\sourceLayer \in [1, \ldots, \numSourceLayers]$. We conduct a T-test with Bonferroni correction to compare the two methods. Despite using no training data and having no restrictions on the output, the \methodName{} achieves a significantly higher accuracy than the probe on six out of twelve tasks $(p\!<\!1e\!-\!5)$, and works comparably well in all but one of the remaining six. These results suggest that, in the majority of cases, the source representation without its original context carries enough information about many attributes that a targeted \methodName{} can extract.
We also study how the accuracy changes across the source layers, and observe that \methodName{} consistently outperforms the baseline in early layers, outperforms or works on par with the baseline in mid layers, and almost all cases where it performs worse than the baseline occur in later layers. Our interpretation is that given the language modeling training objective, the representations shift toward next-token prediction in the later layers. Therefore, the attribute of interest would not be as readily accessible via the model's computation in these layers. This interpretation is also aligned with recent findings that show no decline in using linear relational embedding in predicting $\object$ only when the next token also happens to be $\object$ \cite{hernandez2023linearity}. Note that this pattern explains the higher standard deviation of \methodName{} accuracy observed in Tab.~\ref{tab:attribute_extraction}. We discuss this phenomenon in more detail in \S\ref{sec:appendix-attribute-extraction} (see Fig.~\ref{fig:attribute_extraction_lineplots_appendix}).

\subsection{Analyzing Entity Resolution in Early Layers}
\label{sec:input_processing}

The previous sections focused on analyzing the information encoded in a single hidden state. Here we turn to consider a more global question of how LLMs resolve entity mentions across multiple layers. Concretely, given a subject entity name, such as \textit{``the summer Olympics of 1996''}, how does the model contextualize the input tokens of the entity and at which layer is it fully resolved?

Answering these questions is hard with existing methods; vocabulary projections focus on the output prediction and fail to show clear patterns in early layers, and probing is restricted to outputs from a fixed number of classes, which may not be expressive enough to describe this process. 
Alternative approaches have studied this process indirectly via interventions \cite{meng2022locating}, showing that the model constructs a subject representation at the last token of the entity name. However, it is still unclear how this contextualization is performed. 

We analyze how LLMs contextualize input entity names by leveraging \methodFamilyName{}.
Particularly, we craft a target prompt for generating a description of a given subject, and apply it to the hidden representation at the last subject position in the source prompt -- where the model forms the subject representation \cite{geva-etal-2023-dissecting, hernandez2023measuring} -- across the early layers. This will allow us to see how the model describes the subject in each layer.

\begin{table}[t!]
\caption{Illustrating entity resolution via a qualitative example. The expressive generations show that as we go through the layers, more tokens from the context get integrated into the current representation. The ``Generation'' column shows the automatically generated text. The ``Explanation'' column shows our own manually coded interpretation, aiming to specify what entity the generation refers to and how that relates to the tokens processed. $\targetModel=\sourceModel \is \text{Vicuna (13B)}$, $\targetLayer = \sourceLayer$, $\sourcePrompt \is \texttt{"Diana,\,Princess\;of\;Wales"}$.}
\vspace{.5em}
\label{tab:entity-resolution-samples}
\def\arraystretch{1.85}
\resizebox{1.\columnwidth}{!}{%
\begin{tabular}{m{1.35em}m{19em}m{7.5em}}
    \toprule
    \boldmath$\sourceLayer$  & \textbf{Generation}    & \textbf{Explanation} \\
    \toprule
    1-2 &
    : Country in the United Kingdom &
    \textbf{Wales} \\
    3 &
    : Country in Europe &
    \textbf{Wales} \\
    4 &
    : Title held by female sovereigns in their own right or by queens consort &
    \textbf{Princess of Wales} \newline (unspecific) \\
    5 &
    : Title given to the wife of the Prince of Wales (and later King) &
    \textbf{Princess of Wales} \newline (unspecific)\\
    6 &
    : Diana, Princess of Wales (1961-1997), the first wife of Prince Charles, Prince of Wales, who was famous for her beauty and humanitarian work &
    \textbf{Diana, \newline Princess of Wales} \\
  \bottomrule
\end{tabular}%
}
\vspace{-0.5em}
\end{table}

\paragraph{Analysis Setting}
We use a few-shot target prompt template for decoding an entity description: ``\texttt{subject$_1$:\;description$_1$,\;\ldots, subject$_k$: description$_k$, x}'', while patching the last position corresponding to \texttt{x}. 
We take the 200 most popular and 200 least popular subject entities from the PopQA dataset \cite{mallen-etal-2023-trust}. The popular entities should appear frequently in LLMs' pre-training data, and are thus likely to be captured by the model, while resolving the rare entities is expected to be more challenging \citep{kandpal2023large,mallen-etal-2023-trust}.
Then, for the source prompt we use the entity name, and for the target prompt we sample $k=3$ random subject entities. We obtain a short (up to one sentence) description of every subject entity from Wikipedia. Our target prompt and more technical details are provided in \S\ref{sec:appendix-entity-resolution-info}.
We patch the last position representations from the first 10 layers of Vicuna 13B to the target prompt and evaluate the generated subject name and description. Specifically, the generated descriptions are evaluated against the descriptions from Wikipedia using RougeL \cite{lin-2004-rouge}. Evaluation with Rouge1 \cite{lin-2004-rouge} and Sentence-Bert \citep{reimers-2019-sentence-bert} shows similar trends (see \S\ref{sec:appendix-entity-resolution}).

\paragraph{Results}

Tab.~\ref{tab:entity-resolution-samples} illustrates the generations by Vicuna 13B for a sample subject entity, when patching its representation at different layers to the target prompt (see more examples in~\S\ref{sec:appendix-entity-resolution-qualitative}). For most entities, the contextualization process is spread over the first layers, with the last subject token gradually encompassing more distant positions across layers.

\begin{figure}[t]
    \centering
    \includegraphics[width=0.99\columnwidth]{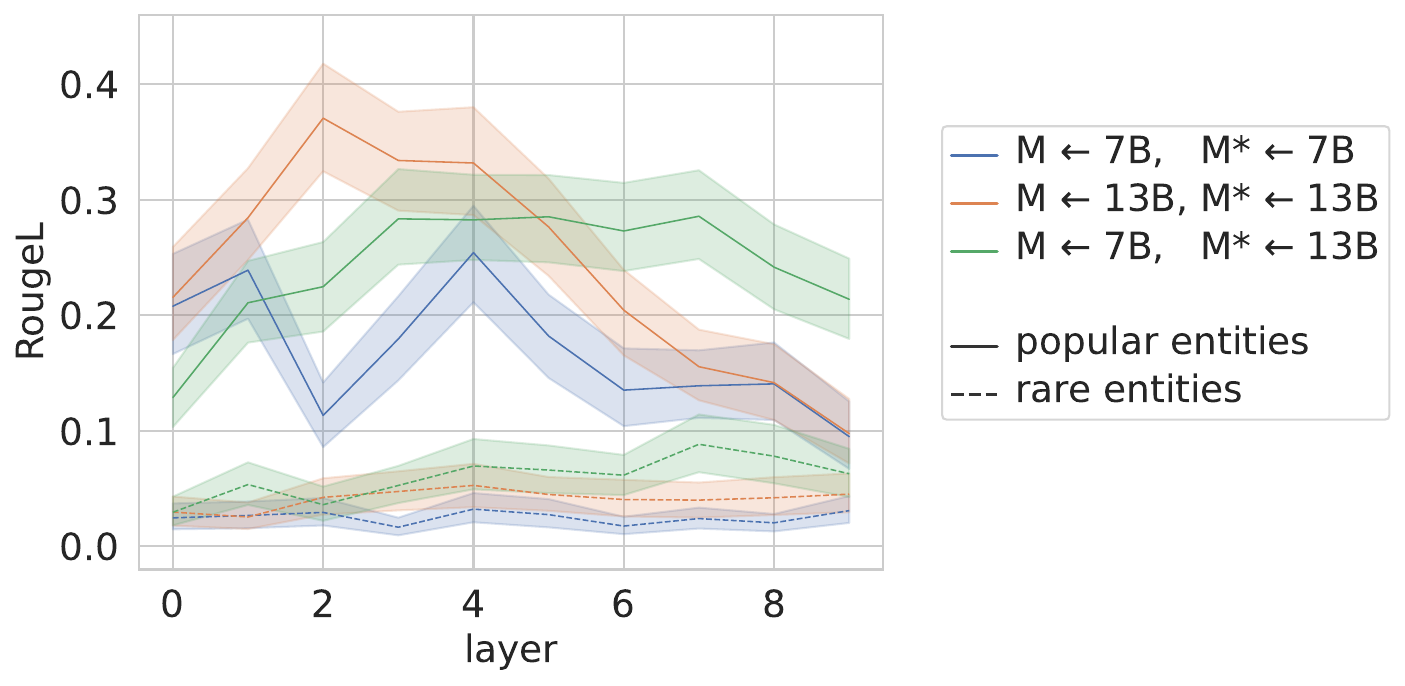}\vspace{-1em}
    \caption{RougeL scores of the generated descriptions against descriptions from Wikipedia, using Vicuna models.}
    \label{fig:entity_processing}
\end{figure}

This trend can be quantitatively observed by the similarity between the generated descriptions and the descriptions from Wikipedia, as measured by RougeL. See Fig.~\ref{fig:entity_processing} where $\sourceModel = \targetModel$. For both models, similarity increases in the first 5 layers and then slowly decreases. This decrease could potentially be attributed to contamination caused by the representation of the placeholder token ``\texttt{x}'' remaining in the early layers, when patching is applied to a later layer. Note that this potential issue is only applicable to multi-token generation scenarios as future positions can still attend to the placeholder position in early layers, potentially interfering with the model's ability to accurately generate descriptions for the patched token. See \S\ref{sec:appendix-entity-resolution-qualitative} for qualitative examples corroborating this interpretation.
As expected, the scores for rare, long-tail entities are significantly lower than those of popular entities. See \S\ref{sec:appendix-entity-resolution} for additional results with Pythia where the smaller model seems to outperform the larger model, possibly because the larger model is biased toward output generation at the expense of input contextualization. To summarize, this analysis shows \methodFamilyName{}' utility for inspecting the contextualization process in early layers.

\begin{figure*}[h!]
    \centering
    \includegraphics[width=0.8\textwidth]{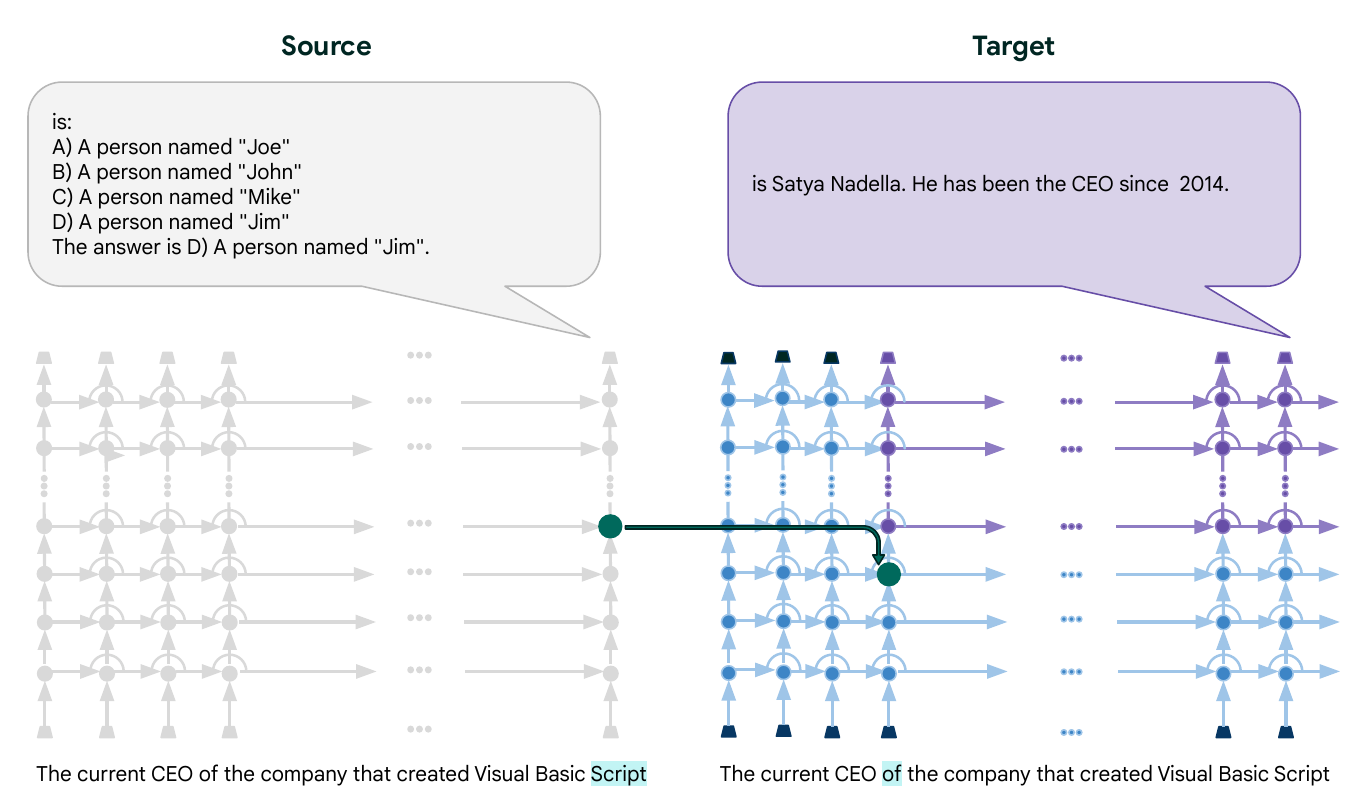}
    \caption{An illustration of CoT \methodName{} on a single example. In this example, $\pi_1 \is ``\texttt{the company that created Visual Basic Script}"$, $\pi_2 \is ``\texttt{The current CEO of}"$, $\sourcePrompt = \targetPrompt \is [\pi_2][\pi_1]=``\texttt{The current CEO of the company that created Visual Basic Script}"$. Note that $\sourceModel=\targetModel$ and $\transformation \is \identity$.}
    \label{fig:cot-example}
\end{figure*}

\subsection{Expressiveness from Cross-Model Patching}
\label{sec:cross_model}

A possible avenue for improving inspection capabilities is to explain a given model with a model that is more expressive \cite{bills2023language}. In \methodFamilyName{}, this means to patch a representation of $\mathcal{M}$ into a more expressive model $\mathcal{M}^*$. However, it is not clear if such an intervention would yield plausible results, due to possible discrepancies between the two models resulting from different architectures, optimization processes, and so on. Here, we present experiments on next-token prediction and entity resolution that exemplify not only the feasibility, but also the opportunities unlocked by patching across models of the same family.

\paragraph{Next-Token Prediction} We repeat the experiment in \S\ref{sec:ntp_decoding}, using the token identity \methodName{}. To overcome discrepancies between the models, we learn affine mappings between their layers (similarly to Tuned Lens).
Our results show that source and target layers on the diagonal exhibit the highest precision values, and that patching representations to an early layer of the larger model are the most effective. Overall, suggesting that when $\targetModel$ and $\sourceModel$ are from the same model family, it is possible to leverage $\targetModel$ for decoding information from the representations of $\sourceModel$. 
For detailed results on Precision@1 and Surprisal, see \S\ref{sec:appendix-cross-model-ntp}.

\paragraph{Entity Resolution in Early Layers} 
We now show that using a large model as $\targetModel$ can enhance the output expressivity. To this end, we repeat our entity resolution experiment in \S\ref{sec:input_processing} with Vicuna model family, setting $\sourceModel\!\!\is\!\!\text{7B}, \targetModel\!\!\is\!\!\text{13B}$. Fig.~\ref{fig:entity_processing} shows the cross-model patching results (green lines) compared to the same-model patching $\sourceModel\!\!\is\!\!\text{7B}, \targetModel\!\!\is\!\!\text{7B}$ (blue line). 
The results show that cross-model patching from a smaller model to its larger version generally improves the ability to inspect the input contextualization, both for popular and rare entities. For Pythia, since the smaller model outperforms the larger one, cross-model patching is not as effective (see \S\ref{sec:appendix-entity-resolution}).

\section{Application: Correcting Multi-Hop Errors}
\label{sec:multihop}
Multi-hop reasoning is a challenging problem~\cite{zhong2023mquake}. 
While a language model may be capable of correctly answering each step independently, it could still fail at processing the connection between different steps, resulting in an incorrect prediction. Recent attempts to improve multi-hop reasoning rely on prompting the model to generate a step-by-step answer autoregressively \citep[e.g.,][]{wei2022chain, yao2023tree, besta2023graph}, some with an iterative process of self-refinement \citep[e.g.,][]{madaan2023self}. 
However, achieving similar benefits might be possible via directly rewiring the model's intermediate computation.

Here, we show that \methodFamilyName{} can improve multi-hop reasoning performance \textit{without}  generating the reasoning steps, particularly in cases where the model fails at completing a multi-hop query despite being successful in each reasoning step independently. Via \methodFamilyName{}, one can surgically operate on the model representations, reroute its intermediate answer to one reasoning step, simplify the consequent step, and ultimately correct the final prediction.

\paragraph{Data} Building on \citet{hernandez2023linearity}, we systematically generate all valid multi-hop factual and commonsense reasoning queries where $\object_1 = \subject_2$. We conduct experiments on Vicuna (13B), focusing on samples where $\sourceModel$ accurately represents both $\tau_1$ and $\tau_2$ independently, that is, $\object$ appears in the next 20 tokens $\sourceModel$ generates conditioned on the prompt $\pi$ that verbalizes $\subject$ and $\relation$. This process yields 1,104 multi-hop reasoning samples, out of which 46 satisfy the above criteria and are used for evaluation. See more details in \S{\ref{sec:appendix-cot}}.

\paragraph{Experimental Setup} Following the notation in \S\ref{sec:feature_extraction}, let $\tau_1 = (\subject_1, \relation_1, \object_1)$ represent the relation $\relation_1$ between a subject entity $\subject_1$ and an object entity $\object_1$. Let $\tau_2 = (\subject_2, \relation_2, \object_2)$ represent another tuple such that $\subject_2 = \object_1$. A multi-hop reasoning query pertaining to $\tau_1$ and $\tau_2$ is a prompt composed of two parts: $\pi_1$ is a verbalization of $\subject_1$ and $\relation_1$ from which $\object_1$ can be inferred; $\pi_2$ is a verbalization of $\relation_2$, from which $\object_2$ can be inferred after its concatenation with $\pi_1$. For example, Let $\tau_1 \is (``\texttt{Visual Basic}", ``\texttt{product of}", ``\texttt{Microsoft}")$ and $\tau_2 \is$ (``\texttt{Microsoft}", ``\texttt{company CEO}", ``\texttt{Satya Nadella}"). An example verbalization of these tuples is $\pi_1 \is $``\texttt{the company that created Visual Basic}'', $\pi_2 \is ``\texttt{The current CEO of}"$, leading to the multi-hop query $[\pi_2][\pi_1]=$``\texttt{The current CEO of the company that created Visual Basic}''.

\paragraph{Method} We introduce a Chain-of-Thought (CoT) \methodName{} to fix multi-hop reasoning via intervening on the computation graph and rerouting representation likely to capture $\object_1$ in place of $\subject_2$. Concretely, $\sourcePrompt$ refers to the formed query discussed above, and we use the following configuration: $\targetPrompt \is \sourcePrompt, \targetModel \is \sourceModel, \sourcePosition \is \sourceSeqLen, \targetPosition \is$ the token preceding $\pi_1$. 
We evaluate the outputs in terms of accuracy, similarly to \S\ref{sec:feature_extraction}. For a sample $\sourcePrompt$, the \methodName{} is considered accurate if $ \exists \, (\sourceLayer, \targetLayer) : \sourceLayer \in [1, \ldots, \numSourceLayers], \targetLayer \in [1, \ldots, \numTargetLayers]$ where the autoregressive generation up to 20 tokens includes $\object_2$.
Fig.~\ref{fig:cot-example} illustrates the CoT \methodName{} with an example. We use the following configuration for CoT \methodName{}: $\sourcePrompt \is \pi_1, \targetPrompt \is \pi_2, \sourcePosition \is \sourceSeqLen, \targetPosition \is \targetSeqLen$, which is equivalent to $\sourcePrompt = \targetPrompt \is [\pi_2][\pi_1]$ and adjusting the attention mask such that no token in $\sourcePrompt$ has visibility to $\pi_2$ and no token in $\targetPrompt$ has visibility to $\pi_1$. In addition, we consider two baselines. 

\paragraph{Vanilla Baseline} For this baseline, we set $\sourcePrompt \is [\pi_1][\pi_2]$, we let the model autoregressively generate up to 20 tokens and check whether $\object_2$ appears in the generation.

\paragraph{Chain-of-Thought Baseline} Here, the setup and evaluation is similar to the vanilla baseline, except that we prepend \texttt{"Let's think step by step."} to $\sourcePrompt$, following \citep{wei2022chain}. We then let the model generate up to 20 tokens and check whether $\object_2$ appears in the generation. Note that this experiment uses Vicuna (13B). Vicuna is based on LLaMA, with supervised finetuning on additional instruction data, which makes it amenable to chain-of-thought prompting.

\paragraph{Results} 
While the vanilla baseline accuracy is only 19.57\%, and CoT baseline accuracy is 35.71\%, our proposed \methodName{} achieves 50\% accuracy. 
For more details about the interaction between $\sourceLayer$ and $\targetLayer$, and how it affects the success rate, see Fig.~\ref{fig:cot_heatmap} in \S\ref{sec:appendix-cot}.

We emphasize that our primary goal in this section is not to devise a new method to solve multi-hop queries that is necessarily a competitor to CoT, but rather to make a proof-of-concept while comparing with CoT as a common reference. We highlight that we have taken advantage of the extra structural information about the queries given the prior knowledge about how they were synthesized. One could potentially automate this by learning the right places to patch. \citet{li2024understanding} recently proposed an optimization-based approach for directly patching multi-head self-attention in mid-layers, which can be viewed as soft position-selection, and works effectively in multihop error correction, suggesting that inferring the right positions to patch could be effective.
However, even if optimal source and target position are automatically decided upfront, Patchscope and CoT may not be directly comparable. CoT generates multiple steps which fundamentally extend the computational power of the LLM \cite{merrill2024expressive}, but a Patchscope with pre-identified $(l, l^*)$ only makes $O(1)$ inference passes.

\section{Conclusion}
\label{sec:conclusion}

We present \methodFamilyName{}, a simple and effective framework that leverages the ability of LLMs to generate human-like text for decoding information from intermediate LLM representations. 
We show that many existing interpretability methods can be cast as specific \methodName{} instances, and even these only cover a small portion of the framework's possible configurations. Moreover, new underexplored \methodNamePlural{} substantially improve our ability to decode various types of information from the model's internal computation, such as the output prediction and knowledge attributes, typically outperforming prominent methods that rely on projection to the vocabulary and probing.
In addition, our framework enables new capabilities, such as analyzing the contextualization process of input tokens in early LLM layers, and can correct multi-hop reasoning. 

There are multiple future research directions to consider. An important factor in the effectiveness of a chosen target prompt is how the information from the patched position propagates during inference to other positions and across layers. 
Understanding how to best use a given target prompt, perhaps automatically, is an important factor for using \methodNamePlural{}. Another avenue for future work is investigating the effectiveness of few-shot target prompts compared to zero-shot/instruction-based prompts. More expressive instruction-based target prompts, for example, could enable extracting more complex information. Additionally, while our cross-model patching focused on models from the same family, it will be valuable to explore which mapping functions would enable patching across models from different families and perhaps with different architectures. Other directions for future work include applications across different domains and modalities, investigating variants with simultaneous multi-token patching or multi-layer patching to mitigate the risks of placeholder contamination, and presenting recipes for task-specific and task-agnostic \methodNamePlural{}.

\section*{Acknowledgements}
We thank Amir Globerson for feedback on writing and presentation of results. We thank Ardavan Saeedi, Martin Wattenberg, Ellie Pavlick, the AI Explorables team\footnote{\url{https://pair.withgoogle.com/explorables}} at Google Research, and Jasmijn Bastings for their helpful comments.

\section*{Impact Statement}

\paragraph{Societal Impact}
This paper presents a new framework for interpreting hidden representations of large language models. Interpretability methods in general can be used to investigate models' reliability and safety prior to deployment. We hope that \methodFamilyName{} framework facilitates progress in this area with the introduction of inspection tools that are more expressive, robust across layers, and do not require training data.

\paragraph{Limitations} 
While our proposed framework is not limited to any particular architecture or domain, the experimental evidence provided in this paper focuses on autoregressive Transformer-based language models, and future work is needed to verify its effectiveness in other setups.

\bibliography{references}
\bibliographystyle{icml2024}

\newpage
\appendix
\onecolumn
\section{Detailed Configuration of Prior Methods as \methodName{} Instances}
\label{sec:method-configs}

In this section, we first reiterate \methodFamilyName{} premise and distinguish it from prior work. We then detail \methodName{} configurations that replicate prior methods. We highlight that clearly formulating \methodFamilyName{} in the form of source and target quintuplets, respectively $(S, i, M, l)$ and $(T, i^*, M^*, l^*)$, is not in itself a contribution of this work. However, it is what facilitates the main contributions that are 3-fold: 1) The framework ties methods that otherwise might seem unrelated, e.g., logit lens and attention knockout,  2) makes it easy to come up with new configurations that significantly improve them, 3) makes it easy to see the space that prior methods do not cover, and therefore open up new applications altogether.

We focus on a few inspection objectives. Note that some of the baselines considered have not been originally designed to address these objectives, but are the closest solutions in literature. Particularly, we ask, given a (set of) hidden representation(s) detached from their context, how well can we do next-token prediction, feature extraction, and analysis of model’s contextualization process.

\subsection{Next-token Prediction}
\paragraph{Goal:} Estimating output probability distribution $p^L$ given a hidden representation $h^l$, particularly to inspect when the model has enough information to make the final prediction.

\paragraph{Prior Work:} Many recent methods to estimate output probability distribution given an inner representation use projection onto the vocabulary space  \citep[e.g.,][]{logitlens, din2023jump, belrose2023eliciting}. Considering $(S, i, M, l)$ and $(T, i^*, M^*, l^*)$ formulation above, all these methods can be seen as \methodNamePlural{} where $\mathbf{M=M^*,  l^*=L}$, where $\mathbf{L}$ is the number of layers.

\paragraph{Distinguishing \methodName{} \& its Improvements:} We propose the token identity \methodName{} that unlike \citep{belrose2023eliciting, din2023jump} does not require a learned mapping, yet is more successful. Note that in this task, setting $\mathbf{l^*=L}$ which all the prior methods discussed above have in common, means that the choice of $\mathbf{T}$ does not matter. However, in our proposed \methodName{}, we show that setting $\mathbf{l^*=l}$ where $l$ is arbitrary layer, and using a few-shot token identity prompt $\mathbf{T}$ that encourages repetition results in significant improvements, and by using the target model’s computation from layer $l^*$ onward, there is no need for training a separate mapping function.

\subsection{Feature Extraction}
\paragraph{Goal:} Extracting specific attributes of the subject given its hidden representation $h^l$ (e.g., extracting ``largest city of'' from the representation of ``States'' in ``United States'').

\paragraph{Prior Work:} Classification probes \citep{alain2017understanding} are the most commonly used methods for this purpose. They work by training a classifier on a dataset of prompts and their corresponding labels (e.g., countries and their corresponding largest cities).

\paragraph{Distinguishing \methodName{} \& its Improvements:} We show an intuitive \methodName{} (i.e., $T=$ ``The largest city in x'') significantly outperforms probing across various tasks and layers. Note that this \methodName{}  1) does not require supervised training, 2) is not limited by a predefined set of labels, and 3) is more accurate than a linear probe.

\subsection{Analysis of Model’s Contextualization Process}
\paragraph{Goal:} Identifying how entity tokens are processed, and at which layer the entity is fully resolved.

\paragraph{Prior Work:} It is hard to answer this question with existing methods. First, methods that use vocabulary projection are focused on output prediction (rather than input processing) and fail particularly in early layers, precisely where entity resolution happens.
Second, probing requires a predefined set of classes, which is not expressive enough to show the gradual entity resolution process, and nontrivial. For example, consider ``Alexander the grea'' input. We found that ``Great Britain'' and ``Great depression'' arose in the gradual resolution process. However, these are non-trivial choices a priori. We can train a multi-class classifier to choose one out of all possible entities (or binary classifiers for every possible entity), which is unlikely to work given the huge space of possible entities.
Lastly, most prior work using activation patching aims to answer a different question: whether a certain representation plays a key role in the model's forward computation and the final prediction \citep[e.g.,][]{meng2022locating, geva-etal-2023-dissecting}. However, they can be viewed as \methodName{} instances, and provide indirect signals about the entity resolution process, namely, showing subject information accrues in the last token after a few layers.

\paragraph{Distinguishing \methodName{} \& its Improvements:} To the best of our knowledge, there are not any methods that verbalize the gradual entity resolution process in coherent text, and the proposed \methodName{} is the first to do that.

In summary, as discussed in \ref{sec:encompass_prior_methods}, many prior methods can be seen as \methodNamePlural{}. Tab. \ref{tab:method-configs} details the configurations of \methodFamilyName{} that yield various methods introduced in prior work.

\begin{table*}[!ht]
\setlength\tabcolsep{3pt}
\centering
\caption{\methodFamilyName{} is a novel framework for inspection of hidden representations in language models. Many prior inspection methods with various objectives can be viewed as \methodNamePlural{}, as detailed in the ``Configuration'' column (see notation description in~\S\ref{sec:method}). The rows highlighted in green show our new configurations that overcome several limitations of prior methods through more expressive inspection that is training-data free and is more robust across layers. Additionally, the generality of this framework enables novel inspection possibilities that were unexplored before. When the target prompt ($\targetPrompt$) is not specified, it means that the output would be invariant to the choice of $\targetPrompt$. When not specified, $\transformation \is \identity$ and $\targetModel = \sourceModel$.}
\vspace{0.1cm}
\label{tab:method-configs}
\resizebox{1.\textwidth}{!}{%
\renewcommand{\arraystretch}{1.2}
\begin{tabular}{m{2cm}m{1cm}m{6cm}m{1.75cm}m{1.75cm}m{1.5cm}||m{12cm}}
\toprule
\textbf{Inspection Objective} & \multicolumn{2}{r}{\textbf{}} &
  \textbf{Expressive} &
  \textbf{Training Data \newline Free} &
  \textbf{Robust Across \newline Layers} & \textbf{Configuration} \\ 
 \toprule
  \multirow{6}{*}{\parbox{2cm}{\textbf{Inspecting \newline Output \newline Distribution}}} &
  \multicolumn{2}{l}{\cellcolor[HTML]{DFF6CC} Few-shot token identity \methodName{} (\S\ref{sec:ntp_decoding})} &
  \cellcolor[HTML]{DFF6CC} \color[HTML]{34A853} {\faCheck \faCheck} &
  \cellcolor[HTML]{DFF6CC} \color[HTML]{34A853} {\faCheck} &
  \cellcolor[HTML]{DFF6CC} \color[HTML]{34A853}{\faCheck \faCheck} & 
  \cellcolor[HTML]{DFF6CC}{$\targetLayer \is \sourceLayer$, \newline $\targetPrompt \is ``\texttt{tok}_1 \rightarrow \texttt{tok}_1 \texttt{; tok}_{2} \rightarrow \texttt{tok}_2 ; \ldots; \texttt{tok}_{k}"$}\\
 &
  \multicolumn{2}{l}{\parbox{8cm}{Logit Lens \cite{logitlens}, Embedding Space \newline Analysis \cite{dar2023analyzing}}} &
   \color[HTML]{34A853}{\faCheck} &
   \color[HTML]{34A853}{\faCheck} &
   \color[HTML]{ff0000}{\faClose} & $\targetLayer \is \numTargetLayers$ \\
 &
  \multicolumn{2}{l}{\parbox{8cm}{Tuned Lens \cite{belrose2023eliciting}}} &
   \color[HTML]{34A853}{\faCheck} &
  {For learning \newline mappings} &
   \color[HTML]{34A853}{\faCheck} & $\targetLayer \is \numTargetLayers, \transformation \is \text{Affine}$\\
  &
  \multicolumn{2}{l}{Future Lens \cite{pal2023future}} &
   \color[HTML]{34A853}{\faCheck} &
  {For learning \newline mappings} &
   \color[HTML]{34A853}{\faCheck \faCheck} & $\targetLayer \is \sourceLayer, \transformation \is \text{Linear}, \targetPrompt \is \text{Fixed or learned soft prompt}$ \\ \hline
   
  \multirow{5}{*}{\parbox{2cm}{\textbf{Feature \newline Extraction}}} &
  \multicolumn{2}{l}{\cellcolor[HTML]{DFF6CC} Zero-shot feature extraction \methodName{} (\S\ref{sec:feature_extraction})} &
  \cellcolor[HTML]{DFF6CC} \color[HTML]{34A853}{\faCheck \faCheck} &
  \cellcolor[HTML]{DFF6CC} \color[HTML]{34A853}{\faCheck} &
  \cellcolor[HTML]{DFF6CC} \color[HTML]{34A853}{\faCheck \faCheck} &
  \cellcolor[HTML]{DFF6CC}{$\targetLayer \is j^{\prime} \in [1, \ldots, \numTargetLayers], \targetPosition \is \targetSeqLen$,\newline $ \targetPrompt \is $ relation verbalization followed by \texttt{x}}\\
 &
  \multicolumn{2}{l}{\parbox{8cm}{LRE Attribute Lens \cite{hernandez2023linearity}}} &
  \color[HTML]{34A853}{\faCheck}  &
  \multicolumn{1}{l}{\parbox{1.75cm}{For linear \newline relation \newline approx.}} &
  \color[HTML]{34A853}{\faCheck \faCheck} & $\targetLayer \is \numTargetLayers, \transformation \is \text{Linear with additional variables},  \targetPrompt \is \sourcePrompt$\\
  & 
 \multicolumn{2}{l}{\parbox{7cm}{Probing \citep[e.g.,][]{belinkov2019analysis, belinkov2022probing, alain2017understanding, wang2023gaussian}}} & \color[HTML]{ff0000}{\faClose}  & 
 {For training \newline probe} &
 \color[HTML]{34A853}{\faCheck} & N/A \\ \hline
 
  \multirow{5.5}{*}{\parbox{2cm}{\textbf{Entity \newline Resolution}}} &
  \multicolumn{2}{l}{\cellcolor[HTML]{DFF6CC} Entity description \methodName{} (\S\ref{sec:input_processing})} &
  \cellcolor[HTML]{DFF6CC} \color[HTML]{34A853}{\faCheck \faCheck} &
  \cellcolor[HTML]{DFF6CC} \color[HTML]{34A853}{\faCheck} &
  \cellcolor[HTML]{DFF6CC} \color[HTML]{34A853}{\faCheck \faCheck} &
  \cellcolor[HTML]{DFF6CC}{$\targetLayer \is \sourceLayer$, $\targetPosition \is \targetSeqLen$, \newline $\targetPrompt \is$ ``\texttt{subject}$_1$\texttt{:\;description}$_1$\texttt{,\;\ldots,\;} \texttt{subject}$_k$\texttt{:\;description}$_k$ \texttt{,\;x}"}\\
  
  &  \multicolumn{2}{l}{\cellcolor[HTML]{DFF6CC} X-model entity description \methodName{}  (\S\ref{sec:cross_model})} &
  \cellcolor[HTML]{DFF6CC} \color[HTML]{34A853}{\faCheck \faCheck \faCheck} &
  \cellcolor[HTML]{DFF6CC}{For learning \newline mappings} &
  \cellcolor[HTML]{DFF6CC} \color[HTML]{34A853}{\faCheck \faCheck} &
  \cellcolor[HTML]{DFF6CC}{$\targetModel \is \text{a larger variant of}\,\sourceModel, \targetLayer \is \sourceLayer, \targetPosition \is \targetSeqLen$, \newline $ \targetPrompt \is$ ``\texttt{subject}$_1$\texttt{:\;description}$_1$\texttt{,\;\ldots\;,\;subject}$_k$\texttt{:\;description}$_k$\texttt{,\;x}"}\\
  
  &
  \multicolumn{2}{l}{Causal Tracing \cite{meng2022locating}} &
   \color[HTML]{ff0000}{\faClose} &
   \color[HTML]{34A853}{\faCheck} &
   \color[HTML]{34A853}{\faCheck \faCheck} & $\targetLayer \is \sourceLayer,
  \targetPrompt \is \sourcePrompt + \epsilon, \epsilon \sim \mathcal{N}(0, \sigma)$ \\
  &
  \multicolumn{2}{l}{\parbox{6cm}{Attention Knockout \cite{wang2022interpretability, conmy2023automated, geva-etal-2023-dissecting}}} &
   \color[HTML]{ff0000}{\faClose} &
   \color[HTML]{34A853}{\faCheck} &
   \color[HTML]{34A853}{\faCheck \faCheck} & $\targetLayer \is \text{Multiple}, \transformation \is \constantZero,
  \targetPrompt \is \sourcePrompt$\\
  \midrule \midrule
  
 \multirow{2.5}{*}{{\parbox{2cm}{\textbf{Inspection \newline Application}}}} &
  \multicolumn{2}{l}{\parbox{8cm}{Early Exiting, e.g., Linear Shortcuts \cite{din2023jump}}} &
  \color[HTML]{34A853}{\faCheck} &
   {For learning \newline mappings} &
  \color[HTML]{34A853}{\faCheck} &
  $\targetLayer \is \numTargetLayers, \transformation \is \text{Affine}$ \\
  
  & 
  \multicolumn{2}{l}{\parbox{8cm}{Caption Generation, e.g., Linear Mapping\newline \cite{merullo2022linearly}}} & 
  \color[HTML]{34A853}{\faCheck} &
  {For learning \newline mappings} & 
  \color[HTML]{34A853}{\faCheck} & 
  $\targetModel \is \text{A language model of choice}, \targetLayer \is \numTargetLayers, \transformation \is \text{Affine}$ \\ 
 \bottomrule
\end{tabular}
}\vspace{-.5em}
\end{table*}

\section{Next-Token Prediction Additional Details and Experimental Results}
In this section, we provide more information about the models, data, and additional target prompts for few-shot token identity \methodName{} configurations. It is worth highlighting that after patching, we continue computations from the target layer onward. Therefore, since in this \methodName{}, source layer and target layer are the same, there are going to be more computations compared to baseline methods like LogitLens \citep{logitlens, geva-etal-2022-lm} and Tuned Lens \citep{belrose2023eliciting,din2023jump} where the target layer is fixed at the final layer.

\subsection{Models}
\label{section:ntp_models}
We use LLaMA2 (13B)\footnote{LLaMA 2 is licensed under the LLAMA 2 Community License, Copyright © Meta Platforms, Inc. All Rights Reserved.} \citep{touvron2023llama}, Vicuna (13B)\footnote{Vicuna is subject to the LLAMA 2 Community License, terms of use of the data generated by OpenAI, and privacy practices of ShareGPT. The code is released under the Apache License 2.0.} \citep{vicuna2023}, GPT-J (6B) \citep{gpt-j}\footnote{GPT-J-6B weights are licensed under version 2.0 of the Apache License.}, and Pythia (12B)\footnote{Pythia is released under Apache License 2.0, Copyright © 2023 EleutherAI.} \citep{pythia}. LLaMA2 was pre-trained on 2T tokens from a mix of publicly available data. Vicuna is a LLaMA1 \citep{touvron2023llama1} model that was pre-trained on 1T tokens and fine-tuned on 70K user-shared conversations~\footnote{Conversations were collected from \url{www.sharegpt.com}.}.
The primary architectural differences between LLaMA2 and Vicuna (LLaMA1) include a different context length and grouped-query attention. Pythia and GPT-J were pre-trained using a deduplicated version of the Pile corpus\footnote{Pile dataset is available under MIT license.} \citep{gao2020pile}, and for about 300B and 402B tokens, respectively.

\subsection{Training and Evaluation Data}
\label{section:ntp_data}
We use 12,000 random samples from the Pile, partitioned into 10,000 examples for training the affine mappings, and 2,000 examples for evaluation. In our pre-processing strategy, we introduce randomness in the patching positions by trimming the input sequence length of each example. 

\subsection{Additional Few-Shot Token Identity Prompts}
\label{section:add_token_id}
In this section, we provide additional details about the selection of the demonstrations for the token identity baseline, and further evaluate the robustness of LLaMA2 (13B) \citep{touvron2023llama} to various token identity prompts. 

\paragraph{Demonstrations Construction} For the demonstrations used in this experiment, we sample a random set of $k$ tokens for all the models, where $k$ was also randomly sampled from the interval $[1,\ldots,10]$.

\paragraph{Robustness to Additional token IDs' Demonstrations} We randomly generate five realizations of token IDs series of varying lengths, formatted as ``\texttt{tok$_1$ $\rightarrow$ tok$_1$ ; tok$_2$ $\rightarrow$ tok$_2$ ; $\ldots$  ; tok$_k$}'', similarly to the procedure from \S\ref{sec:ntp_decoding}. We also include an ablation experiment, introducing the \emph{Single Token Prompt} baseline. In this approach, the representation is transferred from the original multi-token prompt to a single-token prompt, while preserving the same position. This is crucial for maintaining consistent positional biases during the computation in the forward pass. The baseline is implemented by creating a target prompt in the format: ``\texttt{[PAD] [PAD] $\ldots$ [PAD] tok$_k$}''. Here, the number of \texttt{[PAD]} tokens is chosen such that ``\texttt{tok$_k$}'' retains its original position from the source prompt. 

\begin{figure*}[t!]
  \centering
  \subfigure[Precision@1]{\includegraphics[scale=0.45]{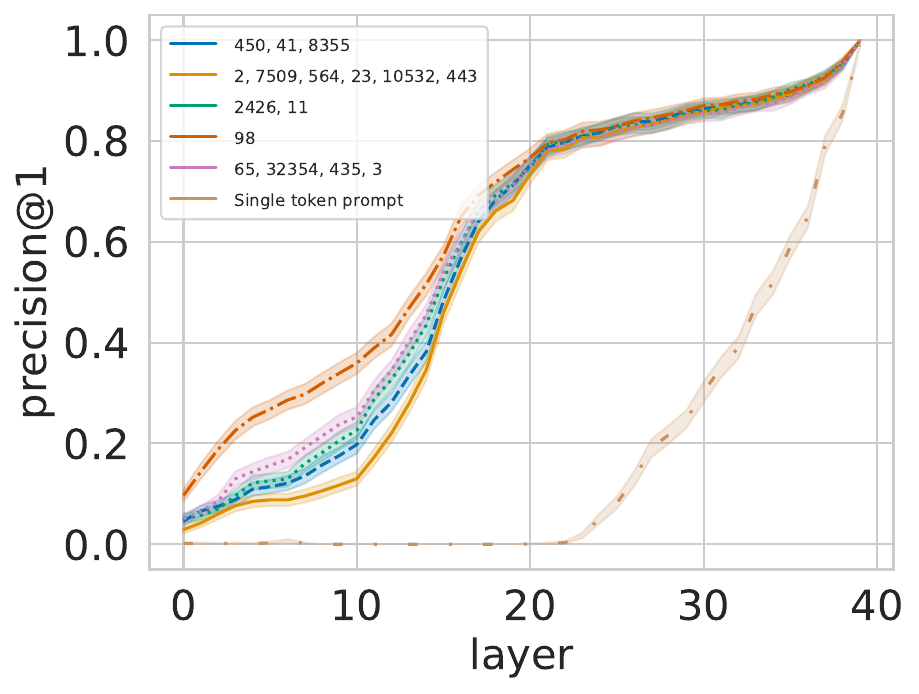}}\quad
  \subfigure[Surprisal]{\includegraphics[scale=0.45]{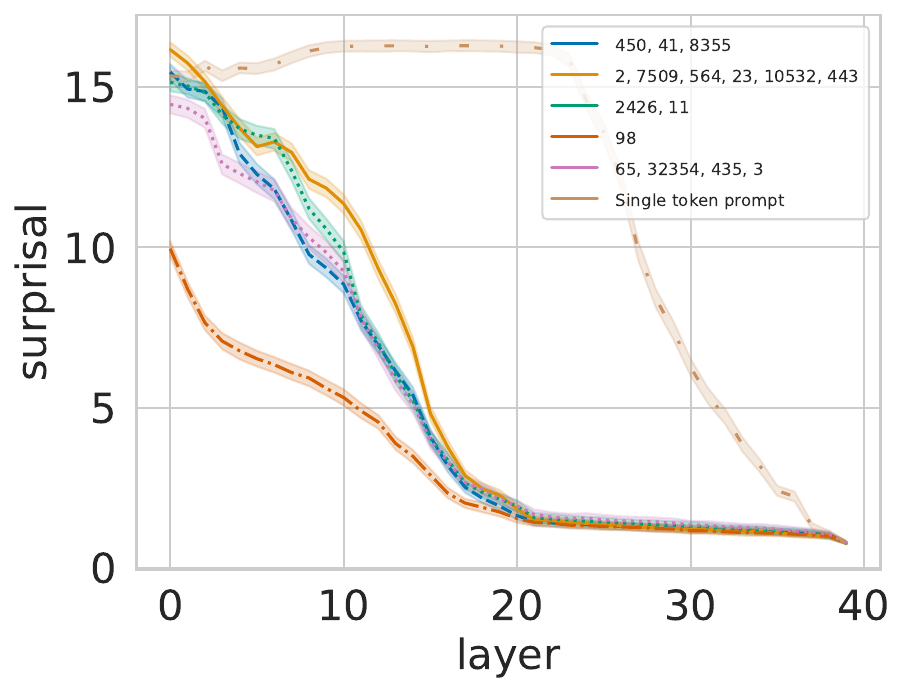}}
\vspace{-.5em}
\caption{Next-token prediction results for LLaMA2, using various token identity demonstrations (the token IDs appear in the legend). We report precision@1 ($\uparrow$ is better), and surprisal ($\downarrow$ is better).}
\label{fig:token_id_add_prompts}
\vspace{-1em}
\end{figure*}

The results are illustrated in Fig.~\ref{fig:token_id_add_prompts}, where a comprehensive overview of the evaluation metrics can be found in \S\ref{sec:ntp_decoding}. The results indicate the stability of the token identity baseline across a range of token identity demonstrations, particularly notable in the upper layers of the model. The \emph{Single Token Prompt} baseline overall seems to be less effective than using our token ID baseline.

\section{More Details on Attribute Extraction Experiments}
\label{sec:appendix-attribute-extraction}

\begin{table*}[t!]
    \centering
    \caption{Comparison between the zero-shot feature extraction \methodName{} and the logistic regression probe shows that despite using \textit{no training data}, the \methodName{} has a significantly higher accuracy than the baseline in most tasks $(p < 1e-5)$. Pairwise t-statistics and the corresponding p-values are included in the table. The columns corresponding to each method show accuracy (mean $\pm$ std).}
    \vspace{.5em}
    \label{tab:attribute_extraction_appendix}
    \resizebox{0.85\textwidth}{!}{%
    \bgroup
    \def\arraystretch{1.1}
    \begin{tabular}{clm{8em}m{10em}ll}
        \toprule
         & \textbf{Task} & \textbf{Probe} & \textbf{\methodName{}} & \textbf{T-statistic} & \textbf{p-value} \\
        \toprule
        \multirow{7}{*}{Commonsense} & Fruit inside color & $37.41 \pm 6.58$ & $37.99 \pm 18.67$ & $0.126$ & $0.901$\\
         & Fruit outside color &  $35.50 \pm 3.09$ & $\mathbf{71.00 \pm 13.26^{**}}$ & $12.426$ & $<1e-5$ \\
         & Object superclass & $\mathbf{68.92 \pm 10.69}^{**}$ & $55.71 \pm 10.81$ & $-5.25$ & $<1e-4$ \\
         & Substance phase & $73.77 \pm 3.74$ & $\mathbf{91.92 \pm 1.73^{**}}$ & $25.647$ & $<1e-5$ \\
        & Task done by person & $0 \pm 0$ & $\mathbf{62.96 \pm 16.513^{**}}$  & $19.632$ & $<1e-5$\\
         & Task done by tool & $10.14 \pm 3.23$ & $\mathbf{48.12 \pm 13.23^{**}}$ & $18.231$ & $<1e-5$\\
         & Work location  & $0 \pm 0$ & $\mathbf{13.58 \pm 9.37^{**}}$ & $7.45990$ & $<1e-5$\\
        \midrule
         \multirow{12}{*}{Factual} & Company CEO  & $4.99 \pm 2.56$ & $\mathbf{47.82 \pm 13.89^{**}}$ & $16.700$ & $<1e-5$ \\
         & Country capital city & 0 $\pm$ 0 & $\mathbf{61.61 \pm 14.14^{**}}$ & $22.426$ & $<1e-5$ \\
         & Country currency & $17.70 \pm 2.20$ & $\mathbf{50.95 \pm 8.85^{**}}$ & $20.293$ & $<1e-5$ \\
         & Country largest city & $0 \pm 0$ & $\mathbf{67.78 \pm 11.47^{**}}$ & $30.427$ & $<1e-5$ \\
         & Food from country & $5.13 \pm 3.66$ & $\mathbf{63.80 \pm 11.34^{**}}$ & $26.710$ & $<1e-5$ \\
         & Person father & $0 \pm 0$ & $\mathbf{25.34 \pm 8.42^{**}}$ & $15.482$ & $<1e-5$ \\
         & {Person plays position in sport} & {\centering $75.89 \pm 9.14$} & $72.20 \pm 7.21$ & $-2.066$ & $0.049$ \\
         & {Person plays pro sport} & $53.87 \pm 10.28$ & $46.28 \pm 14.19$ & $-2.020$ & $0.054$ \\
         & {Product by company} & $58.91 \pm 7.15$  & $63.24 \pm 10.74$ & $1.757$ & $0.091$ \\
         & Star constellation & $17.54 \pm 5.30$ & $18.35 \pm 5.06$ & $-2.98$ & $0.006$ \\
         & Superhero archnemesis & $0 \pm 0$ & $\mathbf{41.73 \pm 18.72^{**}}$ & $11.47044$ & $<1e-5$ \\
         & Superhero person & $0 \pm 0$ & $\mathbf{28.32 \pm 14.05^{**}}$ & $10.37461$ & $<1e-5$ \\
        \bottomrule
    \end{tabular}
    \egroup
    }
\end{table*}

\paragraph{Dataset Details} We start from the factual and commonsense reasoning subsets introduced by \citet{hernandez2023linearity}\footnote{The dataset released by \citet{hernandez2023linearity} is available under MIT license.}. This dataset includes 8 commonsense and 25 factual relations. Recall that each datapoint is representing a triplet $(\subject, \relation, \object)$, where $\subject$ refers to the subject, $\relation$ represents the relation, and $\object$ stands for the object (see \ref{sec:feature_extraction} for notation details). For each data point representing the triplet $(\subject, \relation, \object)$, we sample 5 utterances from \texttt{Wikitext-103} dataset \cite{merity2016pointer} including $\subject$ . We then truncate the sampled text to a window of random length up to 20 tokens that contains $\subject$. This constitutes our source prompt, $\sourcePrompt$. Note that for each model, we filter the data to samples for which the underlying model correctly encodes the tuple. For experiments with zero-shot target prompt $\targetPrompt$ that includes $\relation$ followed by $\subject$, we autoregressively generate the next 20 tokens, and only keep examples where $\object$ appears in the generation. For experiments with few-shot demonstrations in $\targetPrompt$, after generating the next 20 tokens, we do an additional post-processing step. If the demonstration template of an example is identified in the generation, all the following tokens would be dropped. The example is used for evaluation only if $\object$ appears in the post-processed truncated generation. To have a reasonable amount of data for training the classification probe baseline, tasks with fewer than 15 datapoints are dropped from the analysis. For GPT-J, 5 commonsense and 7 factual reasoning tasks remain after applying the above postprocessing steps.

\paragraph{Results on Additional Tasks} We provide additional results on various factual and commonsense reasoning tasks. For a comparison of the zero-shot feature extraction \methodName{} and the logistic regression probe averaged across layers, see t-statistic details in Tab.~\ref{tab:attribute_extraction_appendix}. 

\begin{figure*}
    \centering
        \subfigure[Substance Phase]{\includegraphics[width=0.24\textwidth]{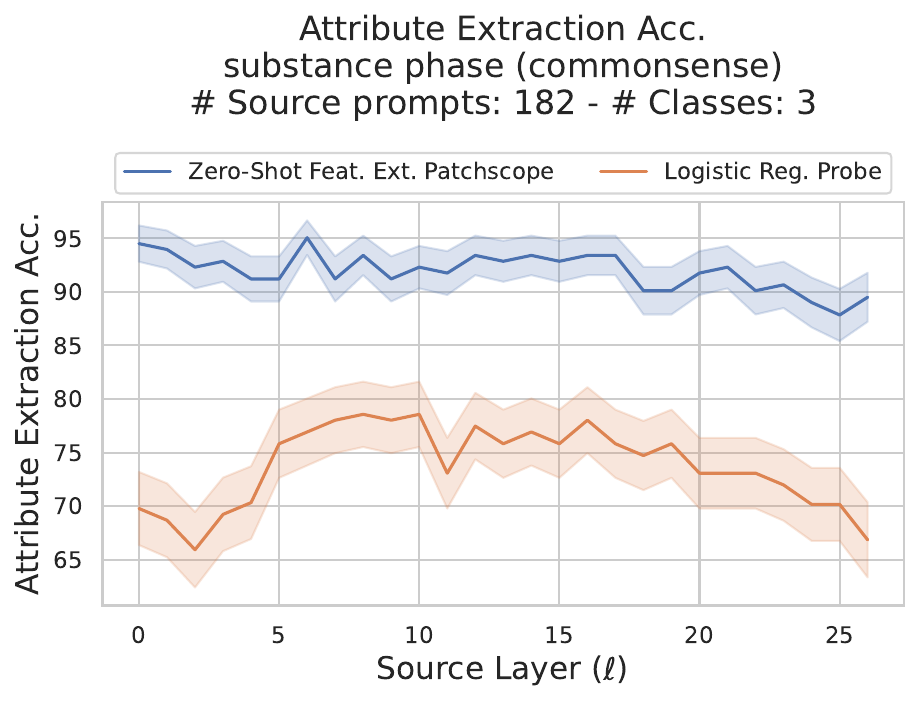}}
        \subfigure[Task Done By Tool]{\includegraphics[width=0.24\textwidth]{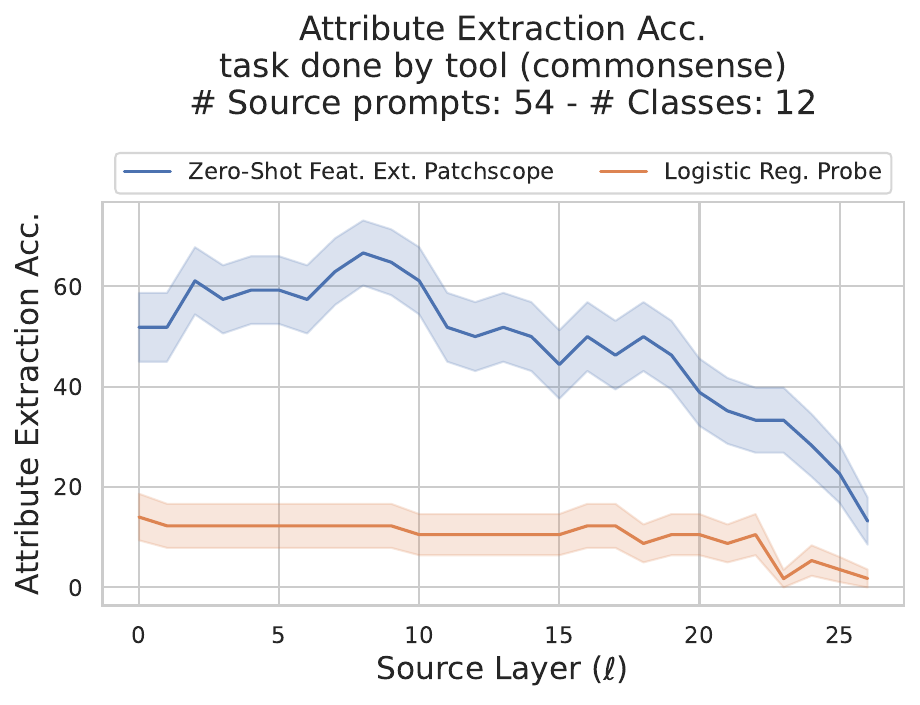}} 
        \subfigure[Country Currency]{\includegraphics[width=0.24\textwidth]{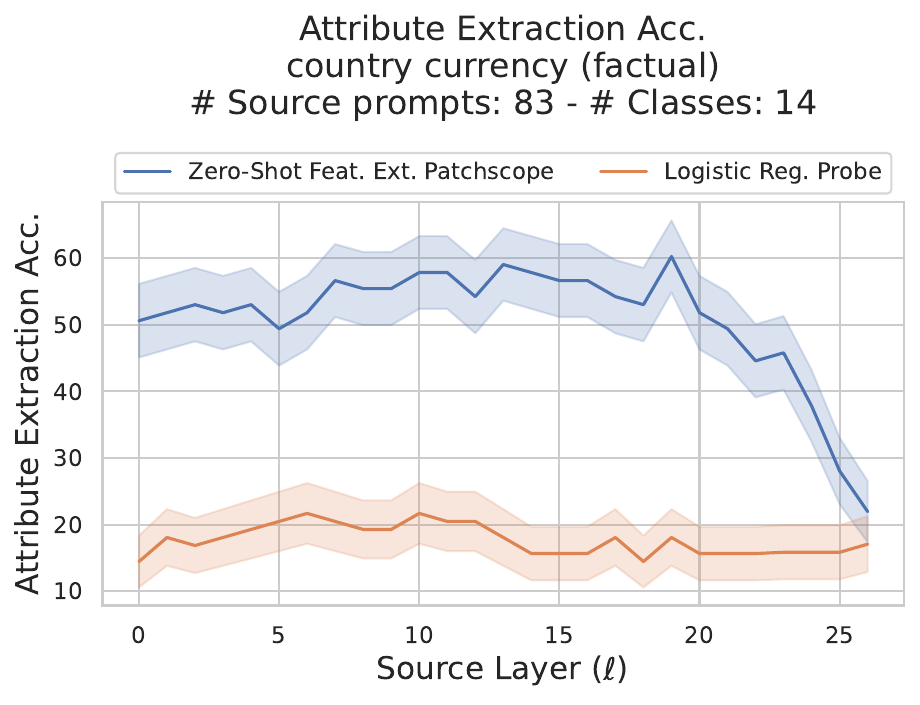}} 
        \subfigure[Food From Country]{\includegraphics[width=0.24\textwidth]{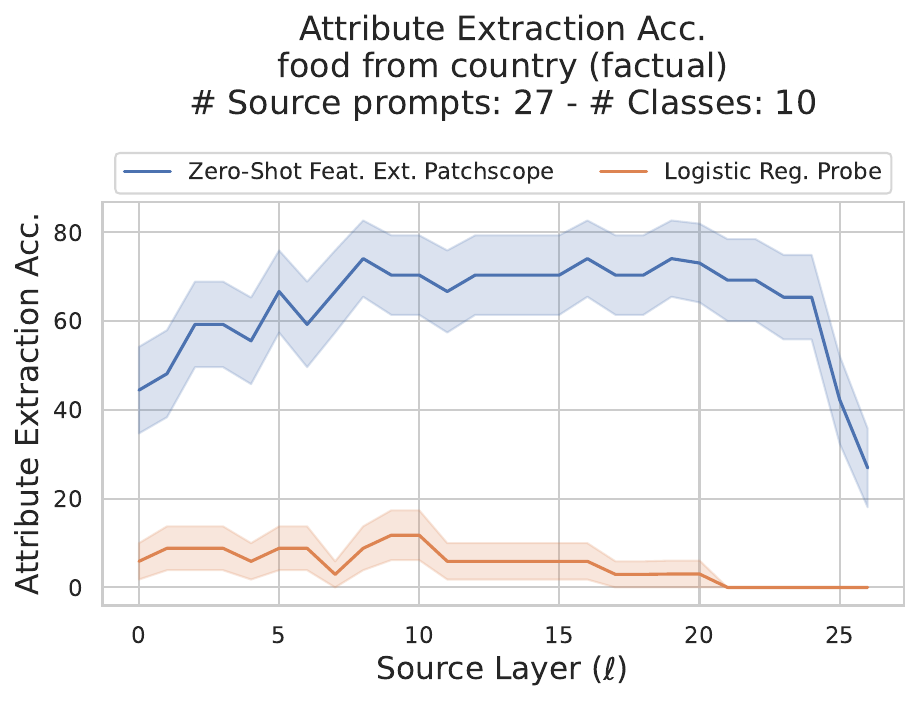}} \\
        \subfigure[Company CEO]{\includegraphics[width=0.24\textwidth]{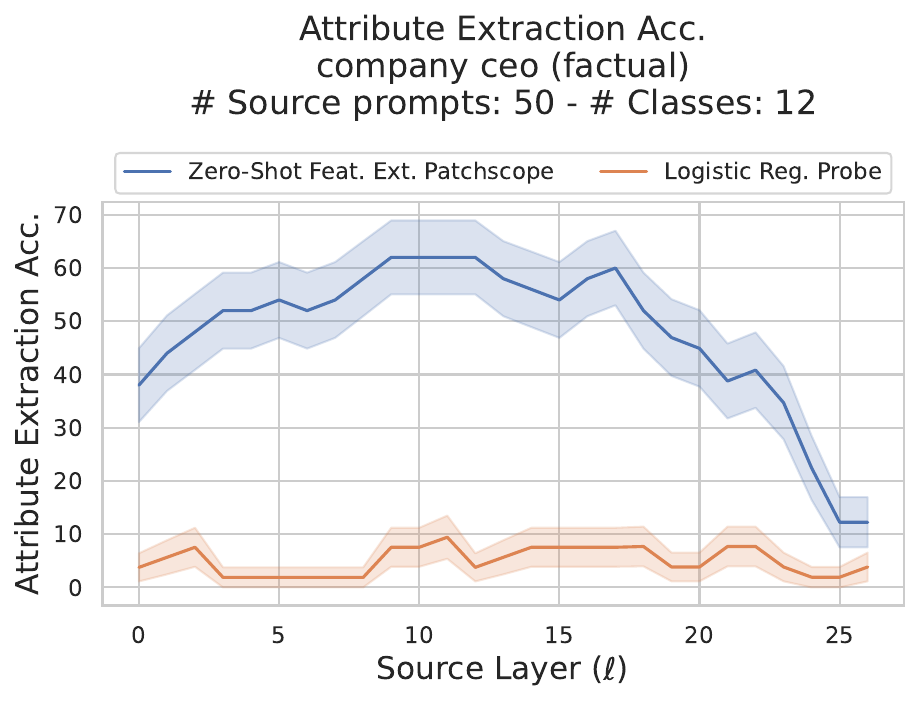}}
        \subfigure[Fruit Outside Color]{\includegraphics[width=0.24\textwidth]{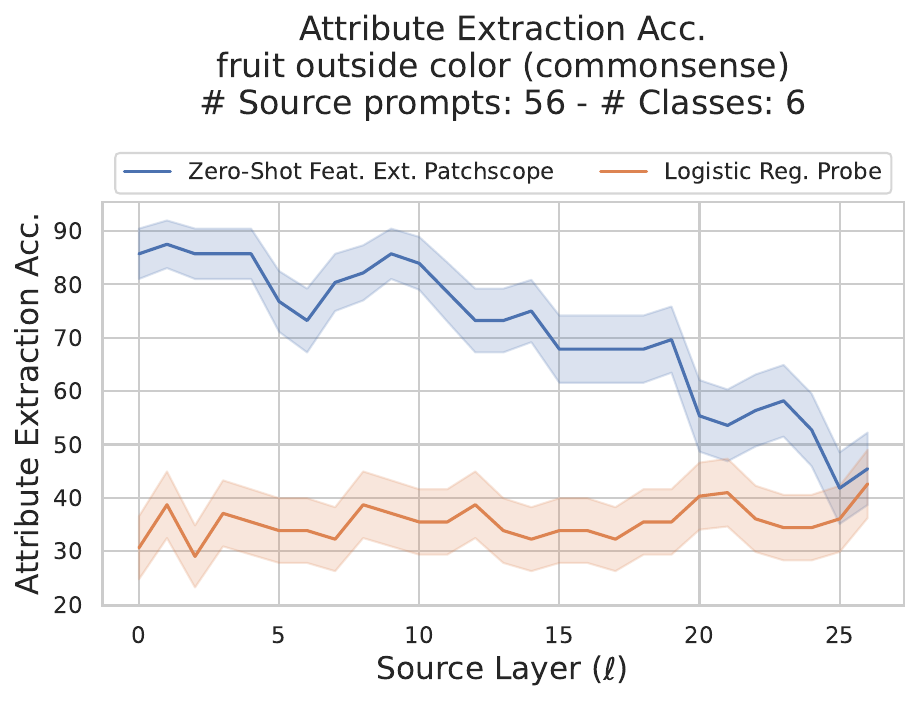}}
        \subfigure[Product By Company]{\includegraphics[width=0.24\textwidth]{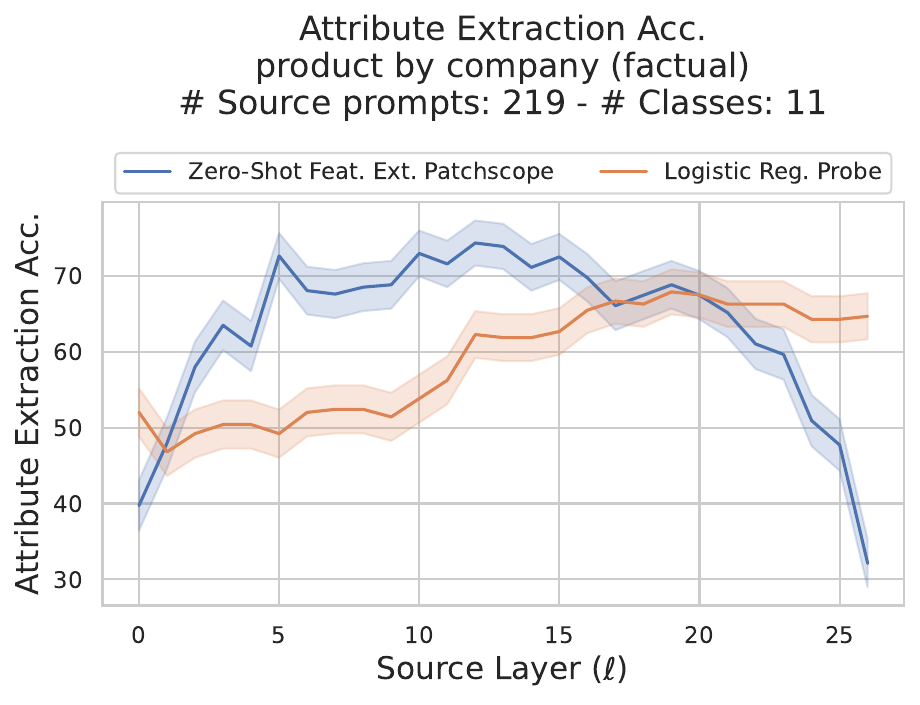}}
        \subfigure[Person Plays Pro Sport]{\includegraphics[width=0.24\textwidth]{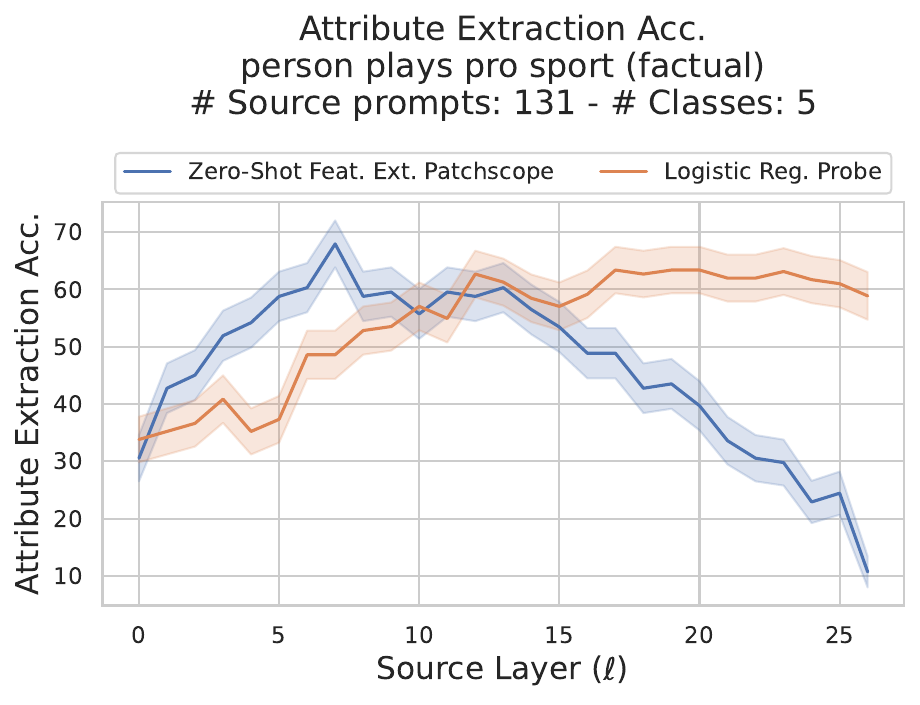}} \\
        \subfigure[Person Plays Position in Sport]{\includegraphics[width=0.24\textwidth]{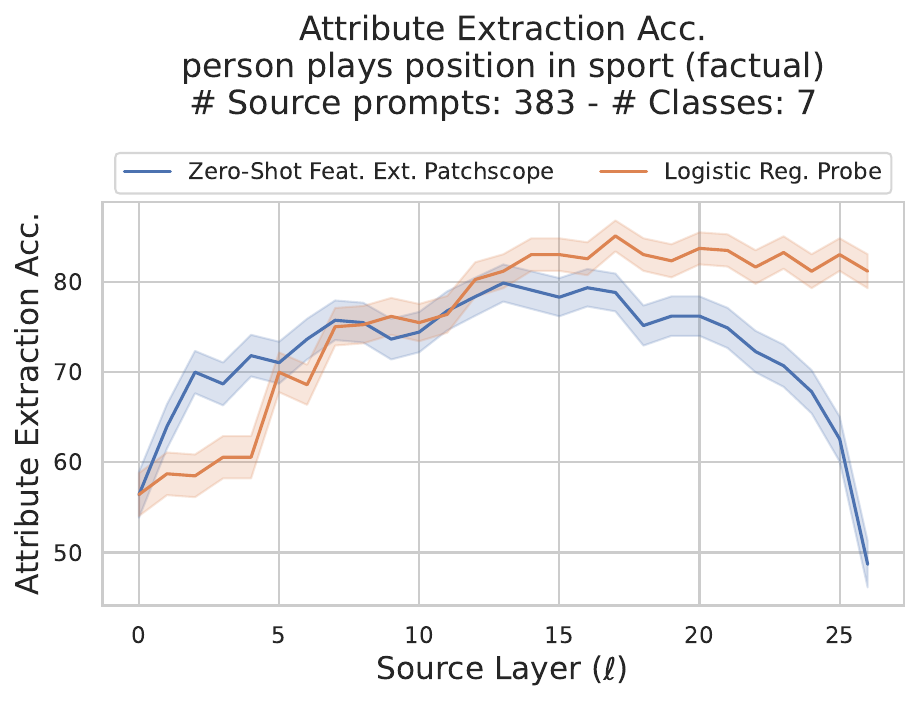}}
        \subfigure[Object Superclass]{\includegraphics[width=0.24\textwidth]{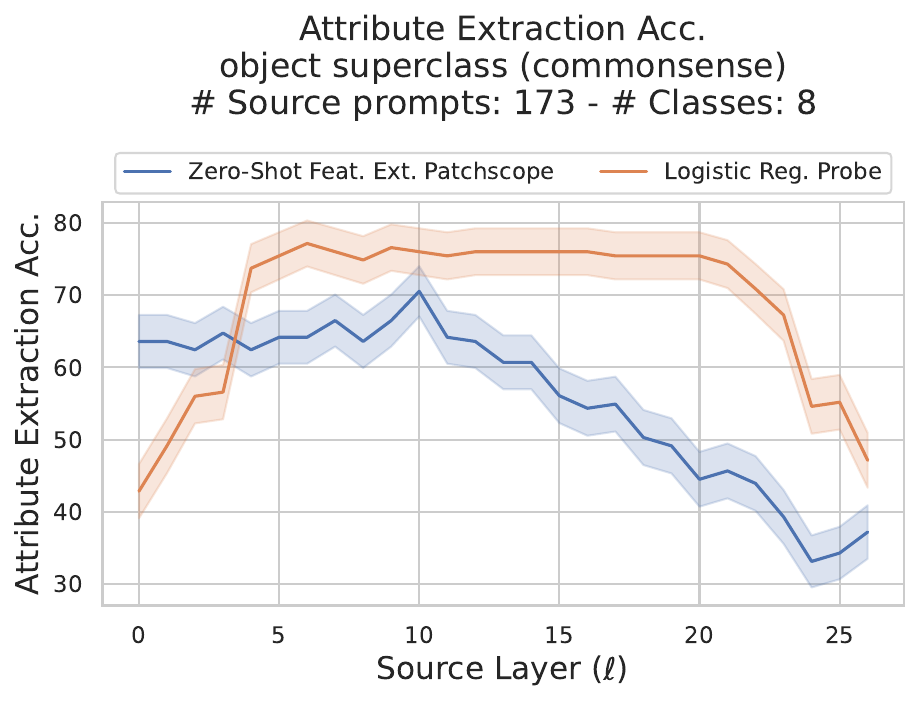}}
        \subfigure[Fruit Inside Color]{\includegraphics[width=0.24\textwidth]{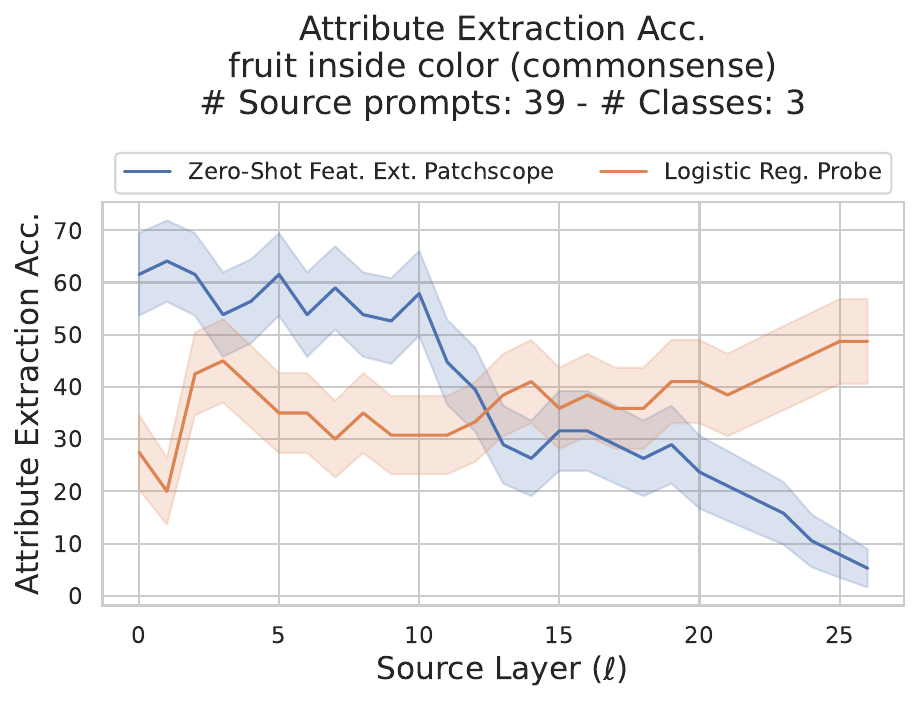}}
        \subfigure[Star Constellation]{\includegraphics[width=0.24\textwidth]{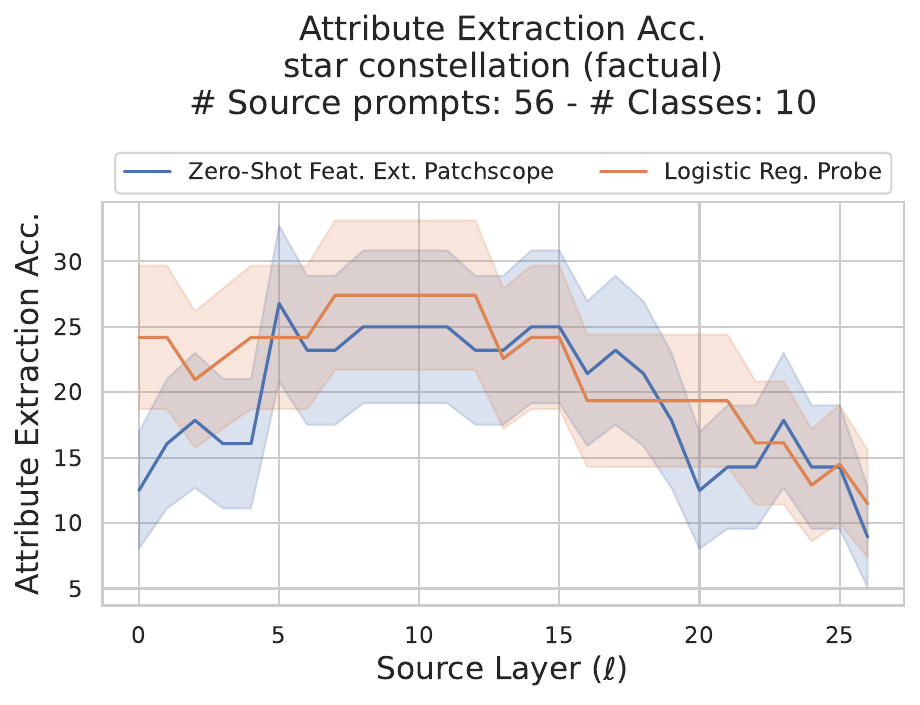}}\\
        \subfigure[Person Father]{\includegraphics[width=0.24\textwidth]{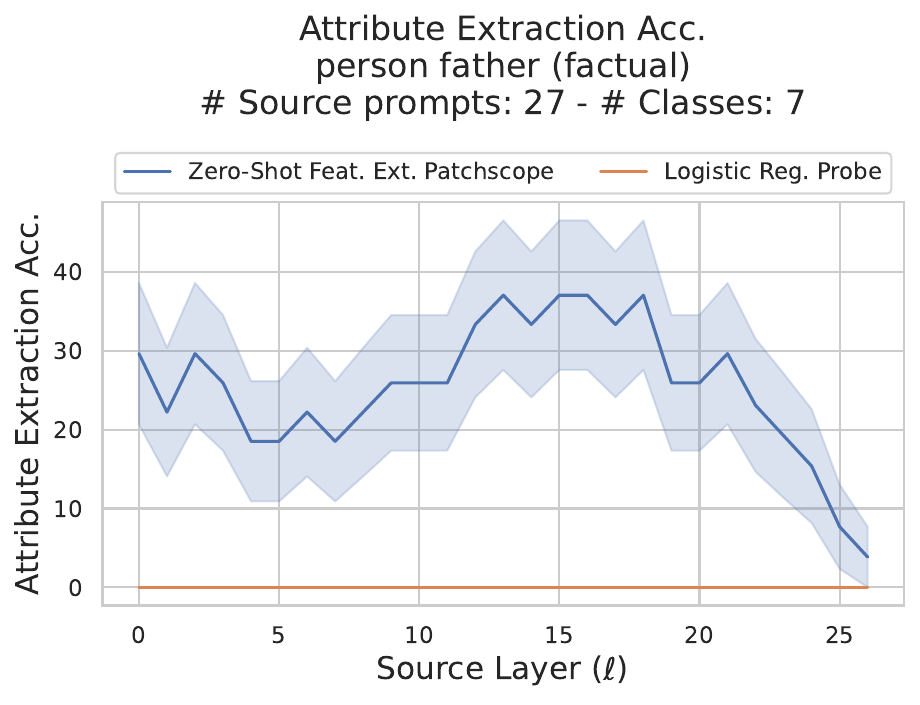}}
        \subfigure[Superhero Archnemesis]{\includegraphics[width=0.24\textwidth]{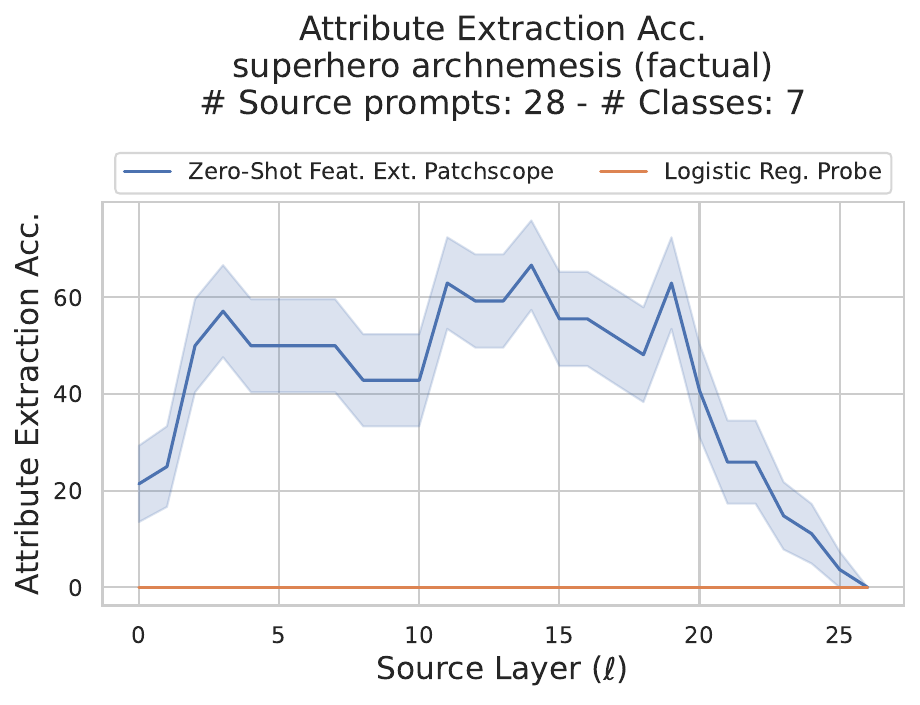}}
        \subfigure[Task Done By Person]{\includegraphics[width=0.24\textwidth]{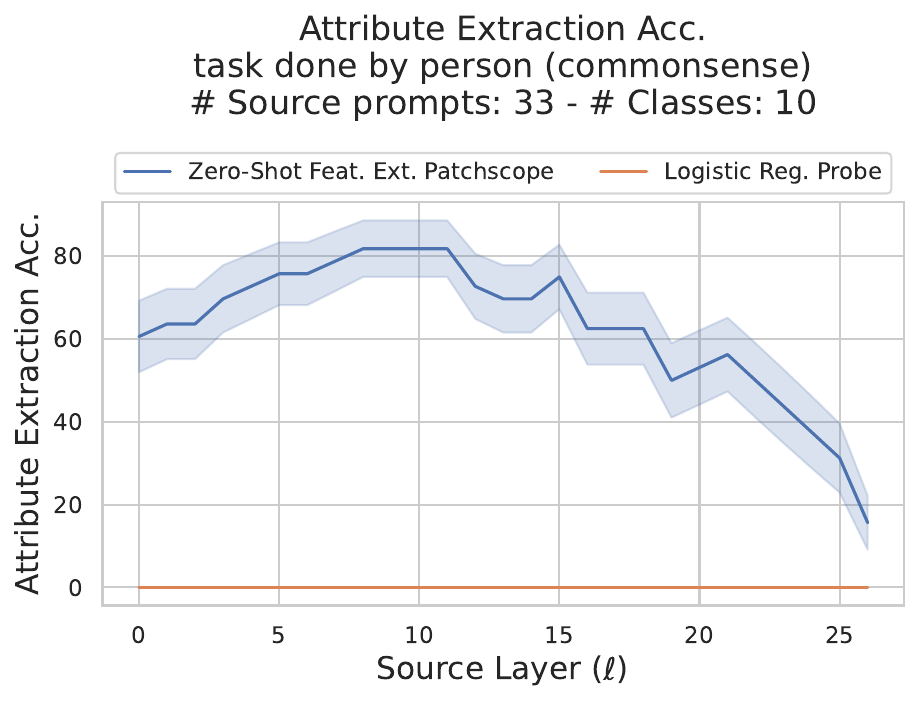}}
        \subfigure[Work Location]{\includegraphics[width=0.24\textwidth]{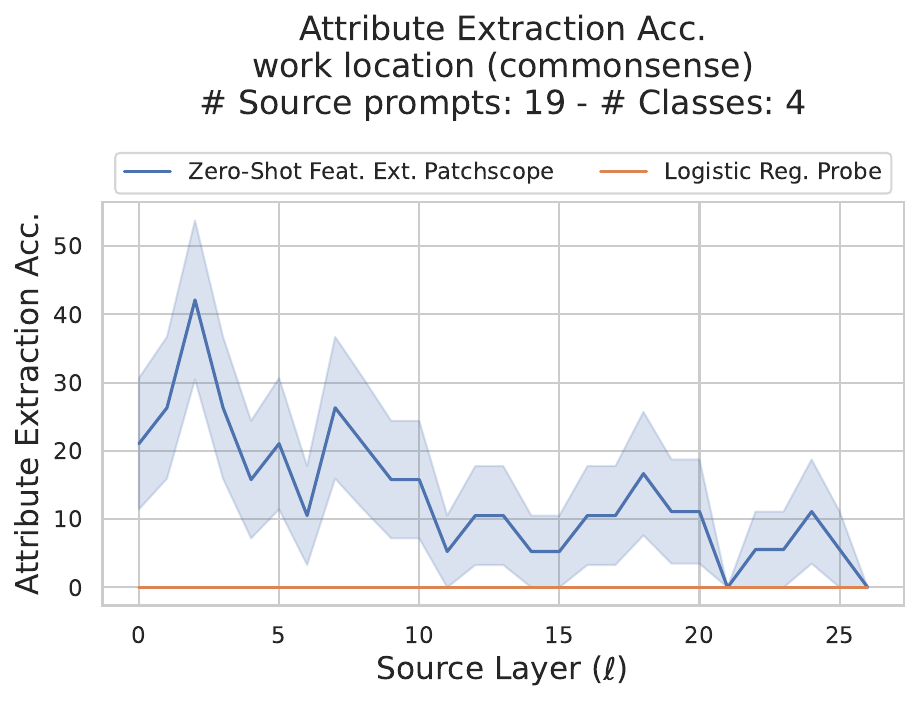}}\\
        \subfigure[Superhero Person]{\includegraphics[width=0.24\textwidth]{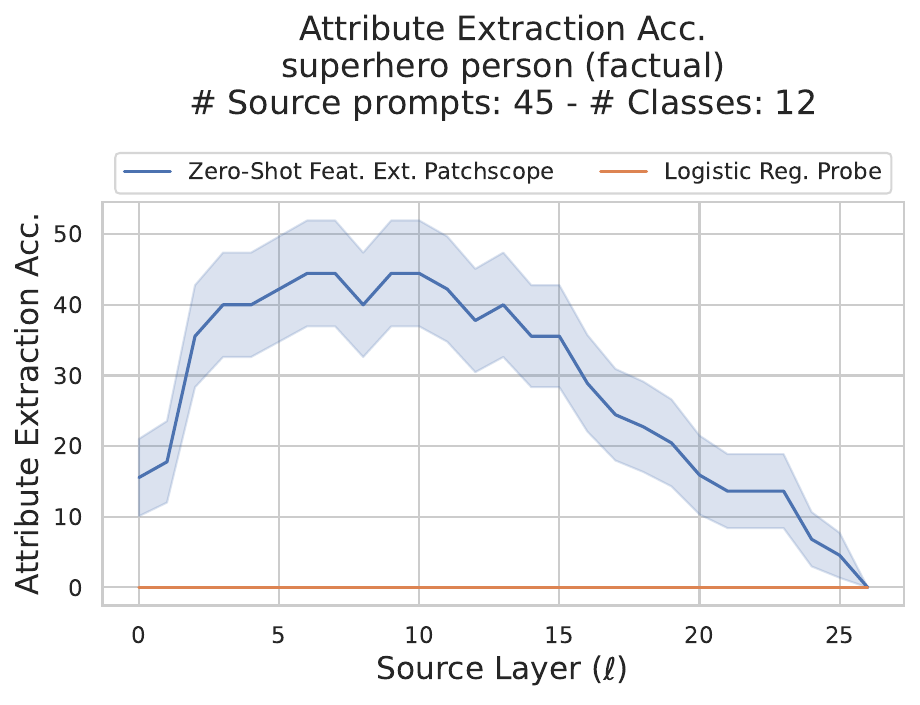}}
        \subfigure[Country Capital City]{\includegraphics[width=0.24\textwidth]{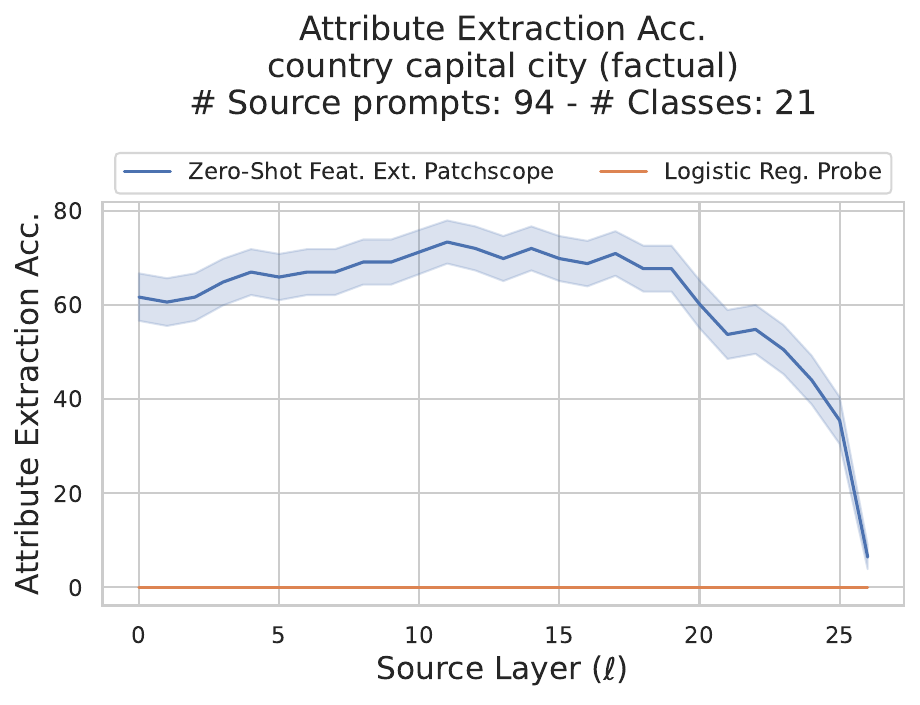}} 
        \subfigure[Country Largest City]{\includegraphics[width=0.24\textwidth]{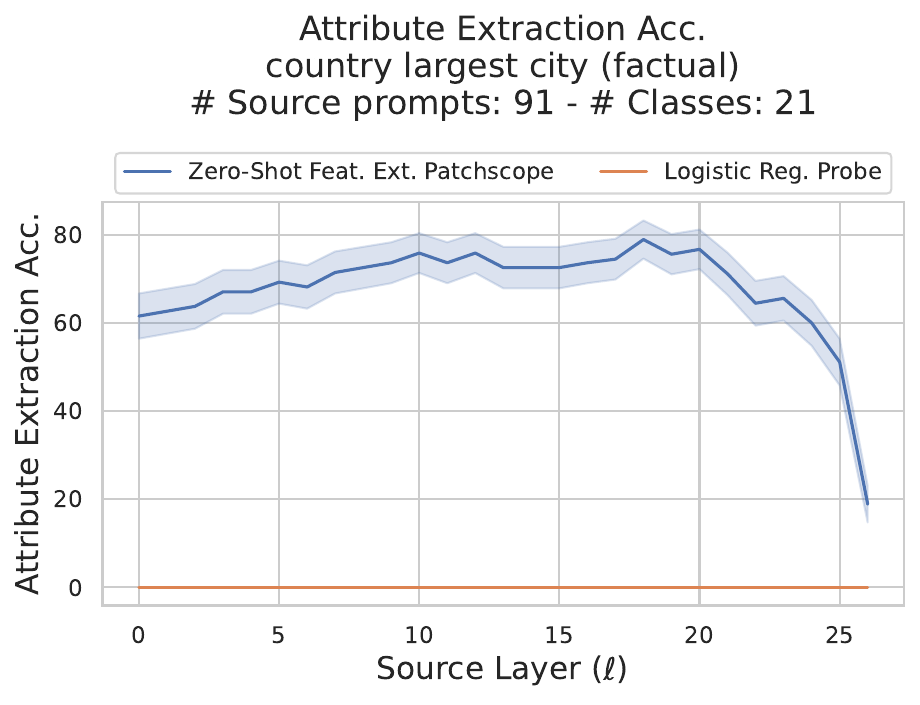}}
        
    \caption{Feature extraction accuracy with respect to source layer ($\sourceLayer$) across various factual and commonsense reasoning tasks. The zero-shot feature extraction \methodName{} works consistently better than the logistic regression probe in early layers, and mostly in mid layers. There is a decline in \methodName{} accuracy in later $\sourceLayer$ as the source representations shift toward next-token prediction.}
    \label{fig:attribute_extraction_lineplots_appendix}
\end{figure*}

\paragraph{Performance Breakdown Across Source Layers}
Fig.~\ref{fig:attribute_extraction_lineplots_appendix} depicts how accuracy changes across source layers $\sourceLayer \in [1, \ldots, \numSourceLayers]$. Considering the early layers, we observe that the \methodName{} consistently outperforms the baseline. This further confirms our hypothesis that prior methods are particularly limited in surfacing information in the early layers, which often cannot be decoded via linear functions.
However, \methodName{} is able to extract such attributes significantly earlier. Note that this observation does not mean the attribute is \textit{explicitly} encoded in a representation, but that there is enough information encoded such that the attribute can be extracted from the representation alone, without its original context, using the model's computation.
In the middle layers, \methodName{} works similarly or better than the baseline.
Interestingly, we observe that almost all cases where \methodName{} performs worse than the baseline occur in later layers. Our interpretation is that given the language modeling training objective, the representations shift toward next-token prediction in the later layers. Therefore, the attribute of interest would not be as readily accessible via the model's computation in these layers.
This interpretation is also aligned with recent findings that show no decline in using linear relational embedding in predicting $\object$ only when the next token also happens to be $\object$ \cite{hernandez2023linearity}. We postulate that when the immediate next token is not the attribute of interest, it does not necessarily mean the attribute information is lost, but rather it may not be as easily accessible on the surface. We hypothesize that using a \methodName{} with a more expressive mapping $\transformation$ could improve attribute extraction accuracy in the later layers, which we leave for future work to verify.

\begin{figure*}
    \centering
        \subfigure[Substance Phase]{\includegraphics[width=0.24\textwidth]{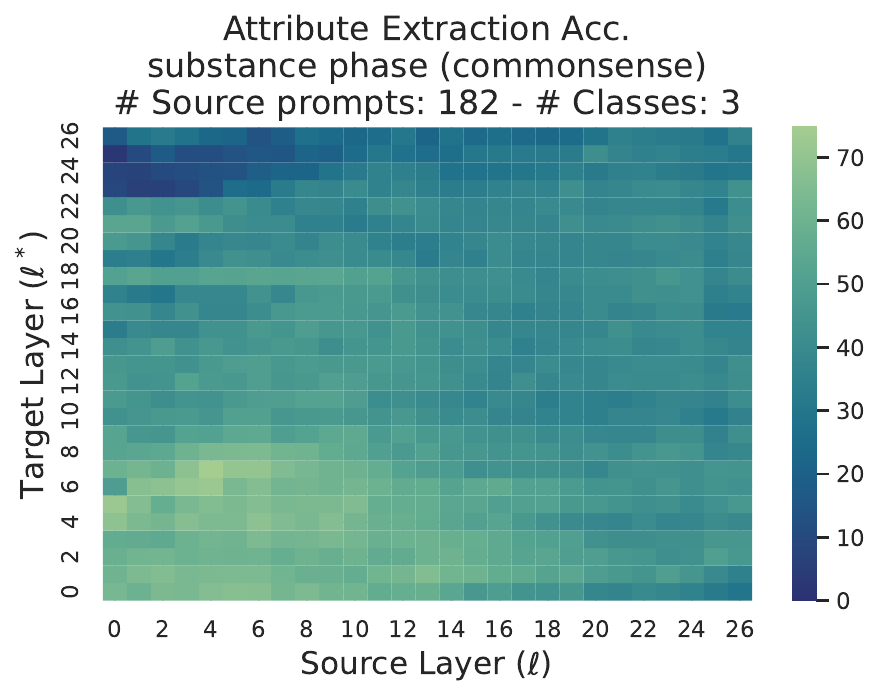}}
        \subfigure[Task Done By Tool]{\includegraphics[width=0.24\textwidth]{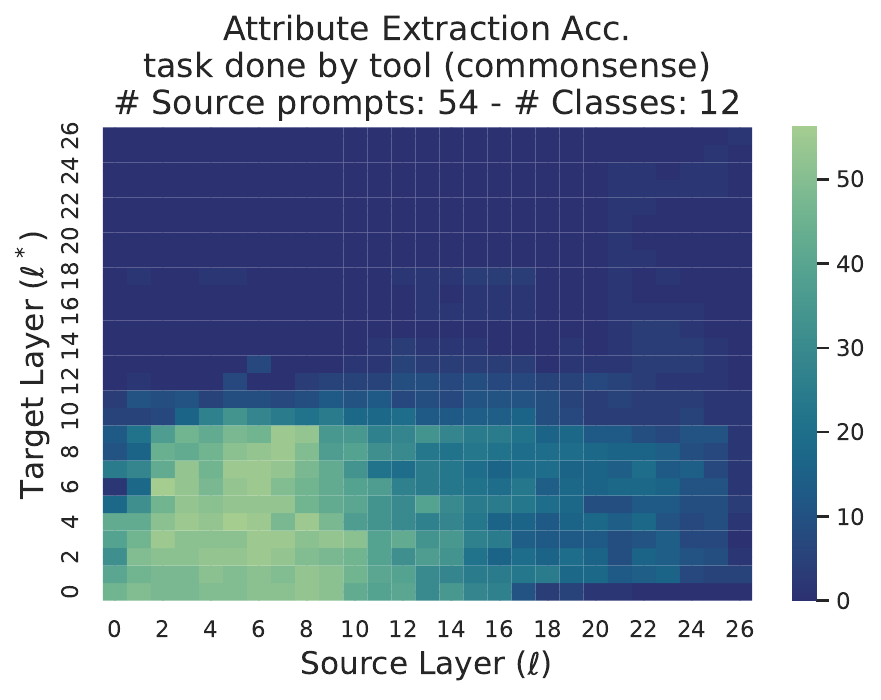}} 
        \subfigure[Country Currency]{\includegraphics[width=0.24\textwidth]{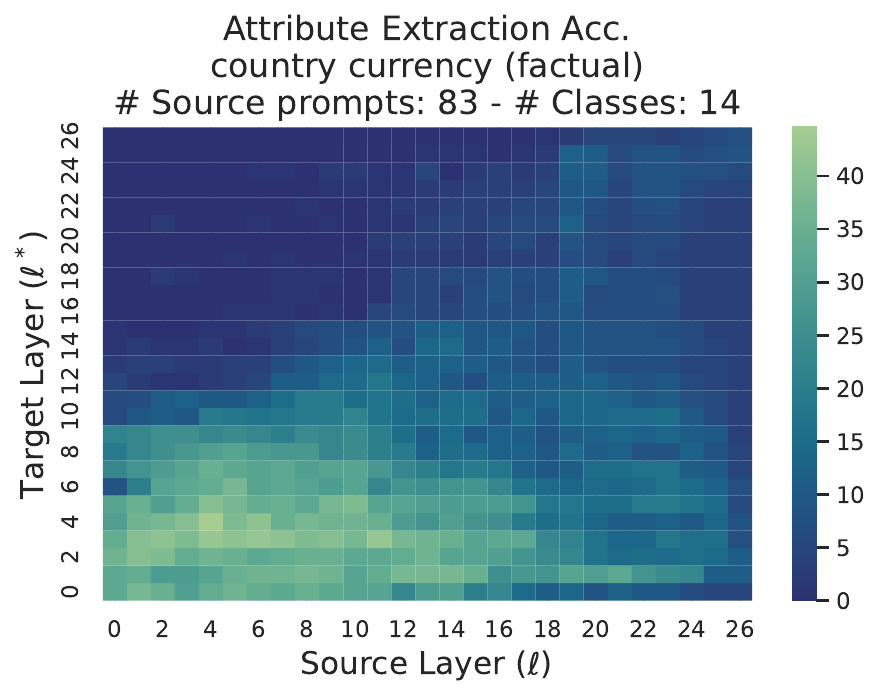}} 
        \subfigure[Food From Country]{\includegraphics[width=0.24\textwidth]{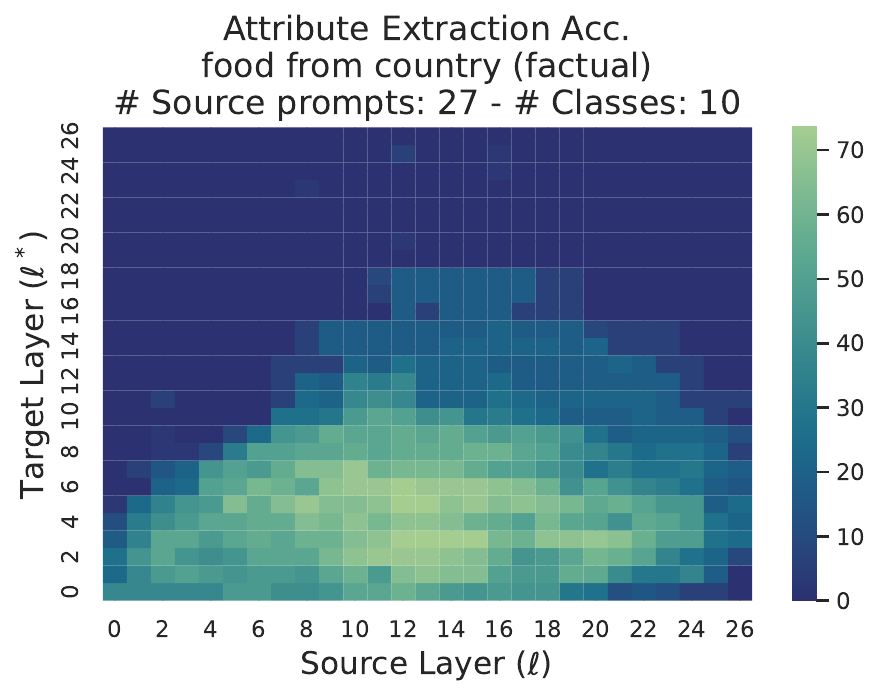}} \\
        \subfigure[Company CEO]{\includegraphics[width=0.24\textwidth]{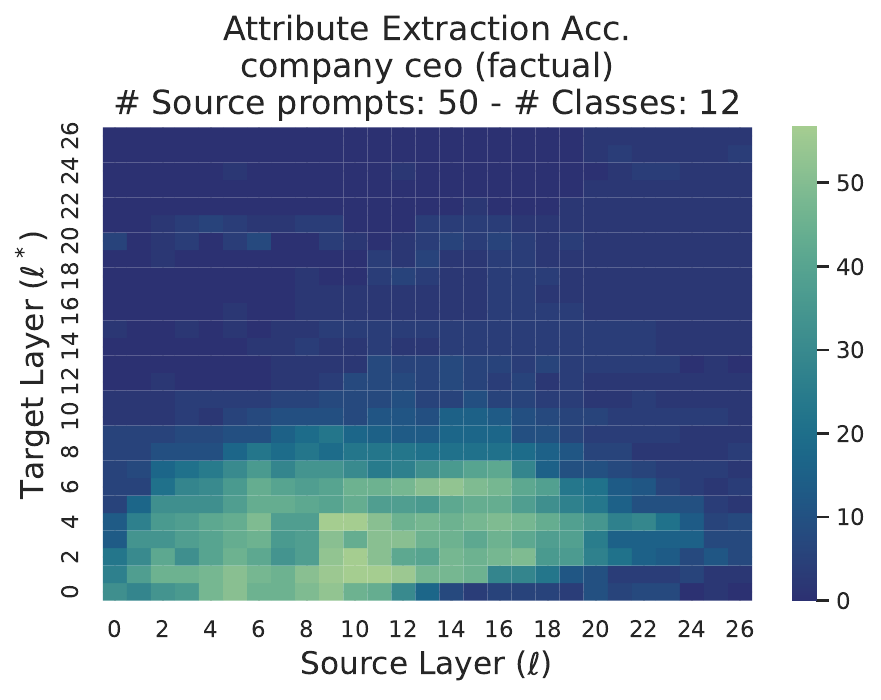}}
        \subfigure[Fruit Outside Color]{\includegraphics[width=0.24\textwidth]{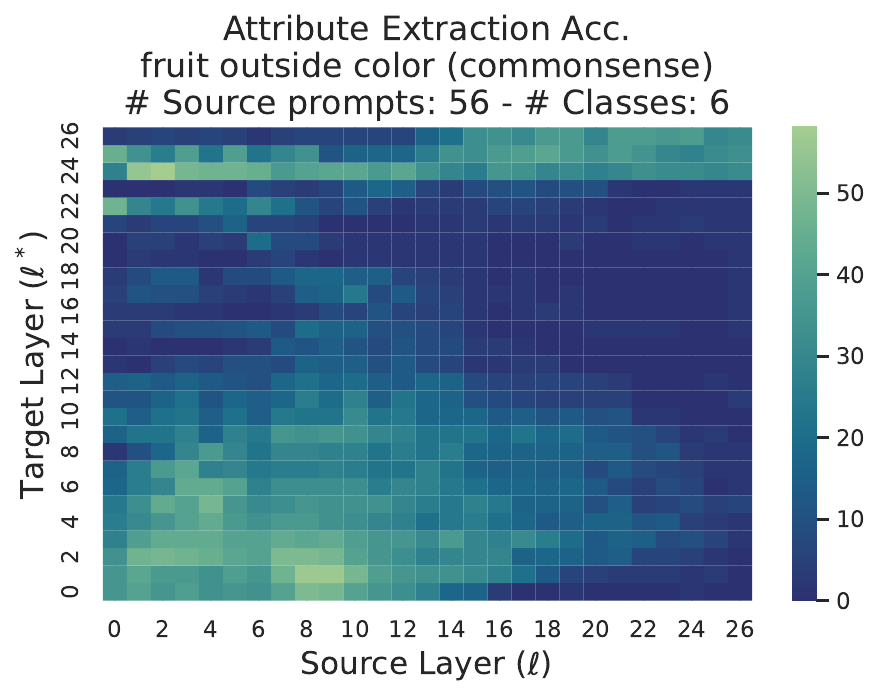}}
        \subfigure[Product By Company]{\includegraphics[width=0.24\textwidth]{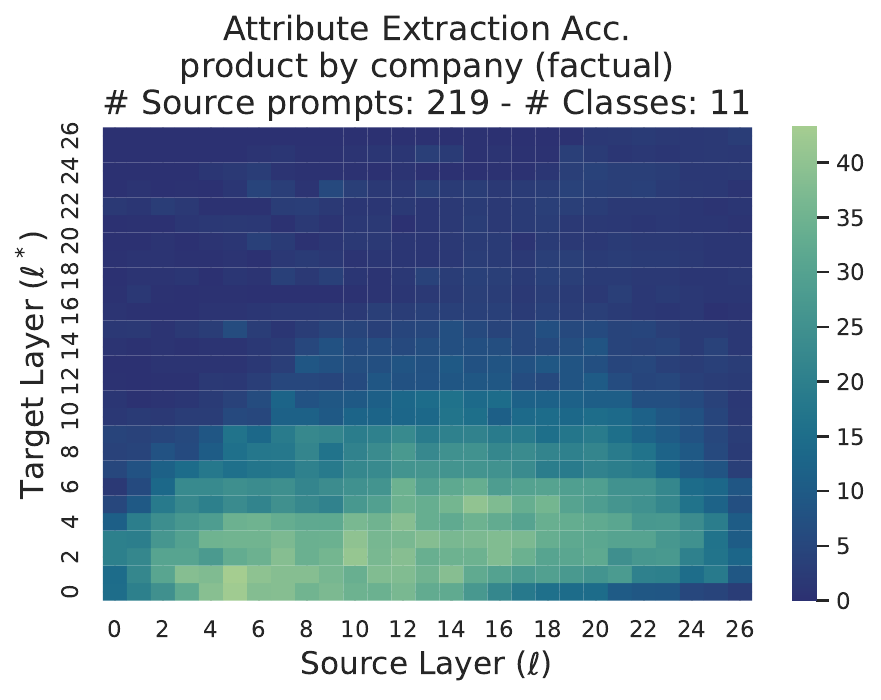}}
        \subfigure[Person Plays Pro Sport]{\includegraphics[width=0.24\textwidth]{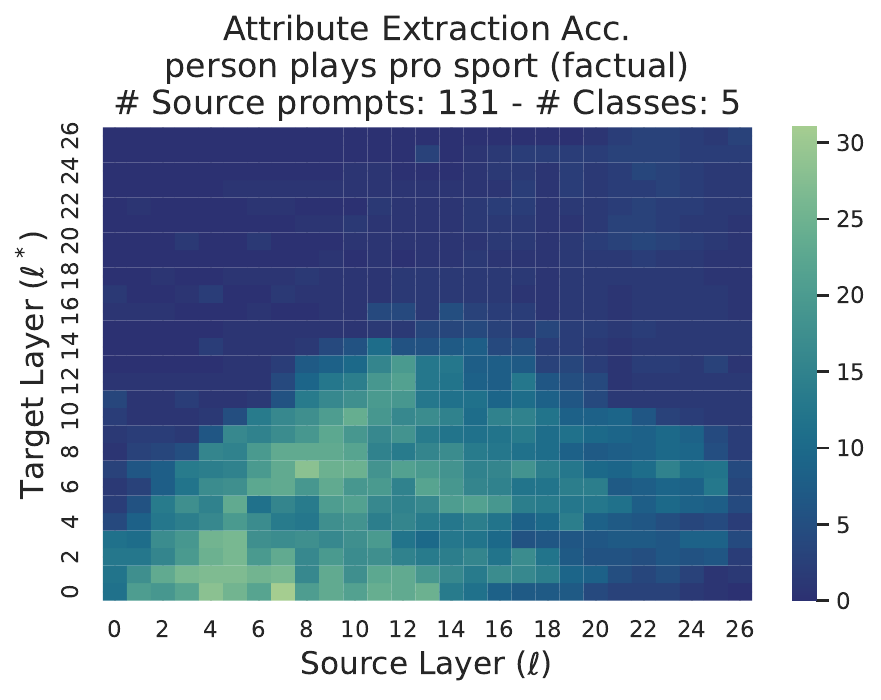}} \\
        \subfigure[Person Plays Position in Sport]{\includegraphics[width=0.24\textwidth]{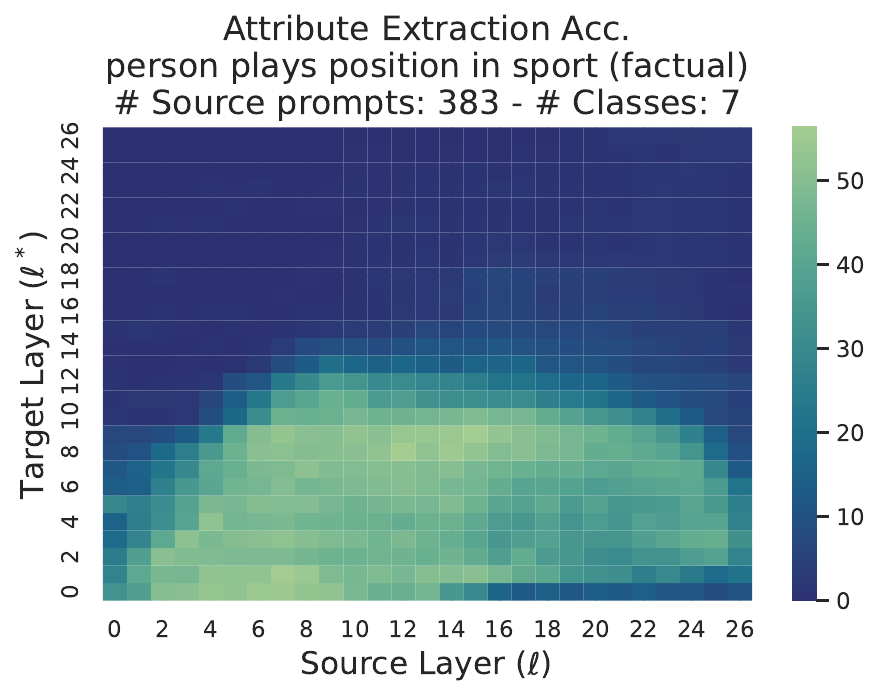}}
        \subfigure[Object Superclass]{\includegraphics[width=0.24\textwidth]{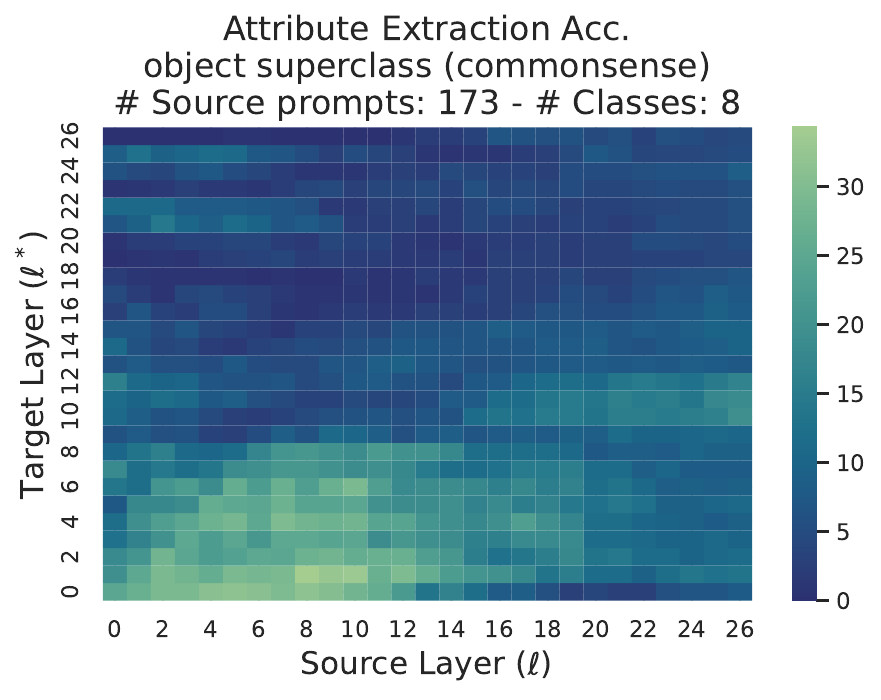}}
        \subfigure[Fruit Inside Color]{\includegraphics[width=0.24\textwidth]{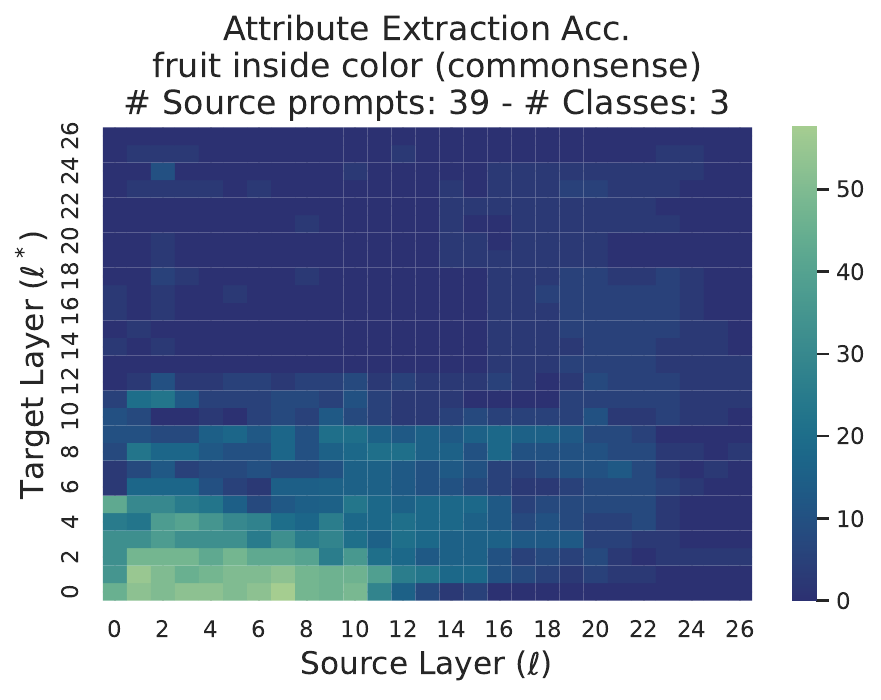}}
        \subfigure[Star Constellation]{\includegraphics[width=0.24\textwidth]{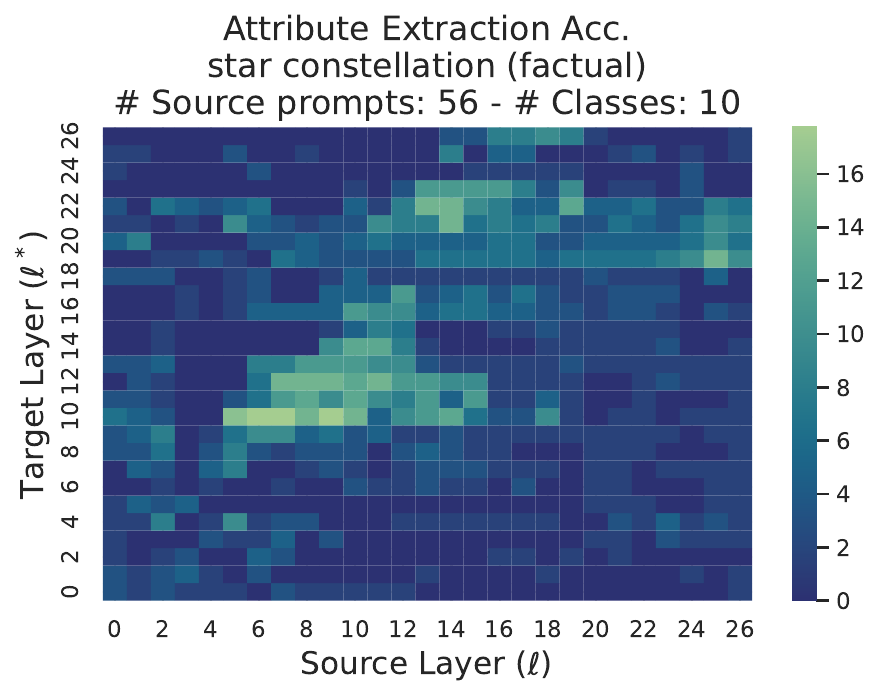}}\\
        \subfigure[Person Father]{\includegraphics[width=0.24\textwidth]{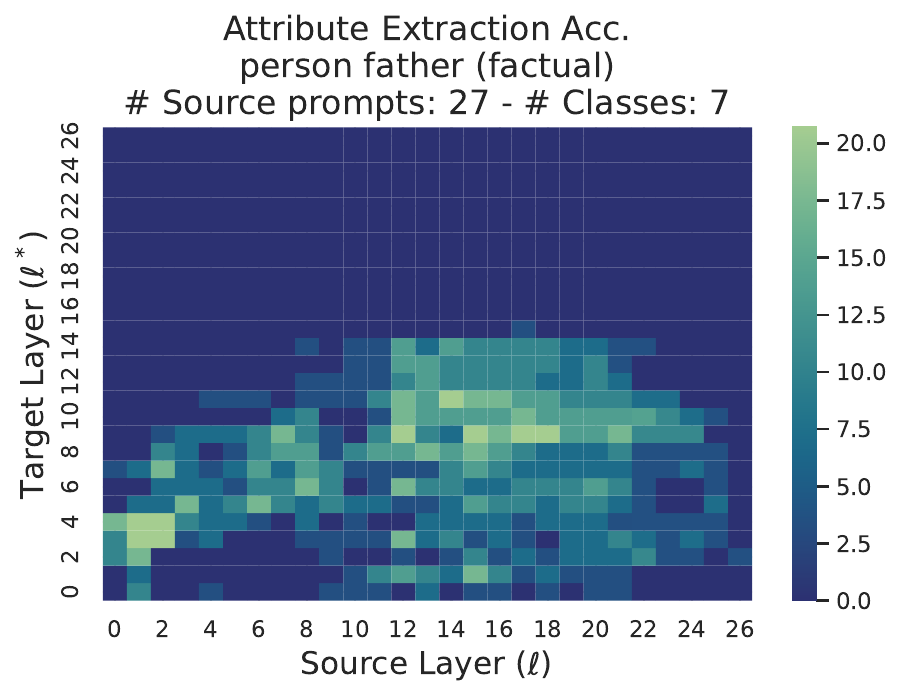}}
        \subfigure[Superhero Archnemesis]{\includegraphics[width=0.24\textwidth]{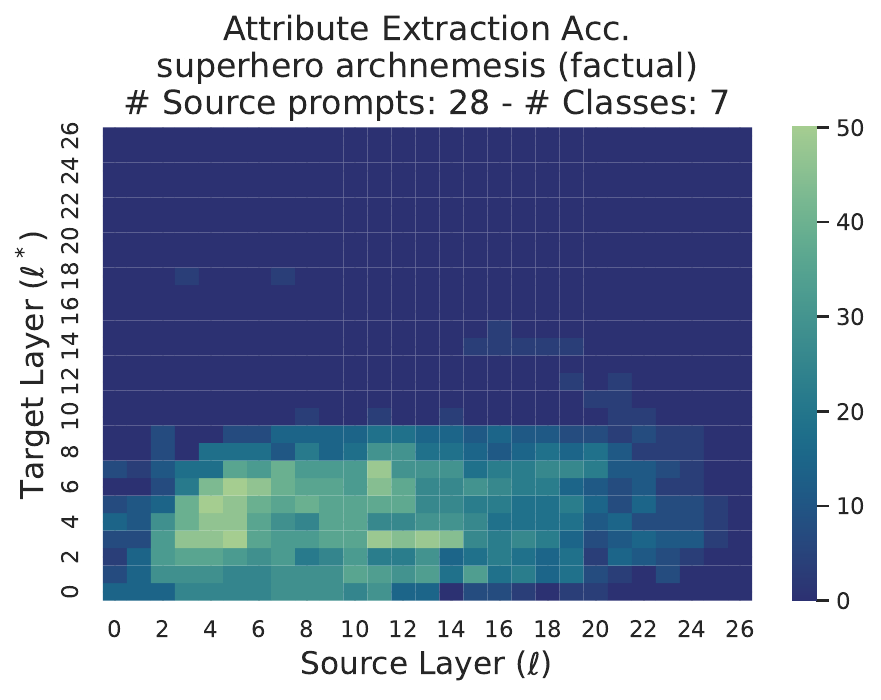}}
        \subfigure[Task Done By Person]{\includegraphics[width=0.24\textwidth]{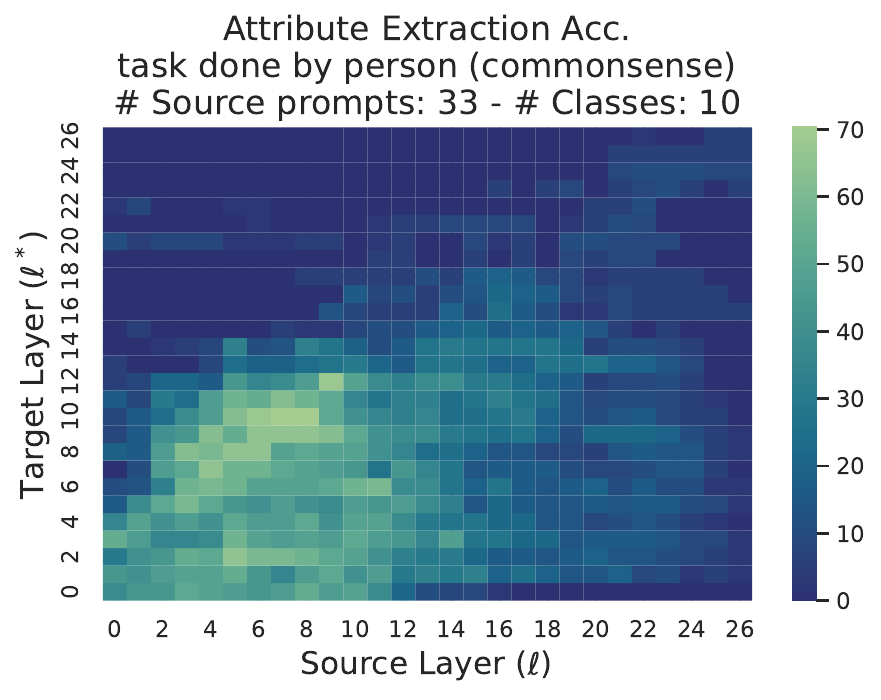}}
        \subfigure[Work Location]{\includegraphics[width=0.24\textwidth]{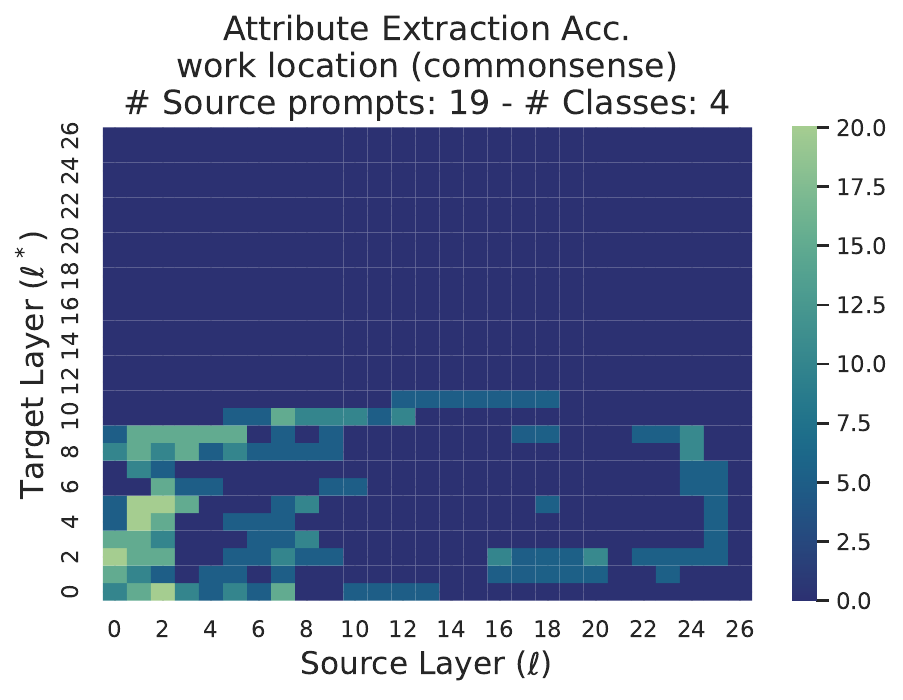}}\\
        \subfigure[Superhero Person]{\includegraphics[width=0.24\textwidth]{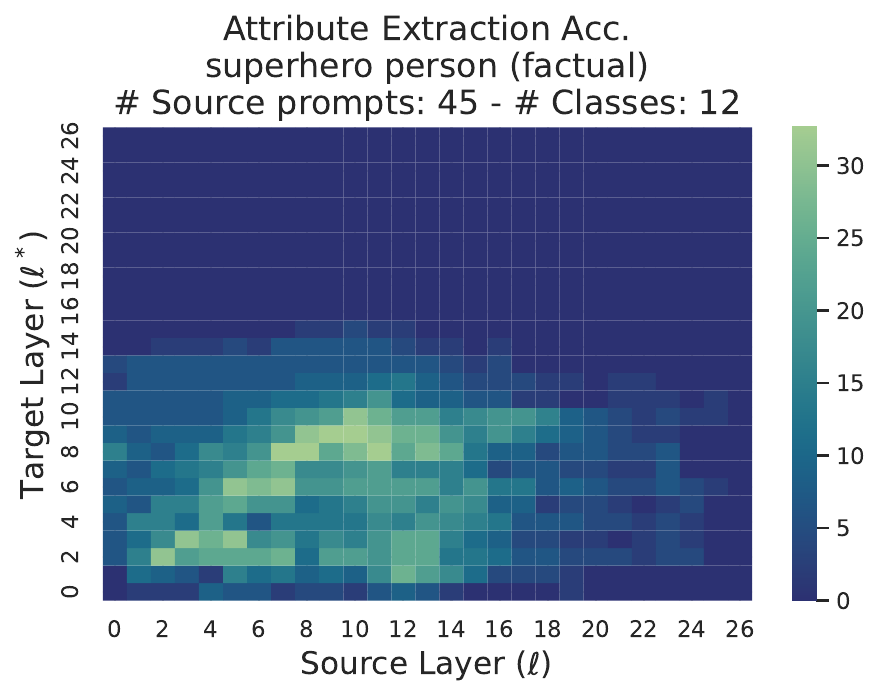}}
        \subfigure[Country Capital City]{\includegraphics[width=0.24\textwidth]{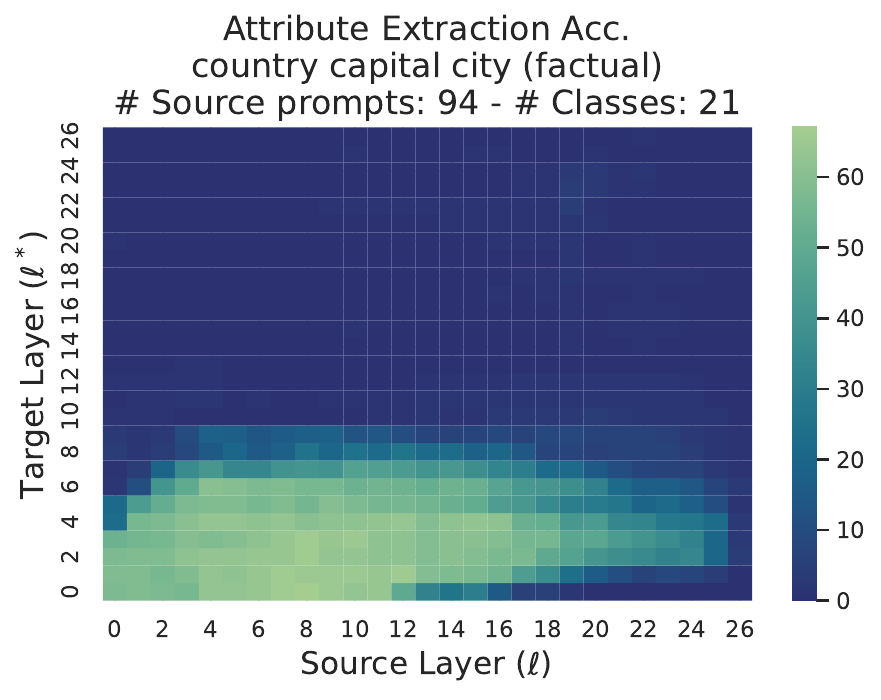}} 
        \subfigure[Country Largest City]{\includegraphics[width=0.24\textwidth]{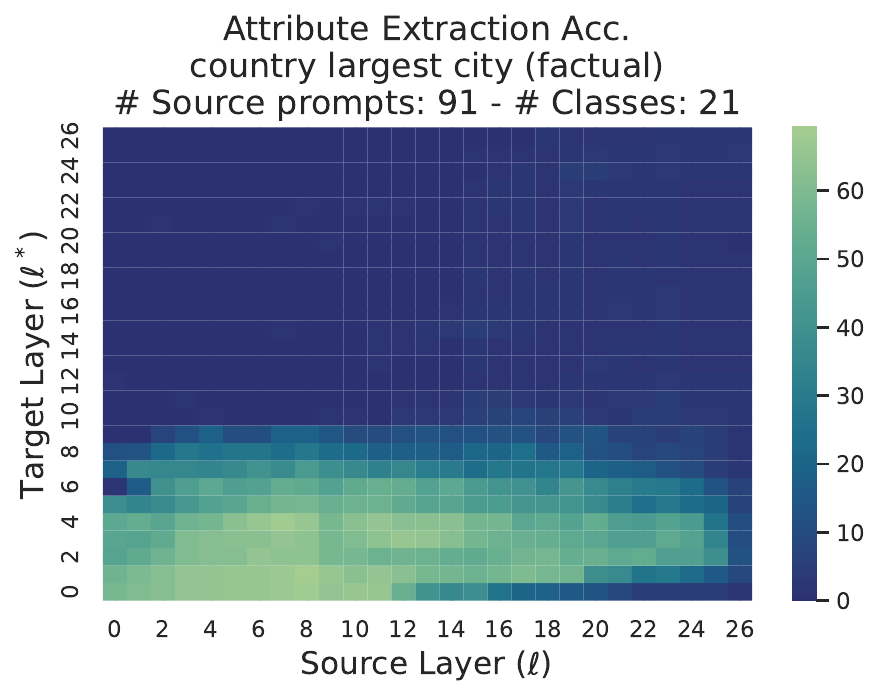}}
        
    \caption{The interaction between source and target layers in the zero-shot feature extraction \methodName{} across various factual and commonsense reasoning tasks. Each cell $(\sourceLayer, \targetLayer)$ in the heatmap shows the attribute extraction success rate where source and target layers are fixed to $\sourceLayer$ and $\targetLayer$, respectively. Particularly, there is a higher success rate in the lower left quadrants, representing early to mid source and target layer combinations. The right half of the heatmaps shows late source layers, where the source representation has shifted toward next-token prediction, leading to a lower success rate in attribute extraction. The top half of the heatmaps shows late target layers. When the accuracy is a function of more than a single next-token, the placeholder token representation still remains in the early layers, leading to a lower attribute extraction rate.}
    \label{fig:attribute_extraction_heatmaps_appendix}
\end{figure*}
\paragraph{Source-Target Layer Interplay}
Fig.~\ref{fig:attribute_extraction_heatmaps_appendix} visualizes the interaction between $\sourceLayer$ and $\targetLayer$. These heatmaps show attribute extraction success rate for the zero-shot feature extraction \methodName{} for a fixed $(\sourceLayer, \targetLayer)$ combination. The lower left quadrants show setups where both  $\sourceLayer$ and $\targetLayer$ represent early to middle layers, and the success rate is maximal. The right half of the heatmaps represents late $\sourceLayer$, which achieves lower success rate due to token representations shifting toward next-token prediction as discussed earlier. In addition, we notice lower success rate in the top half of the heatmaps which represents late $\targetLayer$. 
It is worth noting that in this task, the accuracy is not only based on the immediate next-token prediction, but rather whether $\object$ appears in the next 20 autoregressively generated tokens. The placeholder token ``\texttt{x}'' does still remain in the input, and its representation persists in the early layers in the target computation. This explains why a lower attribute extraction rate is observed in later $\targetLayer$ values. We leave it to the future work to investigate adaptations to \methodFamilyName{} to control contamination from the placeholder tokens and make them more amenable to late $\targetLayer$ choices.

\paragraph{Statistical Test Details} The null hypothesis is: ``There is no significant difference between probing accuracy and patchscope accuracy''. We calculate attribute extraction accuracy per layer, in a 40-layer model, across various tasks. Since we run a test for each task independently (see Tab.~\ref{tab:attribute_extraction}), we use the conservative Bonferroni correction to address the multiple comparison problem. That is, for a desired overall $\alpha=0.05$, we consider $\alpha/\text{number\,of\,tasks}=\frac{0.05}{12}=0.004$ for each individual task.

\section{Additional Information and Results on the Entity Resolution Experiment}
\begin{figure}[t!]
    \centering
    \subfigure[Vicuna]{\includegraphics[width=0.45\textwidth]{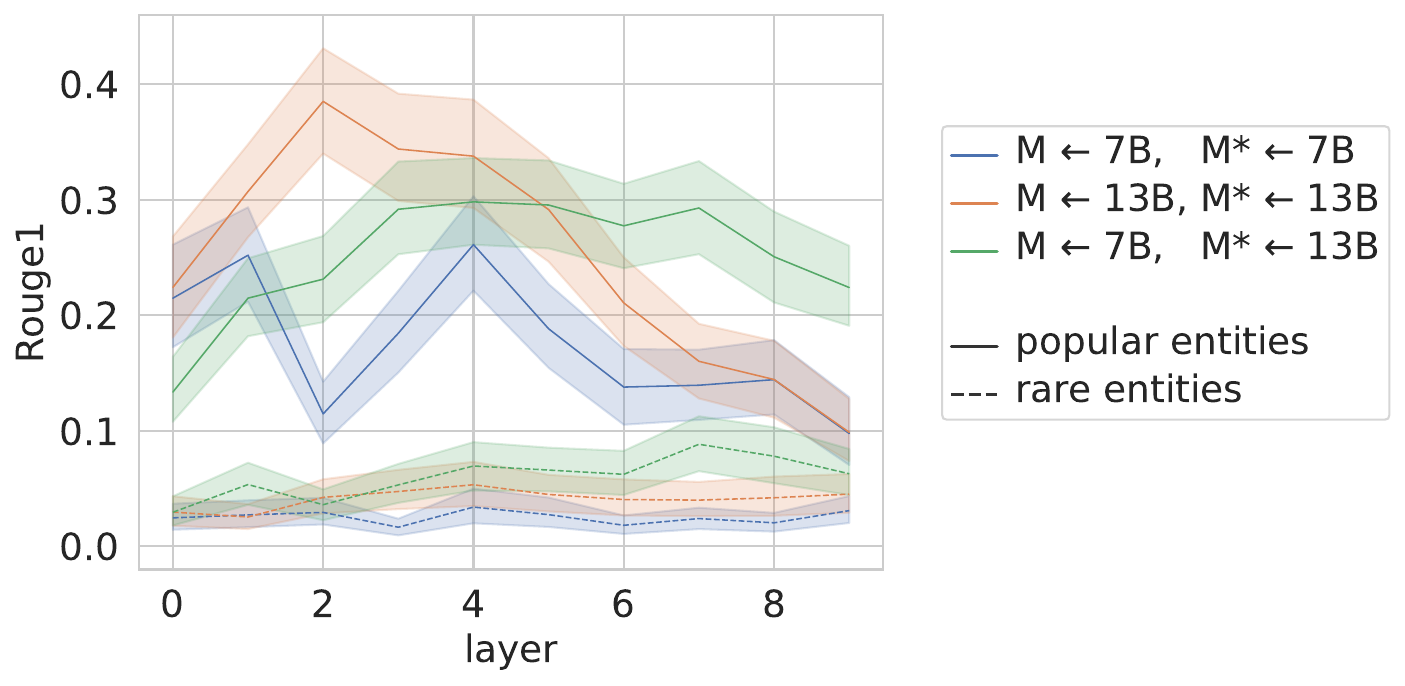}}
    \subfigure[Pythia]{\includegraphics[width=0.45\textwidth]{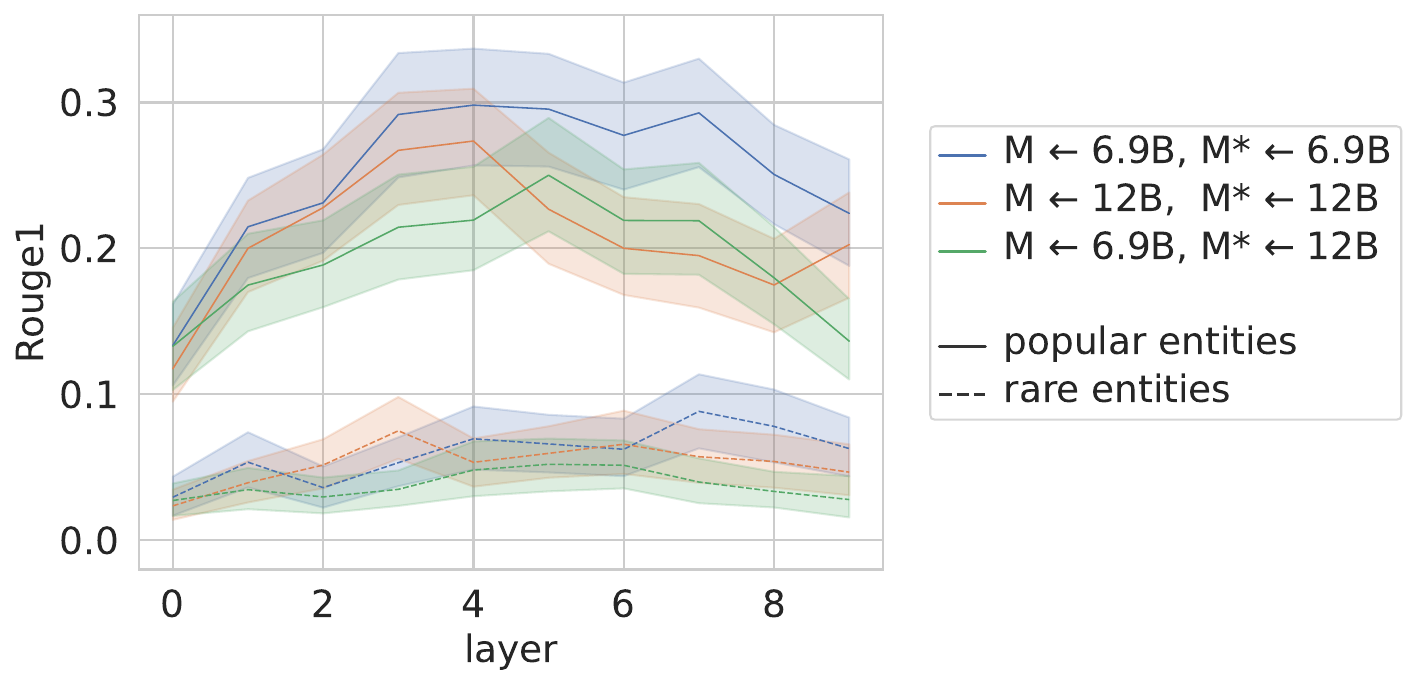}}
\vspace{-1.2em}
\caption{Rouge1 scores of the generated descriptions against descriptions from Wikipedia.}
\vspace{-1em}
\label{fig:entity_processing_rouge1}
\end{figure}
\begin{figure}[t!]
    \centering
    \subfigure[Vicuna]{\includegraphics[width=0.45\textwidth]{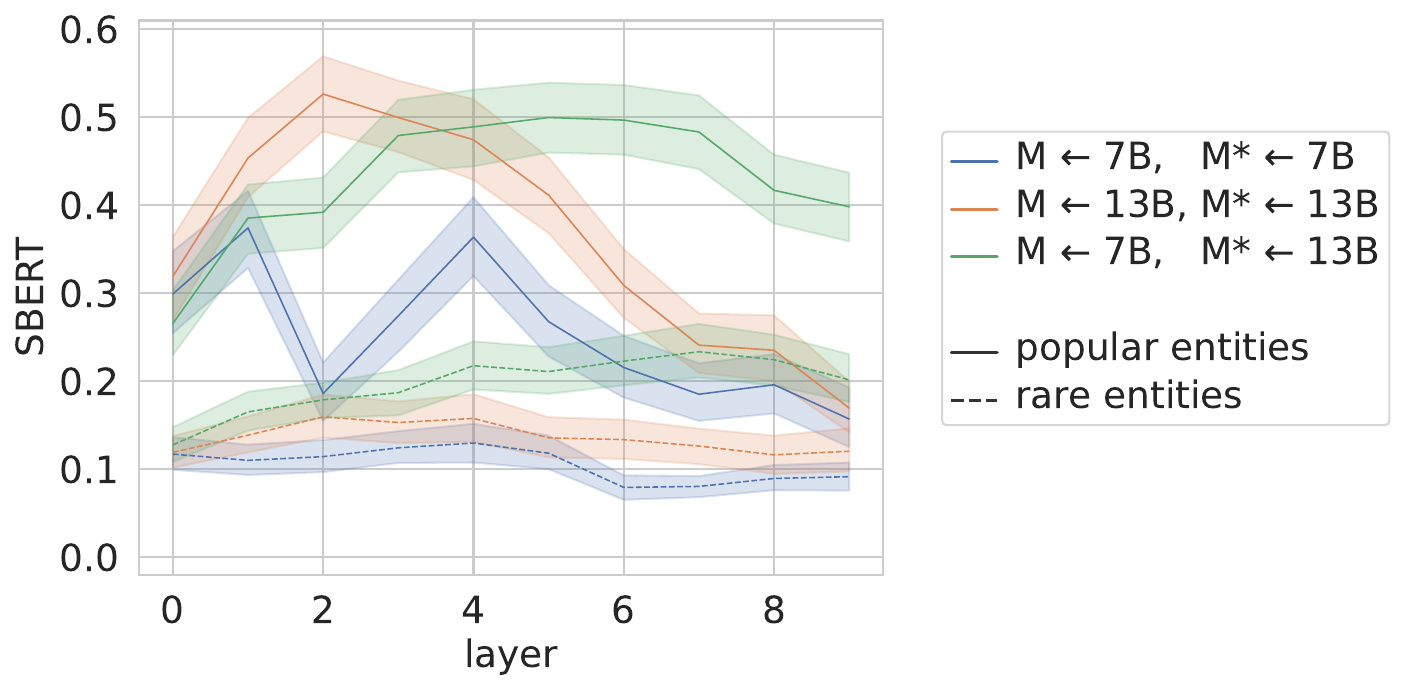}}
    \subfigure[Pythia]{\includegraphics[width=0.45\textwidth]{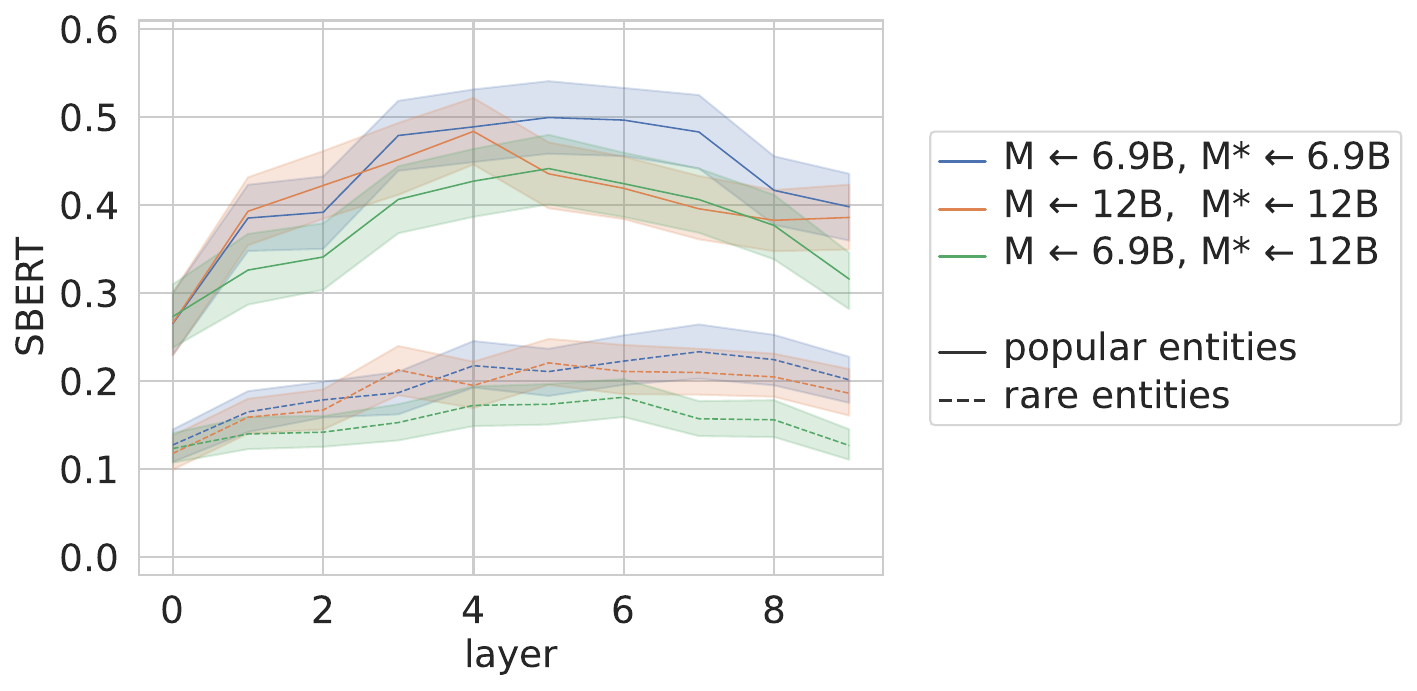}}
\vspace{-1.2em}
\caption{SBERT scores of the generated descriptions against descriptions from Wikipedia.}
\label{fig:entity_processing_sbert}
\vspace{-1em}
\end{figure}

\subsection{Experimental Setup}
\label{sec:appendix-entity-resolution-info}
Recall that we use a few-shot target prompt template for decoding an entity description: $``\texttt{subject}_1\;: \texttt{description}_1\texttt{, subject}_{2}: \texttt{description}_2 \texttt{, \ldots, subject}_{k}: \texttt{description}_k \texttt{, x}"$, while patching the last position which corresponds to ``\texttt{x}''.  Specifically, we use the following target prompt, obtained randomly: ``\texttt{Syria: Country in the Middle East, Leonardo DiCaprio: American actor, Samsung: South Korean multinational major appliance and consumer electronics corporation, x}'' and task the model to generate the completion after the patched representation in ``\texttt{x}''. For the subject description, which is composed of $k=3$ random subject entities, we used the wptools\footnote{\url{https://github.com/siznax/wptools/}} python package for obtaining a  description of every subject entity from Wikipedia.

\subsection{Additional Quantitative Results}
\label{sec:appendix-entity-resolution}

In this section, we present the Rouge1 \cite{lin-2004-rouge} and SBERT score \citep{reimers-2019-sentence-bert} results, as well as the results for the Pythia models \citep{pythia}.\footnote{We used the package from \url{https://www.sbert.net/}, with the sentence-transformers/all-MiniLM-L6-v2 model.} In Fig.~\ref{fig:entity_processing_rouge1} and Fig.~\ref{fig:entity_processing_sbert}, we present the Rouge1 and the SBERT results, respectively, complementary to the RougeL results in Fig.~\ref{fig:entity_processing} from \S\ref{sec:input_processing}.  Note that for Pythia, the smaller model (6.9B) outperforms the larger one (12B), and hence, our cross-model patching method would not be able to improve the inspection of the smaller model, unlike the trends we observed for Vicuna.

\subsection{Additional Qualitative Results}
\label{sec:appendix-entity-resolution-qualitative}

In this section, we provide more examples and discuss our observations about the gradual process of entity resolution. As shown in Tab.~\ref{tab:entity-resolution-vicuna-samples} and Tab.~\ref{tab:entity-resolution-pythia-samples}, it is interesting to observe that the resolution process for the same input can look different across models, suggesting they assign different likelihoods to different entities, and weigh context differently. For example, as Vicuna 13B processes ``\texttt{Will Smith}'', it goes from \textit{``Smithsonian Museum''} to the \textit{``Smith rock band''} to the actor and rapper, \textit{``Will Smith''}. However, Pythia 12B starts with \textit{``Smith \& Wesson weapon manufacturing company''} before it resolves the entity as the American actor, \textit{``Will Smith''}. 

We also observe another phenomenon which we refer to as placeholder contamination. It is the case where the remaining representation of the  placeholder entity ``\texttt{x}'' in the early layers interferes with the model's capability in generating descriptions for the patched token. For example, see Vicuna 13B response to ``\texttt{Paris Hilton}'' entity in Tab.~\ref{tab:entity-resolution-vicuna-samples}. First, we see the gradual process of going from \textit{``Rigatoni''} pasta, to \textit{``Hilton Hotels''}, to the socialite \textit{``Paris Hilton''} in layers 1-6. But in layers 7 and onward, the generation seems to describe the placeholder token ``\texttt{x}''rather than the ``\texttt{Paris Hilton}'' entity: \textit{``Placeholder for a variable or concept''}, \textit{``Variable representing any number of things or concepts''} or \textit{``x is a placeholder''}. For future, we would like to quantify these qualitative observations, study to what extend this contamination can be mitigated with a different placeholder choice, and why some models might be more susceptible to this contamination than others.

\begin{figure}[t!]
    \centering
    \resizebox{.7\columnwidth}{!}{%
    \subfigure[\scriptsize{Vicuna: $\sourceModel\!\is\!\text{7B}, \targetModel\!\is\!\text{13B}$}]{\includegraphics[width=0.396\textwidth,trim={0 0 1.7cm 0},clip]{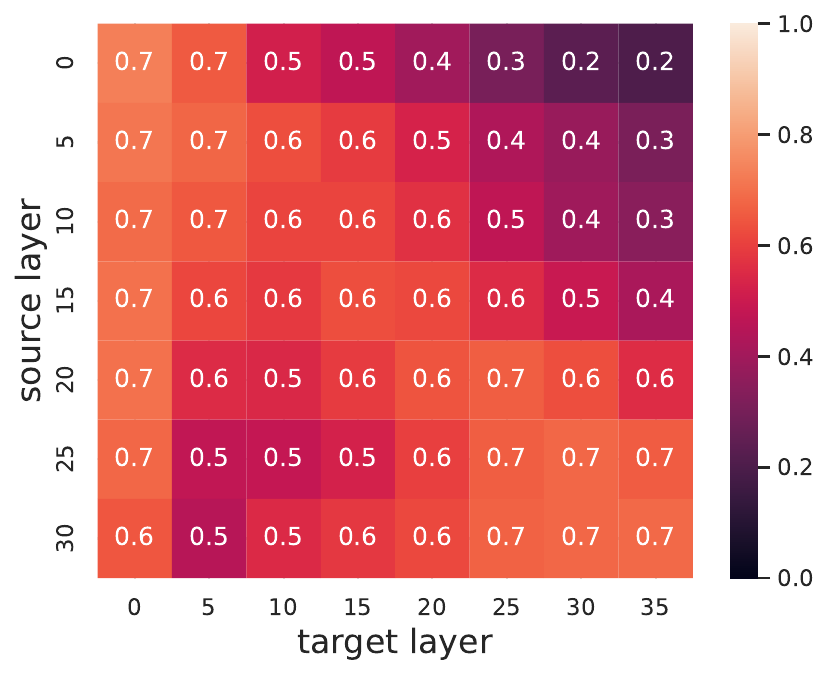}}
    \subfigure[\scriptsize{Pythia: $\sourceModel\!\is\!\text{6.9B}, \targetModel\!\is\!\text{12B}$}]{\includegraphics[width=0.45\textwidth]{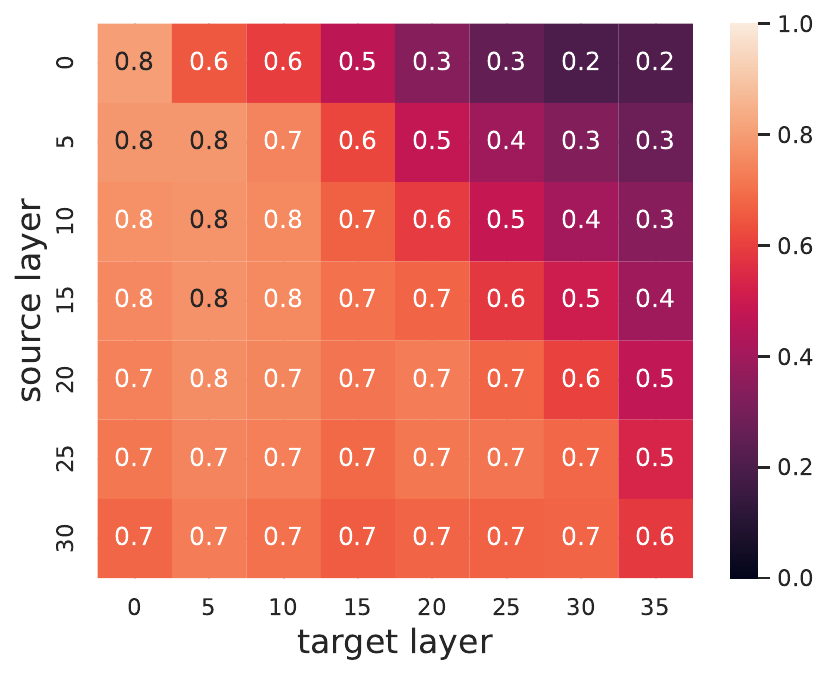}}
    }
\caption{Precision@1 scores ($\uparrow$ is better) on next-token prediction estimation in Vicuna and Pythia with cross-model \methodFamilyName{}}
\label{fig:cross_model_ntp}
\end{figure}
\begin{figure}[t!]
    \centering
    \resizebox{.7\columnwidth}{!}{%
    \subfigure[\scriptsize{Vicuna: $\sourceModel\!\is\!\text{7B}, \targetModel\!\is\!\text{13B}$}]{\includegraphics[width=0.396\textwidth,trim={0 0 1.7cm 0},clip]{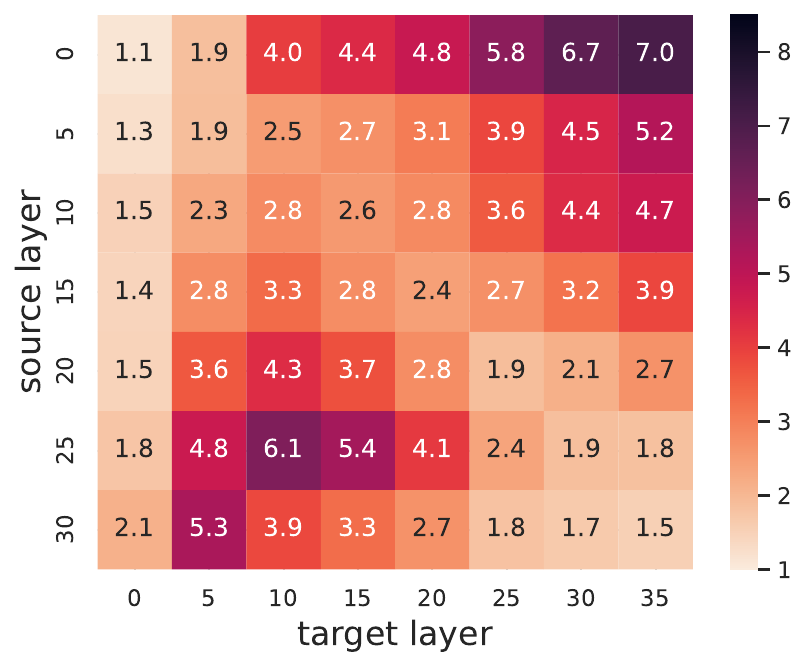}}
    \subfigure[\scriptsize{Pythia: $\sourceModel\!\is\!\text{6.9B}, \targetModel\!\is\!\text{12B}$}]{\includegraphics[width=0.45\textwidth]{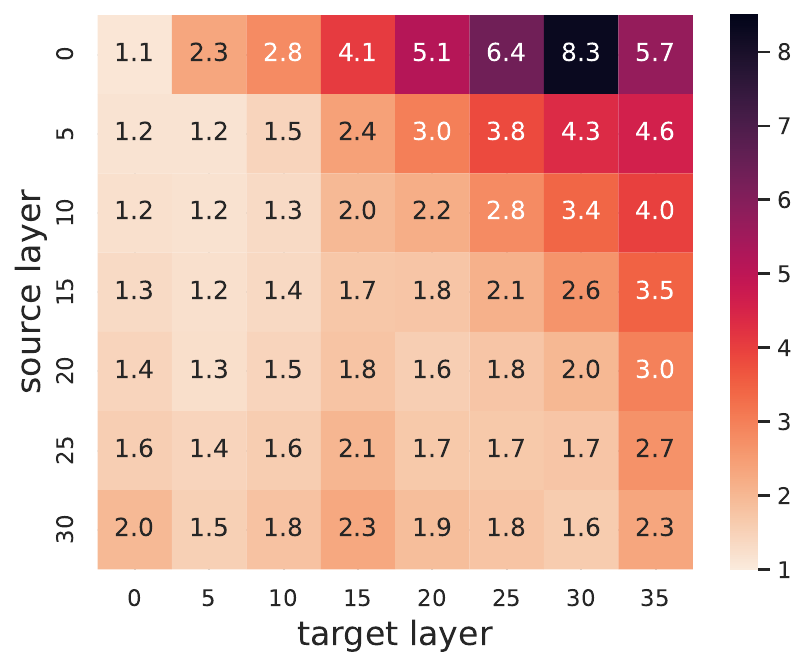}}
    }
\caption{Surprisal scores ($\uparrow$ is better) on next-token prediction estimation in Vicuna and Pythia cross-model \methodFamilyName{}.}
\label{fig:cross_model_ntp_surprisal}
\end{figure}

\begin{table}[ht!]
\caption{Additional qualitative examples illustrating entity resolution in \textbf{Vicuna 13B}. The expressive generations show that as we go through the layers, more tokens from the context get integrated into the current representation, referred to as ``Tokens Covered". Here, $\sourceModel \is \text{Vicuna 13B}, \targetModel \is \sourceModel, \targetLayer \is \sourceLayer$.}
\label{tab:entity-resolution-vicuna-samples}
\def\arraystretch{1.85}
\vspace{0.5em}
\resizebox{0.95\textwidth}{!}{%
\begin{tabular}{cllm{40em}m{12em}}
    \toprule
    $\sourcePrompt$ & Tokens Covered & $\sourceLayer$  & Generation    & Notes/Entity Described by the Generation\\
    \toprule
    \multirow{3.5}{*}{\rotatebox{90}{\texttt{"Will Smith"}}}  &
    \texttt{"Smith"} & 
    1 &
    sonian: National museum and research complex in Washington, D.C. &
    \textbf{Smith}sonian Museum \\
    &
    \texttt{"Smith"} & 
    2 &
    sonas: American rock band &
    An amalgam of \textbf{Smith} band and SONA band\\
    &
    \texttt{"Will Smith"} &
    3 &
    : American actor and rapper &
    \textbf{Will Smith} \\
\midrule
    \multirow{6}{*}{\rotatebox{90}{\texttt{"Queen Victoria"}}}  &
    \texttt{"Victoria"} & 
    1 &
    s Secret: American multinational women's lingerie, beauty and body-care products company &
    \textbf{Victoria}'s Secret \\
    &
    \texttt{"Victoria"} & 
    2 &
    s Secret: American lingerie and beauty products retailer &
    \textbf{Victoria}'s Secret \\
    &
    \texttt{"Queen Victoria"} &
    3 &
    : Queen of the United Kingdom of Great Britain and Ireland from 1837 to 1901 &
    \textbf{Queen Victoria} \\
    &
    \texttt{"Queen Victoria"} & 
    4-8 &
    : Queen of the United Kingdom of Great Britain and Ireland from 1837 until her death in 1901 &
    \textbf{Queen Victoria} \\
    &
    \texttt{"Queen Victoria"} &
    9 &
    : Queen of England (reigned from 1837-1901) &
    \textbf{Queen Victoria} \\
    &
    \texttt{"Queen Victoria"} &
    10 &
    : 19th century British queen who ruled from 1837-1901 &
    \textbf{Queen Victoria} \\
\midrule
    \multirow{8.5}{*}{\rotatebox{90}{\texttt{"Paris Hilton"}}}  &
    \texttt{"ton"} & 
    1 &
    y: a type of pasta &
    Riga\textbf{ton}i/Riga\textbf{ton}y/Riga\textbf{ton}\'{e} \\
    &
    \texttt{"Hilton"} & 
    2 &
    Hotels: American multinational hospitality company &
    \textbf{Hilton} Hotel \\
    &
    \texttt{"Hilton"} &
    3 &
    : Hotel chain founded by Conrad Hilton &
    \textbf{Hilton} Hotel \\
    &
    \texttt{"Hilton"} &
    4 &
    : Hotel and resorts company &
    \textbf{Hilton} Hotel\\
    &
    \texttt{"Paris Hilton"} &
    5-6 &
    : Socialite and television personality &
    \textbf{Paris Hilton} \\
    &
    N/A &
    7 &
    : Placeholder for a variable or concept &
    Placeholder contamination \\
    &
    N/A &
    8-9 &
    : Variable representing any number of things or concepts &
    Placeholder contamination \\
    &
    N/A &
    10 &
    : x is a placeholder & 
    Placeholder contamination \\
\midrule
    \multirow{8.5}{*}{\rotatebox{90}{\texttt{"James Bond"}}}  &
    \texttt{"Bond"} & 
    1 &
    : Unit of measurement for the strength of a liquid &
    \textbf{Bond} Unit \\
    &
    \texttt{"James Bond"} & 
    2 &
    : James Bond, the fictional British secret agent created by novelist Ian Fleming and portrayed by actor Daniel Craig in the 2006 film "Casino Royale," &
    \textbf{James Bond} \\
    &
    \texttt{"James Bond"} &
    3-5 &
    : Fictional British secret agent &
    \textbf{James Bond} \\
    &
    \texttt{"James Bond"} &
    6 &
    : Code name for fictional British secret agent James Bond &
    \textbf{James Bond}\\
    &
    \texttt{"James Bond"} &
    7 &
    : Code name for a fictional British secret agent created by novelist Ian Fleming &
    \textbf{James Bond} \\
    &
    \texttt{"James Bond"} &
    8 &
    : Subject of a fictional British secret agent, created by novelist Ian Fleming and portrayed in a series of films. &
    \textbf{James Bond} \\
    &
    \texttt{James Bond} &
    9 &
    : Subject of the 007 novels and films &
    \textbf{James Bond} \\
    &
    N/A &
    10 &
    : Variable representing the number of the film in the James Bond series & 
    Placeholder contamination \\
\midrule
    \multirow{8.5}{*}{\rotatebox{90}{\texttt{"Dwayne Johnson"}}}  &
    \texttt{"Johnson"} & 
    1-2 &
    \& Johnson: American multinational corporation in the fields of pharmaceuticals, medical devices, and consumer packaged goods &
    \textbf{Johnson} \& Johnson \\
    &
    \texttt{"Dwayne Johnson"} & 
    3-4 &
    : American actor and professional wrestler, also known as "The Rock" &
    \textbf{Dwayne Johnson} \\
    &
    \texttt{"Dwayne Johnson"} &
    5 &
    : Also known as "The Rock," American actor and professional wrestler &
    \textbf{Dwayne Johnson} \\
    &
    \texttt{"Dwayne Johnson"} &
    6 &
    : Former professional wrestler and actor &
    \textbf{Dwayne Johnson}\\
    &
    N/A &
    7 &
    : Abbreviation for the United States &
    Placeholder contamination \\
    &
    N/A &
    8 &
    : Variable representing any number of things or concepts &
    Placeholder contamination \\
    &
    N/A &
    9 &
    : Variable representing any number of other options. &
    Placeholder contamination \\
    &
    N/A &
    10 &
    : Variable representing any number of other options or possibilities.
    &
    Placeholder contamination\\
  \bottomrule
\end{tabular}%
}
\end{table}

\begin{table}[ht!]
\caption{Additional qualitative examples illustrating entity resolution in \textbf{Pythia 12B}. The expressive generations show that as we go through the layers, more tokens from the context get integrated into the current representation, referred to as ``Tokens Covered". Here, $\sourceModel \is \text{Pythia 12B}, \targetModel \is \sourceModel, \targetLayer \is \sourceLayer$.}
\label{tab:entity-resolution-pythia-samples}
\def\arraystretch{1.85}
\vspace{0.5em}
\resizebox{0.95\textwidth}{!}{%
\begin{tabular}{cllm{28em}m{17em}}
\toprule
    $\sourcePrompt$ & Tokens Covered & $\sourceLayer$  & Generation    & Notes/Entity Described by the Generation\\
    \toprule
    \multirow{2}{*}{\rotatebox{90}{\parbox{4em}{\texttt{"Will Smith"}}}} &
    \texttt{"Smith"} &
    1  &
     \& Wesson: American firearms manufacturer &
    \textbf{Smith} \& Wesson \\
    &
    \texttt{"Will Smith"} &
    2    &
    : American actor     &
    \textbf{Will Smith} \\
\midrule
    \multirow{8}{*}{\rotatebox{90}{$\sourcePrompt \is \texttt{"Alexander\;the\;Great"}$}} &
    \texttt{"Great"} &
    1  &
    Britain: Country in the European Union &
    \textbf{Great} Britain \\
    &
    \texttt{"the Great"} &
    2    &
    Wall Street Crash of 1929: Financial crisis in the United States     &
    \textbf{the Great} Depression \\
    &
    \texttt{"the Great"} &
    3 &
    Wall Street Bubble: The Great Depression &
    \textbf{the Great} Depression \\
    &
    \texttt{"the Great"} &
    4 &
    Wall Street: Wall Street in New York City  &
    Wall Street \newline (related to the Great Depression)\\ 
    &
    \texttt{"Alexander the Great"} &
    5 &
    : Ancient Greek ruler, and the first to rule all of the then known world  &
    \textbf{Alexander the Great}\\ 
\midrule
    \multirow{8}{*}{\rotatebox{90}{\texttt{"George Washington"}}} &
    \texttt{"Washington"} &
    1  &
    : Capital of the United States &
    \textbf{Washington} D.C. \\
    &
    \texttt{"Washington"} &
    2    &
    : American capital city     &
    \textbf{Washington} D.C. \\
    &
    \texttt{"George\;Washington"} &
    3 &
    : American president &
    \textbf{George Washington} \\
    &
    \texttt{"George\;Washington"} &
    4-6 &
    : American revolutionary & 
    \textbf{George Washington} \\ 
    &
    \texttt{"George\;Washington"} &
    7 &
    : American president & 
    \textbf{George Washington} \\
    &
    \texttt{"George\;Washington"} &
    8 &
    : George Washington & 
    \textbf{George Washington} \\ 
    &
    \texttt{"George\;Washington"} &
    9 &
    : George Washington, the first President of the United States & 
    \textbf{George Washington} \\
    &
    \texttt{"George\;Washington"} &
    10 &
    : George Washington, Thomas Jefferson, Abraham Lincoln, and Theodore Roosevelt: American presidents & 
    \textbf{George Washington} \\
\midrule
    \multirow{6}{*}{\rotatebox{90}{\texttt{"Titanic"}}} &
    \texttt{"c"} &
    1  &
    loud: Apple's cloud-based storage service &
    \textbf{c}loud \\
    &
    \texttt{"Titanic"} &
    2    &
    : Sinking ocean liner     &
    \textbf{Titanic} \\
    &
    \texttt{"Titanic"} &
    3 &
    : The ship that sank in the Atlantic Ocean in 1912 &
    \textbf{Titanic} \\
    &
    \texttt{"Titanic"} &
    4-6 &
    : British passenger ship & 
    \textbf{Titanic} \\ 
    &
    \texttt{"Titanic"} &
    7-9 &
    : The ship that sank & 
    \textbf{Titanic} \\
    &
    \texttt{"Titanic"} &
    10 &
    : Titanic, Titanic: British luxury passenger liner & 
    \textbf{Titanic} \\ 
\midrule
    \multirow{5}{*}{\rotatebox{90}{\parbox{4em}{\texttt{"Saturday Night\;Live"}}}} &
    \texttt{"Live"} &
    1  &
     Nation: American concert promoter &
    \textbf{Live} Nation \\
    &
    \texttt{"Live"} &
    2    &
     Nation: American television network     &
    \textbf{Live} Nation\\
    &
    \texttt{"Saturday Night Live"} &
    3 &
    : American television program &
    \textbf{Saturday Night Live} \\
    &
    \texttt{"Saturday Night Live"} &
    4-10 &
    : American television comedy show & 
    \textbf{Saturday Night Live} \\ 
\midrule
    \multirow{9.5}{*}{\rotatebox{90}{\texttt{"Nineteen Eighty-Four"}}} &
    \texttt{"Four"} &
    1  &
      Seasons: American hotel chain &
    \textbf{Four} Seasons Hotel \\
    &
    \texttt{"Four"} &
    2    &
      Seasons: The four seasons of the year     &
    \textbf{Four} Seasons (of the year)\\
    &
    \texttt{"ighty-Four"} &
    3 &
     Lions: Chinese professional football club &
    Cangzhou M\textbf{ighty} Lions F.C. \\
    &
    \texttt{"Nineteen Eighty-Four"} &
    4 &
    : Number of the novel 1984 by George Orwell & 
    \textbf{Nineteen Eighty-Four} (Novel) \\ 
    &
    \texttt{"Nineteen Eighty-Four"} &
    5 &
     : George Orwell's novel, and the 1984 film. &
     \textbf{Nineteen Eighty-Four} (Novel \& Film) \\
    &
    \texttt{"Nineteen Eighty-Four"} &
    6 &
    : George Orwell’s novel & 
    \textbf{Nineteen Eighty-Four} (Novel) \\ 
    
    &
    \texttt{"Nineteen Eighty-Four"} &
    7 &
     : George Orwell: British novelist &
     \textbf{Nineteen Eighty-Four} (Novel) \\
    &
    \texttt{"Nineteen Eighty-Four"} &
    8 &
    : 1984 novel by George Orwell & 
    \textbf{Nineteen Eighty-Four} (Novel) \\ 
    &
    \texttt{"Nineteen Eighty-Four"} &
    9-10 &
    : 1984 novel by George Orwell, and the 1984 film adaptation directed by Michael Radford. & 
    \textbf{Nineteen Eighty-Four} (Novel \& Film) \\ 
  \bottomrule
\end{tabular}%
}
\end{table}

\section{Additional Results on the Cross-Model Patching Experiment}
\label{sec:appendix-cross-model-ntp}

Fig.~\ref{fig:cross_model_ntp} depicts the Precision@1 scores for different combinations of source-target layers, showing that patching with a simple affine \methodName{} proves to be effective with precision of up to 0.7 and 0.8 for Vicuna and Pythia, respectively. Specifically, patching representations to an early layer of the larger model seems to be the most effective.
Furthermore, it appears that there is a subtle matching among some layers of the models, with the diagonal consistently exhibiting higher values. 
As shown in Fig.~\ref{fig:cross_model_ntp_surprisal}, similar trends are observed in the surprisal results for both models. Overall, these results show that, when $\targetModel$ and $\sourceModel$ are from the same model family, it is possible to leverage $\targetModel$ for decoding information from the representations of $\sourceModel$. Notably, our findings extend observations by \citet{csiszarik2021similarity} from patching representations across deep convolutional neural networks with the same architecture but different initializations.

\section{More Details on the Multi-Hop Reasoning Experiment}
\label{sec:appendix-cot}

\paragraph{Data} Building on \citet{hernandez2023linearity}, we systematically generate all valid multi-hop factual and commonsense reasoning queries where the object $\object_1$ in the first triplet $\tau_1$ is equal to the subject $\subject_2$ in the second triplet $\tau_2$. This process yields 1,104 multi-hop reasoning samples. We then filter examples to the ones where the model accurately represents both triplets independently. Tab.~\ref{tab:multihop-samples} summarizes information about the samples where Vicuna (13B) correctly represents each reasoning step. Out of 1,104 samples that require two steps of commonsense or factual reasoning, 46 satisfy the above criteria, with 8 unique relation combinations.

\begin{table}[t!]
\caption{Sample statistics for the multi-hop reasoning experiment where $\sourceModel$ correctly represents both $\tau_1$ and $\tau_2$.}
\label{tab:multihop-samples}
\vspace{0.5em}
\centering
\resizebox{.5\textwidth}{!}{%
\begin{tabular}{lll}
\hline
 $\relation_1$      & $\relation_2$        & \# Samples \\ \hline
 Company CEO        & Person Father        & 4     \\
 Food from Country  & Country Capital City & 10    \\
 Food from Country  & Country Currency     & 3     \\
 Food from Country  & Country Language     & 9     \\
 Food from Country  & Country Largest City & 11    \\
 Person Father      & Person Father        & 1     \\
 Person Father      & Person Mother        & 1     \\
 Product by Company & Company CEO          & 7     \\ \hline
 Total              &                      & 46    \\ \hline
\end{tabular}%
}
\end{table}

\paragraph{More Detailed Results}

\begin{figure}[t!]
    \centering
    \includegraphics[width=0.5\textwidth]{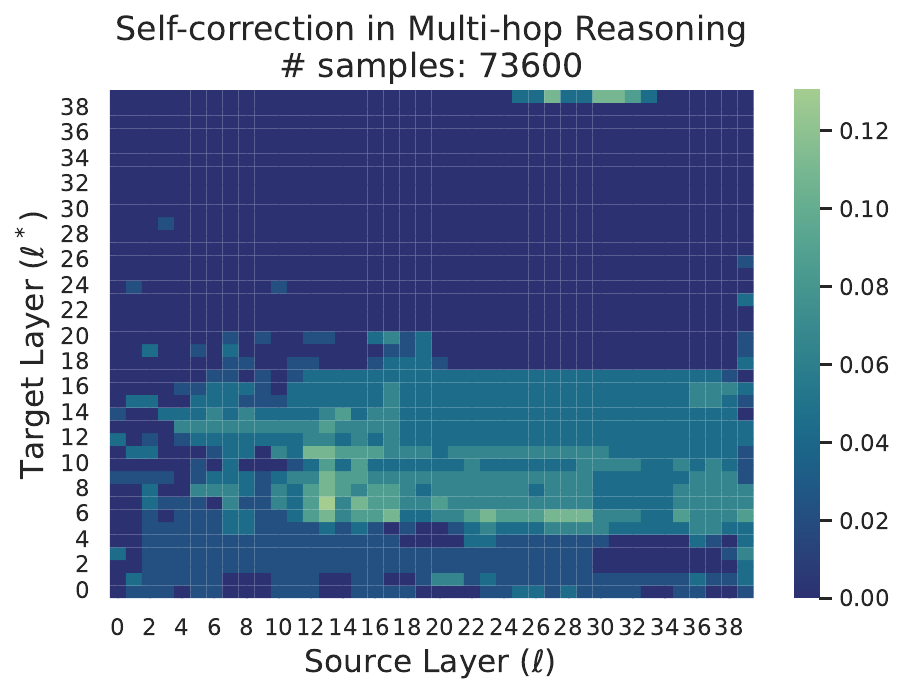}
    \caption{The interaction between source ($\sourceLayer$) and target ($\targetLayer$) layers in correcting multi-hop reasoning errors. The majority of success cases correspond to early-to-mid $\targetLayer$. We observe higher cumulative success rate in the lower right triangle which corresponds to $\targetLayer \le \sourceLayer$.}
    \label{fig:cot_heatmap}
\end{figure}

Fig.~\ref{fig:cot_heatmap} shows the interaction between $\sourceLayer$ and $\targetLayer$ and how it affects the success rate. Patching representations from most source layers $\sourceLayer$ into early-to-mid $\targetLayer$ (6-16) is the most effective in making the right prediction. Our interpretation is that patching into late $\targetLayer$ is not effective because $\relation_2$ has already been processed and the result has been copied to the last position, therefore it cannot incorporate the proxy of $\subject_2$. We also observe that $\targetLayer \le \sourceLayer$ is more successful on average.

\section{Compute Resources}
All experiments were conducted using A100 80GB GPUs, with the exception of experiments with GPT-J \cite{gpt-j} where we used A100 40GB GPUs.


\end{document}